\documentclass{article}

\usepackage{arxiv}
\usepackage[square,numbers]{natbib}
\usepackage[title]{appendix}
\usepackage[utf8]{inputenc} 
\usepackage[T1]{fontenc}    
\usepackage{hyperref}       
\usepackage{url}            
\usepackage{booktabs}       
\usepackage{amsfonts}       
\usepackage{nicefrac}       
\usepackage{microtype}      
\usepackage{lipsum}		
\usepackage{graphicx}
\usepackage{natbib}
\usepackage{doi}
\usepackage{xspace}
\usepackage{amsmath}
\usepackage[table]{xcolor}
\usepackage{multirow}
\usepackage{bm}

\title{SteerPose: Simultaneous Extrinsic Camera Calibration and Matching from Articulation}

\author{%
  Sang-Eun Lee \\
  Graduate School of Informatics\\
  Kyoto University\\
  \And
  Ko Nishino \\
  Graduate School of Informatics\\
  Kyoto University\\
  \AND
  Shohei Nobuhara \\
  Information and Human Sciences\\
  Kyoto Institute of Technology\\
}

\newcommand{\bmvaOneDot}{.\xspace}
\newcommand{\wrt}{w.r.t.\ }

\def\eg{\emph{e.g}\bmvaOneDot}

\def\etal{\emph{et al}\bmvaOneDot}
\def\ie{\emph{i.e}\bmvaOneDot}
\def\etc{\emph{etc}\bmvaOneDot}

\def\Dpig{\textit{Bama Pig}\xspace}
\def\Ddog{\textit{Beagle Dog}\xspace}
\def\Dpigeon{\textit{Pigeon}\xspace}
\def\Dcheetah{\textit{Cheetah}\xspace}

\def\Dtoddler{\textit{Toddler}\xspace}

\def\Dvolleyball{\textit{Volleyball}\xspace}
\def\DMaster{MASt3R\xspace}

\begin{document}
\maketitle
\begin{figure}[htbp]
    \centering
    \includegraphics[width=\linewidth]{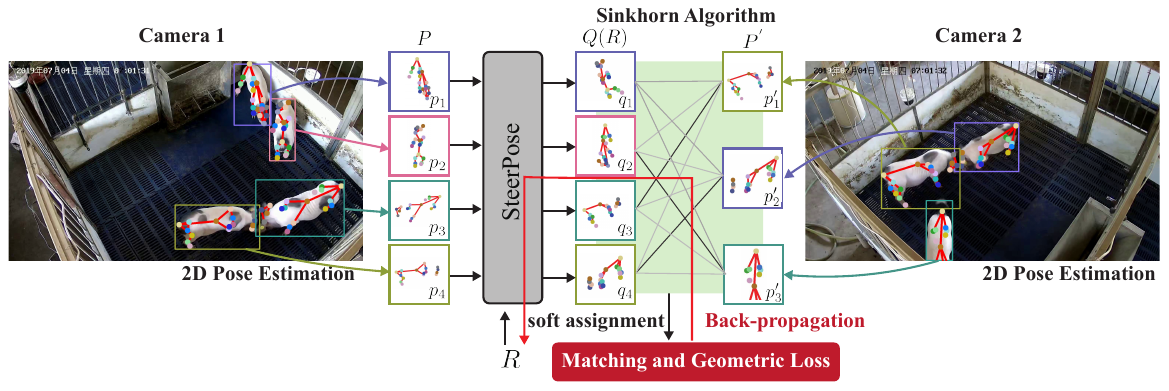}
    \caption{Camera calibration by \textit{SteerPose}.  Given a set of 2D poses $P$ of targets with known articulation, SteerPose predicts their appearance under a relative camera rotation $R$ as $Q(R)$ which can be matched against 2D poses $P'$ observed in another view in a differentiable manner. Novel matching and geometric losses evaluate the validity of $R$ as well as the matching between $Q(R)$ and $P'$, and their gradients are back-propagated to $R$ to optimize the relative rotation between the views and the matching simultaneously.}
\label{fig:opening}
\end{figure}

\begin{abstract}
Can freely moving humans or animals themselves serve as calibration targets for multi-camera systems while simultaneously estimating their correspondences across views?  We humans can solve this problem by \emph{mentally rotating} the observed 2D poses and aligning them with those in the target views.  Inspired by this cognitive ability, we propose \emph{SteerPose}, a neural network that performs this rotation of 2D poses into another view. By integrating differentiable matching, \emph{SteerPose} simultaneously performs extrinsic camera calibration and correspondence search within a single unified framework. We also introduce a novel geometric consistency loss that explicitly ensures that the estimated rotation and correspondences result in a valid translation estimation. Experimental results on diverse in-the-wild datasets of humans and animals validate the effectiveness and robustness of the proposed method. Furthermore, we demonstrate that our method can reconstruct the 3D poses of novel animals in multi-camera setups by leveraging off-the-shelf 2D pose estimators and our class-agnostic model.
\end{abstract}


\section{Introduction}

Motion capture is crucial for understanding human or animal behavior and social interactions, with a wide range of applications in fields such as biomechanics, psychology, and engineering~\cite{naik20233d, waldmann20243d,an2023three,Sigal2010HumanEvaSV,nath2019using,lauer2022multi}. Multi-camera systems facilitate accurate motion capture without the need for intrusive markers; however, they require specialized tools and a cumbersome process of calibrating all cameras prior to on-site deployment. Furthermore, in the case of multiple objects in a scene, their correspondence across views should be estimated for 3D triangulation.  These requirements limit the applicability of markerless motion capture, especially when studying subjects with unpredictable behavior in uncontrolled environments, such as animals in the wild or elderly people in their daily lives.

Can freely moving humans or animals themselves serve as calibration targets for multi-camera systems while simultaneously estimating their correspondences? If so, we can streamline motion capture from uncalibrated 
multi-view videos of freely moving animals such as livestock pigs (Fig.~\ref{fig:opening}), eliminating the need for manual calibration and correspondence search.

This paper proposes a novel method for automatic extrinsic camera calibration and correspondence search
which leverages the inherent articulated structure of moving subjects. By exploiting robust geometric constraints derived from these articulations, our method establishes reliable cross-view correspondences without requiring specialized tools and procedures. This enables markerless motion capture of freely moving subjects using only multi-view video sequences captured in unconstrained environments, thereby improving the flexibility and accessibility of motion capture.

Our key idea is inspired by \textit{mental rotation}~\cite{shepard1971mental}, a fundamental cognitive process in the human brain. When aligning an object in two images captured from different views, we can mentally rotate the object until it matches the reference view. Inversely, we can understand the relative pose between the views through this process.  Furthermore, in the case of images with multiple objects, we can simultaneously understand their correspondences across views that are consistent with the estimated rotation.

Mimicking this psychological mechanism, we propose a neural model called \textit{SteerPose}, which transforms a 2D pose from one view to another, given a hypothesized relative rotation between them. SteerPose realizes a rotation-covariant transform. Following Marcos \etal~\cite{Marcos_2017_ICCV}, a function $f(\cdot)$ is said to be rotation-covariant \wrt a transform $g(\cdot)$ if $f(g(\cdot)) = g'(f(\cdot))$ holds, where $g'(\cdot)$ is a second transform, which is itself a function of $g(\cdot)$. Considering a 3D-to-2D projection and a rotation in 3D as $f(\cdot)$ and $g(\cdot)$, SteerPose serves as $g'(\cdot)$ which \textit{mentally rotates} the 2D projection of the latent 3D pose $f(\cdot)$ to be identical to that rotated in 3D by $g(\cdot)$ before projection.

By combining SteerPose with a differentiable bipartite matching module, we propose an end-to-end framework for simultaneous optimization of extrinsic camera calibration and correspondence search as shown in Fig.~\ref{fig:opening}.  Consider two sets of 2D poses of a target, such as humans or animals, extracted from images captured from different viewpoints using an off-the-shelf 2D pose estimator.
Given a current estimate of the 3D rotation $R$ between these views, SteerPose transforms the 2D poses of one side $p_i \in P$ to the other viewpoint as $q_i(R) \in Q(R)$.  A matching module, such as Sinkhorn~\cite{cuturi2013sinkhorn}, finds the best match between $q_i(R) \in Q(R)$ and the 2D poses $p' \in P'$ observed on the other side based on the similarity between 2D poses.  In addition to this 2D pose similarity loss, the estimated match is evaluated using our novel geometric loss, which guarantees that the estimated rotation $R$ has a valid solution for translation.  By optimizing these losses by refining $R$ through back-propagation, we can estimate the relative pose between the views while simultaneously estimating the correspondences.

Our approach frames camera pose estimation and cross-view association as a single unified optimization problem. The key is to use body articulations as \textit{structured} semantic keypoints.  SteerPose learns how the 2D arrangement of body joint positions changes after rotations, and the predicted 2D poses can be used as \textit{structured} correspondences in the matching module.  We argue that body articulation can serve as general prior knowledge valid across a wide variety of targets as shown in Table~\ref{tab:2view_geometric}.

We evaluated our method using diverse datasets covering both humans and several animal species, including dogs, pigs, pigeons, and cheetahs. The experimental results demonstrate that our approach outperforms both traditional epipolar geometry-based and state-of-the-art learning-based methods.  We also quantitatively evaluated the contribution of our novel geometric loss by showing that the estimated rotation without this loss can be trapped by a false match that is valid in terms of 2D pose rotations but is inconsistent in terms of translation.

A notable application of the proposed method is to provide training datasets for image-based monocular 3D pose estimators of novel animals.  As long as the target novel animal shares the same body articulation structure and their 2D poses can be estimated from captured images, SteerPose automatically finds corresponding instances from different views while calibrating the relative camera pose simultaneously. This result allows for triangulating the 3D poses, which can be used as a pseudo ground-truth for training monocular 3D pose estimators. We demonstrate that by leveraging class-agnostic SteerPose and simply capturing freely moving quadruped animals not included in the training data, we can achieve multi-view calibration and establish instance matches across views, resulting in the 3D reconstruction of these novel quadrupeds.

\begin{figure}
    \centering
    \includegraphics[width=0.7\linewidth]{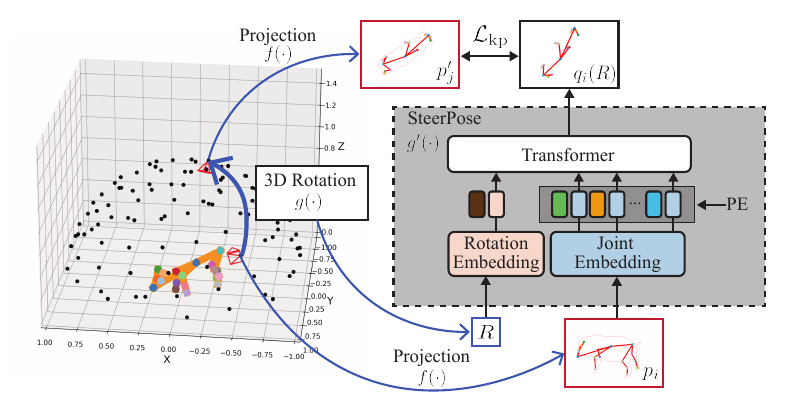}
    \caption{\textit{SteerPose} learns to \textit{rotate} an input 2D pose $p_i$ to the 2D pose $q_i(R)$ observed from another view at relative rotation $R$ in a fully-supervised manner.  SteerPose serves as a function $g'(\cdot)$ that performs \textit{mental rotation} to form a rotation-covariant transformation of 3D poses under projection $f(\cdot)$ satisfying $f(g(\cdot))=g'(f(\cdot))$ where $g(\cdot)$ is the rotation in 3D by $R$.
    The training dataset was synthesized by rendering 3D poses from random viewpoints.
    }
    \label{fig:steerpose_train}
\end{figure}

\section{Related Work}

Extrinsic calibration is the process of estimating the relative poses and positions of cameras. Conventional approaches require specialized reference objects with known geometries such as checkerboards, AR markers, or calibration wands~\cite{zhang2000flexible,aruco2014,mitchelson2003wand}. Since procedures using these tools are usually performed on-site prior to deployment~\cite{Joo_2017_TPAMI,hasler2009markerless,Sigal2010HumanEvaSV,an2023three}, these calibration methods face challenges in ``in-the-wild" scenarios.

Traditional SfM-based methods~\cite{pan2024global,schoenberger2016sfm} utilize static background features such as SIFT~\cite{Lowe:SIFT}.  While this approach is known to be robust for moving cameras such as SLAM~\cite{taketomi2017visual}, it becomes challenging in wide-baseline setups such as motion capture due to appearance changes across views. In contrast, the use of dynamic foreground objects, in particular humans themselves, as calibration targets exploits semantic information such as their inherent motion and structure. This approach not only increases the robustness of the calibration but also eliminates the need for specialized calibration tools and tedious procedures~\cite{leeRAL2022,Takahashi18,yan2021wide-baseline,nakano2021,Garau2020,lee2025spatiotemporal}.  These methods, however, require correspondences of the targets between views, and/or a monocular 3D pose estimator.

Recent studies have integrated learnable components into camera pose estimation. Leveraging learned priors for keypoint detection, feature extraction, and matching~\cite{detone2018superpoint,sarlin2020superglue,lindenberger2023lightglue} has demonstrated improved performance in camera pose estimation by establishing reliable correspondences across different views, particularly in wide-baseline scenarios. \textit{Steerers}~\cite{bokman2024steerers} facilitate keypoint matching in scenarios involving rotations by learning a rotation-equivariant linear mapping in the descriptor space. The Differentiable 8-point algorithm~\cite{roessle2023end2end} unifies feature matching and camera pose estimation in an end-to-end framework, holistically addressing these challenges while eliminating the need for RANSAC~\cite{fischler_bolles:RANSAC}. More recent approaches, such as DUSt3R~\cite{dust3r_cvpr24} and \DMaster~\cite{mast3r_arxiv24}, address correspondence estimation, and 3D reconstruction in a single pipeline by directly regressing dense 3D point maps from unposed images, enabling camera pose estimation without explicit correspondence matching.

Compared with these existing methods, the proposed method realizes simultaneous optimization of camera poses and correspondences using foreground objects as a calibration target, which provides a structured correspondence of semantic features, \ie, joints, across views. A summary is provided in Appendix~\ref{sec:comp_others}

\section{Two-view Extrinsic Calibration}

\subsection{SteerPose}
The goal of SteerPose is to \textit{mentally rotate} a 2D pose found in an image into a 2D pose in another view at rotation $R$. SteerPose realizes a rotation-covariant transform~\cite{Marcos_2017_ICCV} of the latent 3D pose behind the observed 2D pose.

As shown in Fig.~\ref{fig:steerpose_train}, we employ a Transformer~\cite{vaswani2017attention} architecture that encodes articulated poses via multi-head self-attention and feed-forward networks to learn such a mapping.
Relative camera rotation $R$ is embedded into a 32-dim vector as a token to specify the target view into which SteerPose transforms the input 2D poses.
For each pose, we construct 32-dim input tokens by combining the embeddings of the keypoints with learnable positional encodings that capture the hierarchical structure of the articulated body. The model learns to capture both local joint relationships and its global pose structure by processing these tokens through the self-attention layers, making it inherently robust to partial occlusions. 

We train SteerPose in a fully-supervised manner for each class of animals such as humans, quadrupeds, birds, \etc.  Suppose we have a 3D pose dataset of the target.  For each 3D pose, we synthesize random pairs of two views on a unit sphere, and project the 3D pose onto such views.  We can synthesize as many random pairs of 2D poses with known relative rotations for each 3D pose as possible, and we can repeat this for the 3D poses in the dataset.

Using these synthesized pairs, we train SteerPose to minimize the average 2D keypoint displacement $\mathcal{L}_\mathrm{kp}$ defined as
\begin{equation}
    \mathcal{L}_\mathrm{kp}(R) = \frac{1}{M} \textstyle\sum_{i=1}^M D(q_i(R), p'_i)\,, \label{eq:l_kp}
\end{equation}
where $M$ denotes the number of synthesized training pairs. $q_i(R)$ and $p'_i$ denote the 2D pose transformed by SteerPose given $R$ and the target 2D pose respectively.
$D(\cdot)$ computes the dissimilarity between 2D poses as the L2 distance between them. During training, we injected 2D random noise into the input 2D joint positions to mimic ambiguities in joint detection and randomly masked tokens to account for missing joints due to occlusions.

We assume that SteerPose is pretrained for the target class of animals hereafter.  That is, given a 2D pose and the relative rotation to a target view, SteerPose can infer the 2D pose in the target view.  This point is quantitatively evaluated in Sec.~\ref{sec:eval_ransac}.

\subsection{Joint Extrinsic Camera Calibration and Matching}

We simultaneously solve the extrinsic camera calibration and 2D pose matching as an inference-time optimization utilizing a neural network with a pretrained SteerPose and differentiable Sinkhorn-based matching~\cite{cuturi2013sinkhorn} as shown in Fig.~\ref{fig:opening}.  Suppose we have two sets of 2D poses $P = \{p_i\} \; (i=1,\dots,N)$ and $P' = \{p'_{i'}\} \; (i'=1,\dots,N')$ of the target estimated from images captured at different viewpoints $C$ and $C'$.  By specifying a relative rotation $R$ from $C$, the pretrained SteerPose transforms each 2D pose $p_i \in P$ of $C$ to those at another view at rotation $R$ as $q_i(R) \in Q(R)$, and  Sinkhorn algorithm can find the best correspondences between the transformed 2D poses $q_i(R) \in Q(R)$ and the observed 2D poses $p'_{i'} \in P' \; (i'=1,\dots,N')$ at the other view $C'$, based on the similarity between the 2D poses.  If the specified rotation $R$ is identical to the relative rotation between $C$ and $C'$, we can expect that each 2D pose $q_i(R)$ has a similar pose in $P'$, and the Sinkhorn algorithm can find their correspondences correctly.
Note that in the case of $N \neq N'$ due to occlusions, for example, we add dummy targets as described in~\cite{sinkhorn}.

Using this network, we realize simultaneous extrinsic calibration and 2D pose matching as optimization by back-propagation.  Suppose we start the optimization with an initial estimate of the relative rotation $R$, and let the matching module find the best correspondences between the 2D poses under the current $R$.  The key idea is to evaluate the geometric consistency of the current rotation $R$ and the estimated correspondences, in addition to the matching cost used in the correspondence search, using the loss function defined later, and to refine $R$ by back-propagating its gradient while keeping the pretrained SteerPose frozen.  This is an inference-time optimization, and iteratively refining $R$  simultaneously improves the correspondence search in the matching module.

\paragraph{Matching loss}

The correspondence search between the 2D pose sets $Q(R)$ and $P'$ using Sinkhorn algorithm in the matching module assigns a weight to each of the possible combinations of the 2D poses between $Q(R)$ and $P'$. The similarity score for each possible pair of 2D poses $q_j(R) $ and $p'_j $ is defined as   
$s(q_j(R), p_j') = \frac{2}{1+\exp(\alpha\mathcal{L}_\mathrm{kp}(R))}$, where $\alpha$ is a scaling parameter. Denoting the weight and 2D poses of the $j$th pair by $w_j$, $q_j(R) \in Q(R)$, and $p'_j \in P'$, we evaluate the following matching loss to assess the confidence of the correspondence search:
\begin{equation}
    \mathcal{L}_\mathrm{match} = \textstyle\sum_{j=1}^J w_j (1 - s(q_j(R), p'_j))\,.
\end{equation}
To enforce cycle consistency between views, we compute the matching loss bidirectionally.

\paragraph{Geometric consistency loss}

The geometric consistency loss $\mathcal{L}_\mathrm{geom}$ evaluates whether the combination of the current $R$ and the estimated matches yields a valid solution of the relative translation, after enforcing collinearity and coplanarity constraints ~\cite{Hartley00,leeRAL2022}. For all $N$ corresponding 2D pose pairs under current rotation $R$, we transform their positions into a normalized coordinate system. By stacking linear equations derived from collinearity and coplanarity constraints~\cite{leeRAL2022}, we can obtain a linear system of the form $A \begin{bmatrix} x_1 & \dots & x_N & \bm{t} \end{bmatrix}^\top = \bm{0}$,
where unknowns $x_i$ are 3D poses recovered from the collinearity constraint and $\bm{t}$ is determined up to scale and sign ambiguity. This sign ambiguity is resolved via a chirality check~\cite{Hartley00}.

Since the partial derivatives of the singular values $\sigma_2 \ge \sigma_1 $ can be computed in closed-form~\cite{Ionescu_2015_ICCV,roessle2023end2end}, we include the ratio $\sigma_1 / \sigma_2$ as a loss $\mathcal{L}_\mathrm{geom}$ in the optimization of $R$, to guarantee that the estimated $R$ makes the smallest singular value as small as possible, and hence it has a solution of $\bm{t}$.

Consequently, we optimize $\mathcal{L}_\mathrm{match} + \lambda \mathcal{L}_\mathrm{geom}$ as the loss function, where $\lambda$ is an empirically determined weighting parameter. 

\section{Multi-view Extrinsic Calibration}
In the case of multiple cameras, we can simply calibrate them in a pair-wise manner as described in the previous section, and then integrate them into a unified coordinate system by employing motion averaging~\cite{govindu2001combining} and cycle consistency-based matching~\cite{dong2021fast}. The integrated extrinsic parameters can be further optimized by a nonlinear optimization that minimizes the reprojection errors~\cite{Hartley00}.  

\section{Experiments}

\subsection{Experimental Setups}

\paragraph{Datasets} 

\begin{figure}[ht]
    \centering
    \def\vlab#1{\rotatebox[origin=c]{90}{#1}}
    \newcommand{\incg}[2]{%
  \raisebox{-0.5\totalheight}{%
    \includegraphics[width=0.13\linewidth]{#1}%
    \includegraphics[width=0.13\linewidth]{#2}%
  }%
  }
    \begin{tabular}{c|@{}c@{}c@{}c@{}c@{}}
    & LightGlue~\cite{lindenberger2023lightglue} & \DMaster~\cite{mast3r_arxiv24} & Ours & 3D Pose (GT) \\
    \hline
    \vlab{\Dcheetah} &
    \incg{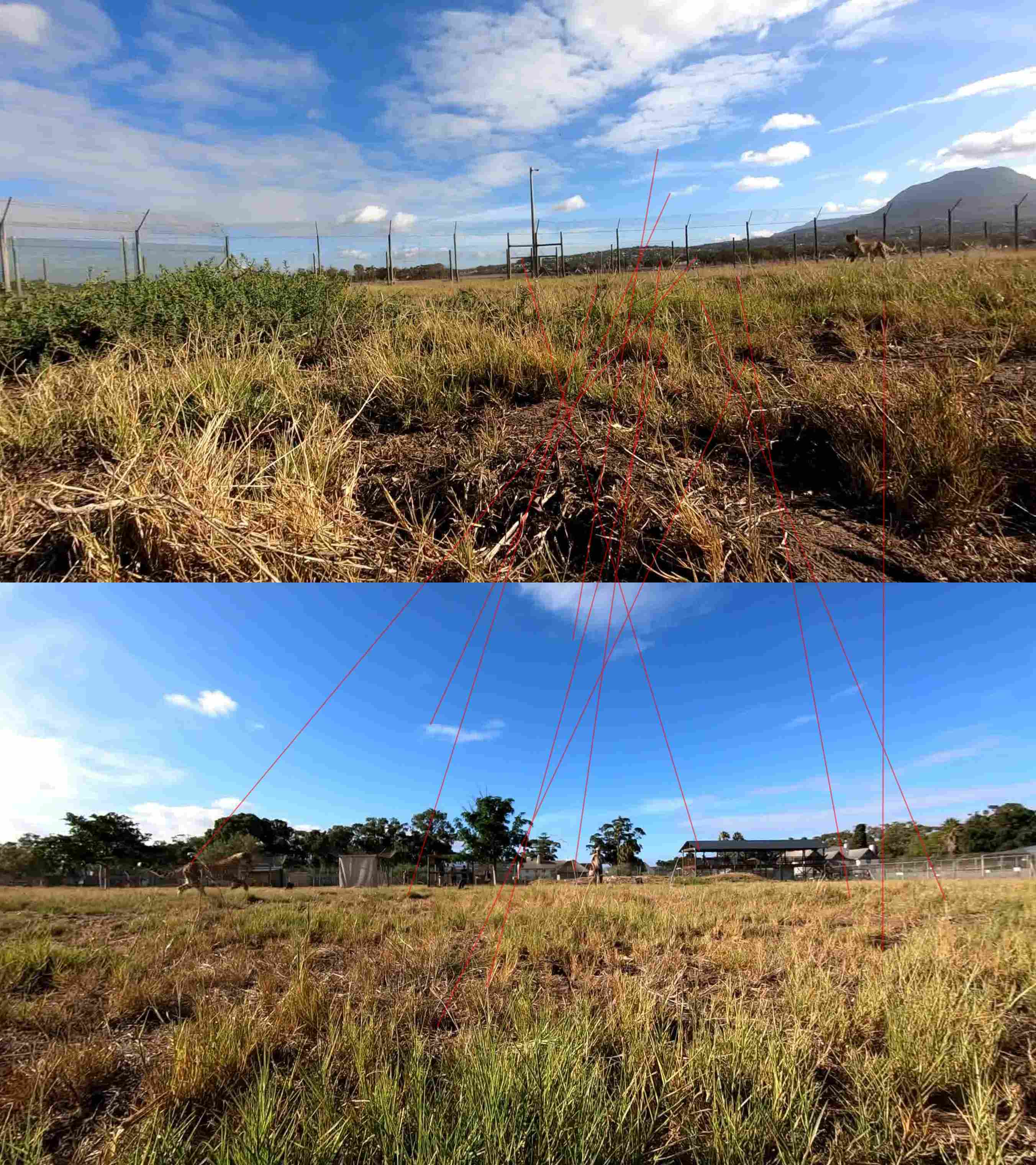}{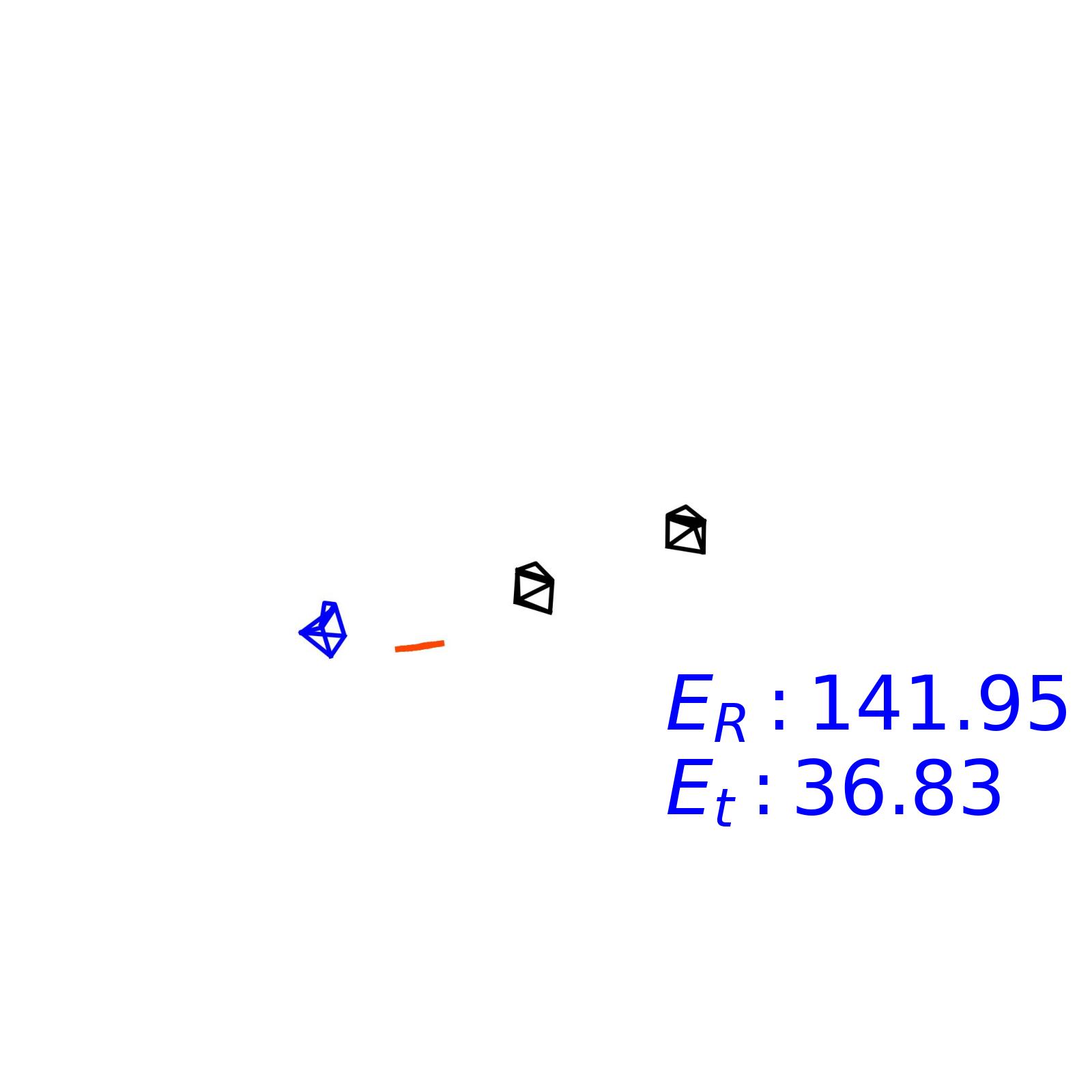} &        
    \incg{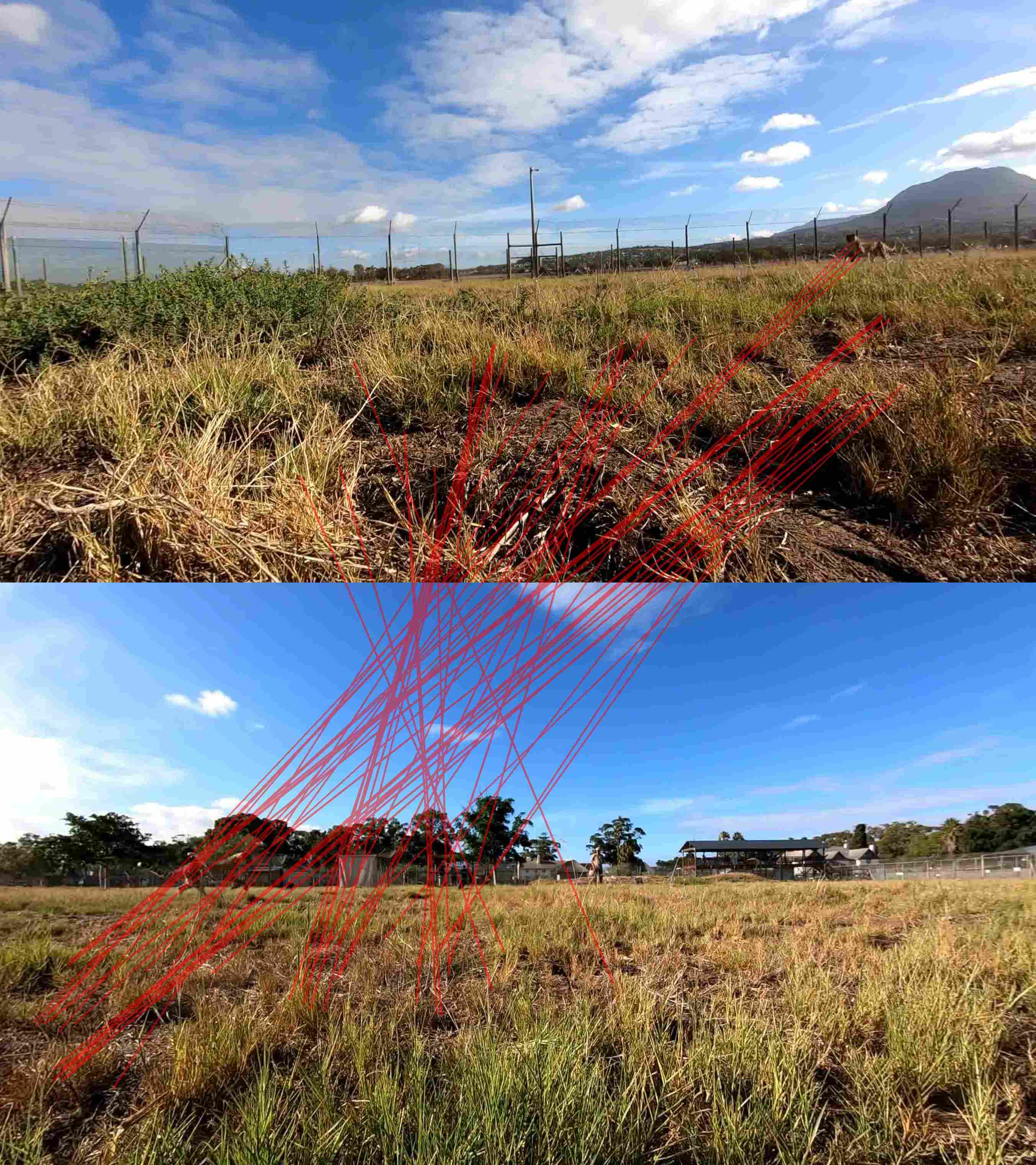}{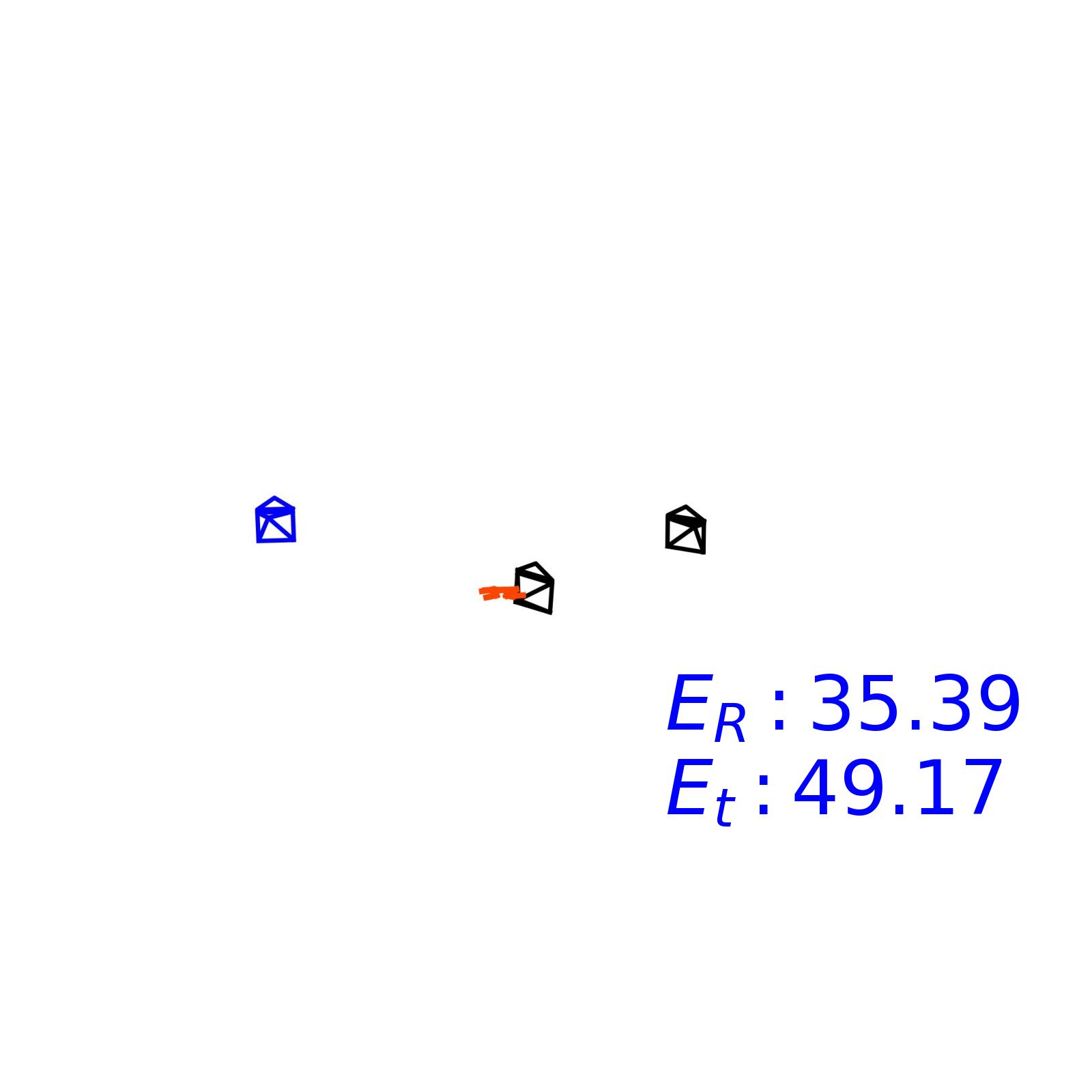}&
    \incg{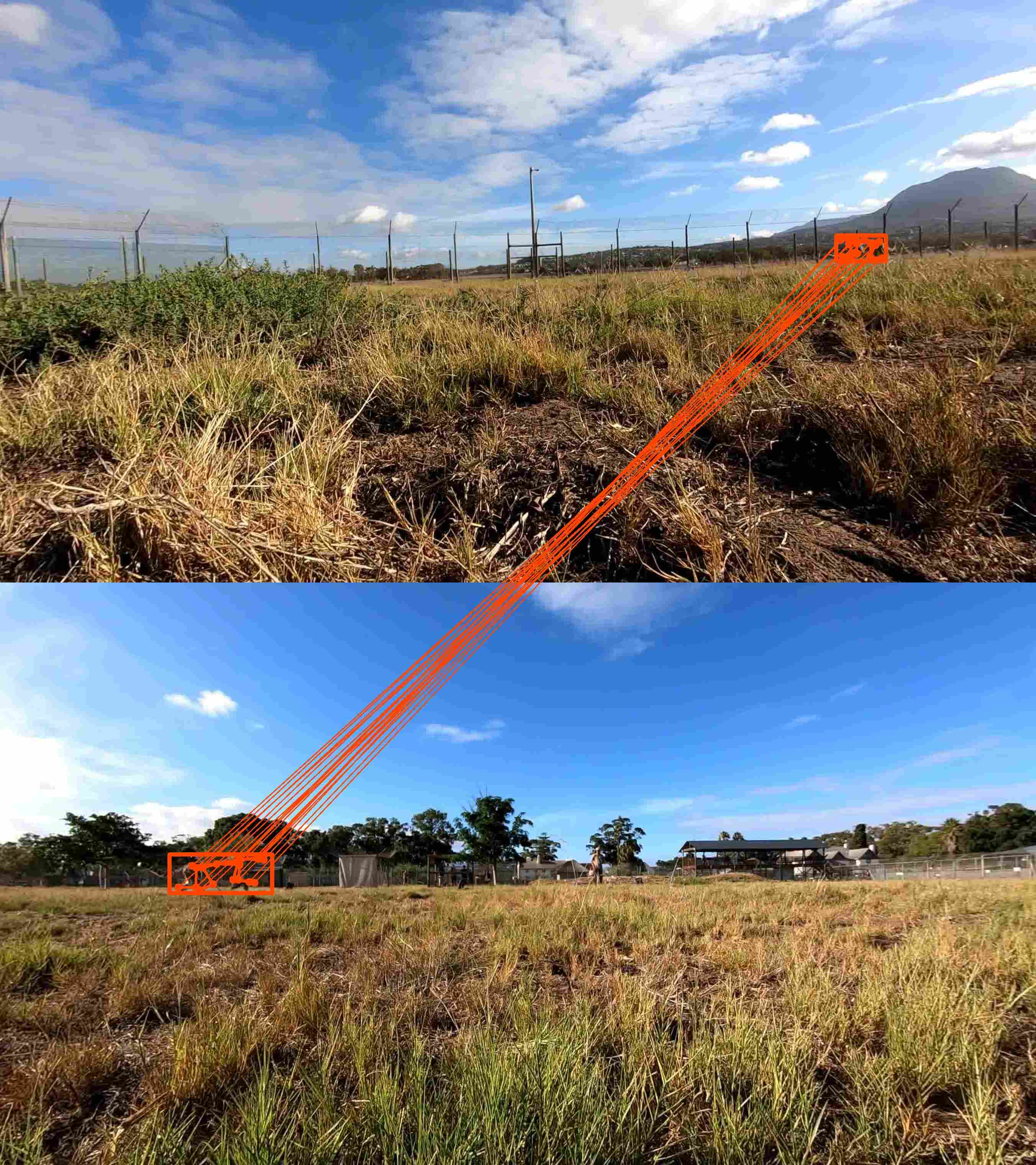}{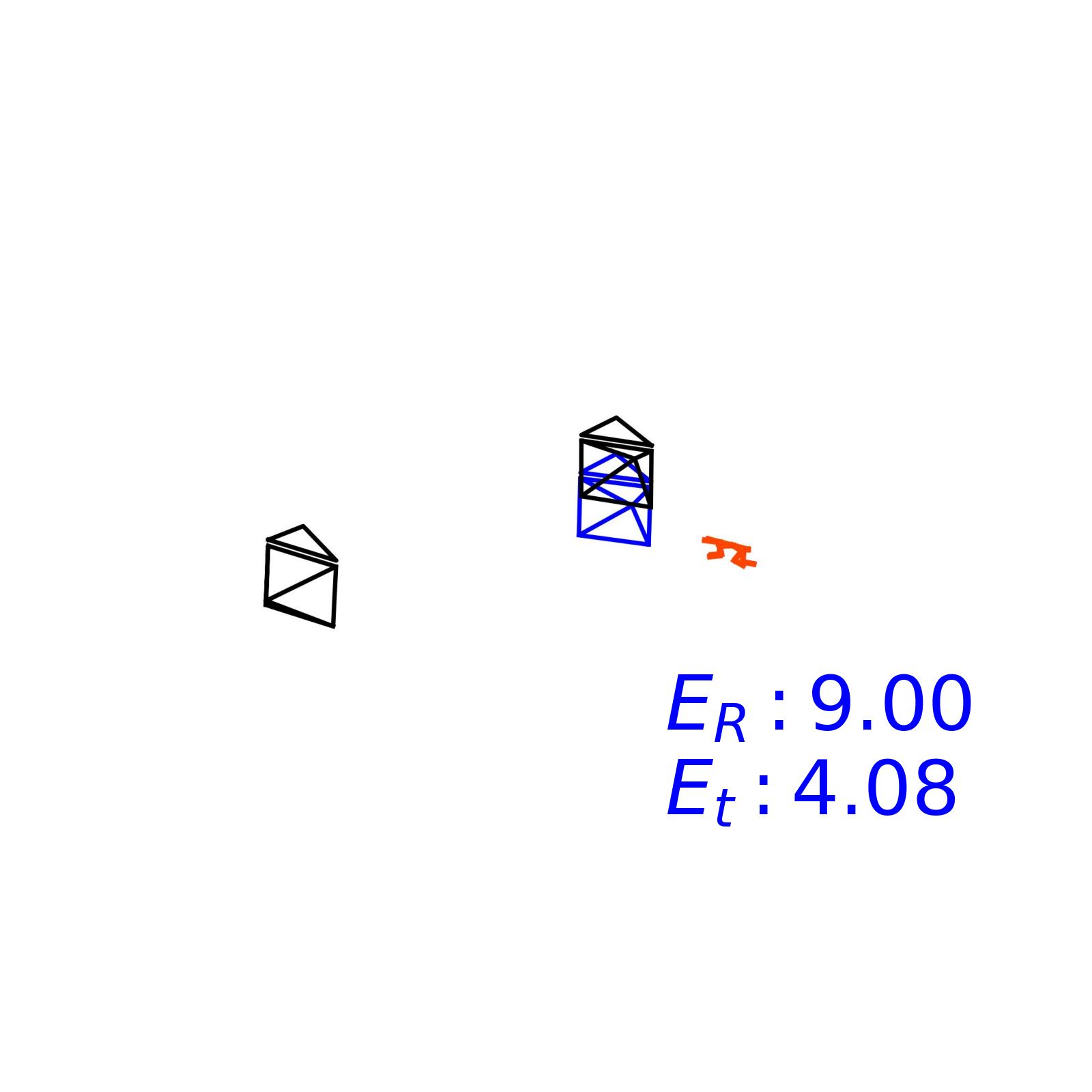} & 
    \raisebox{-0.5\totalheight}{%
    \includegraphics[width=0.13\linewidth]{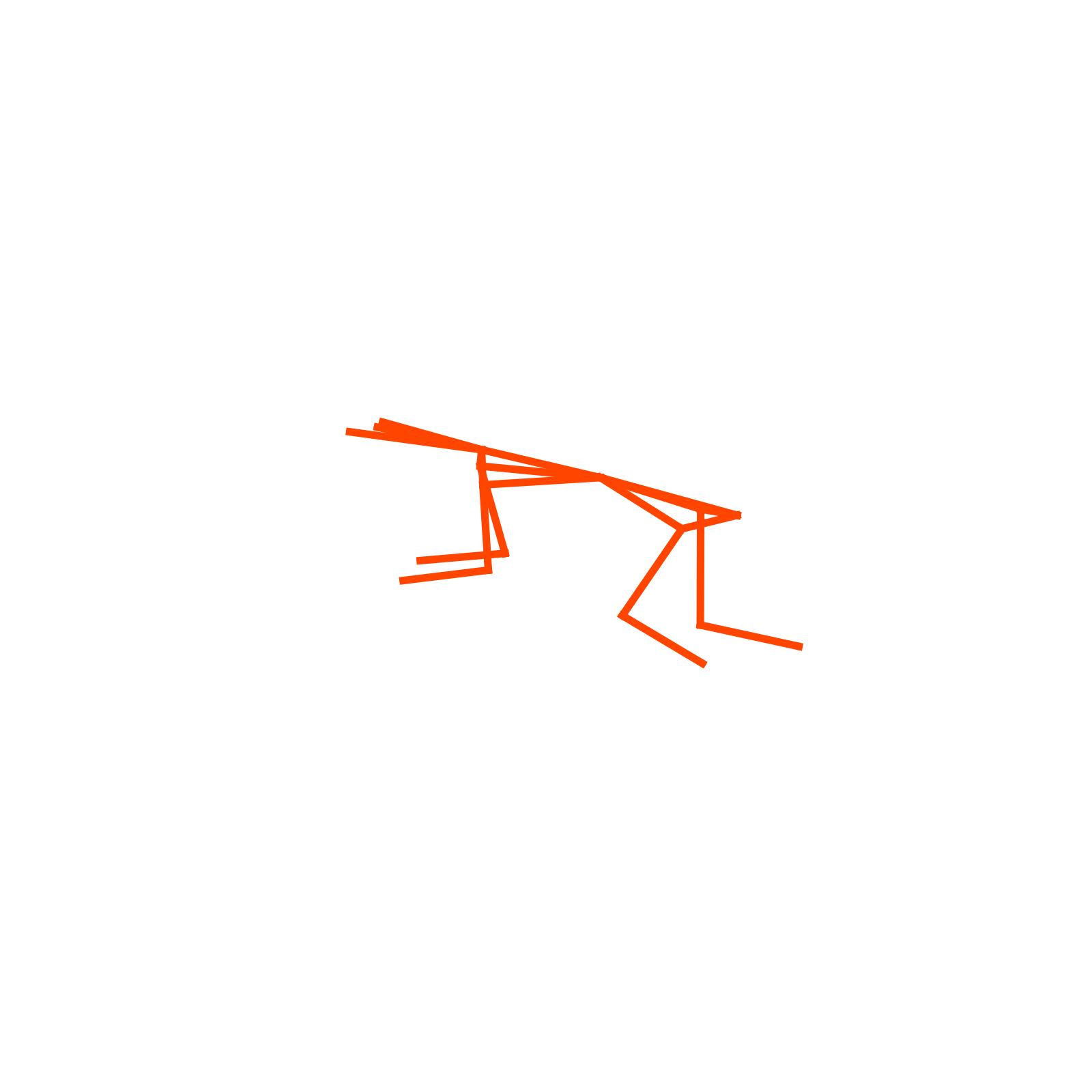}
  }%
    \\
    \hline
    \vlab{\Dpig} &
    \incg{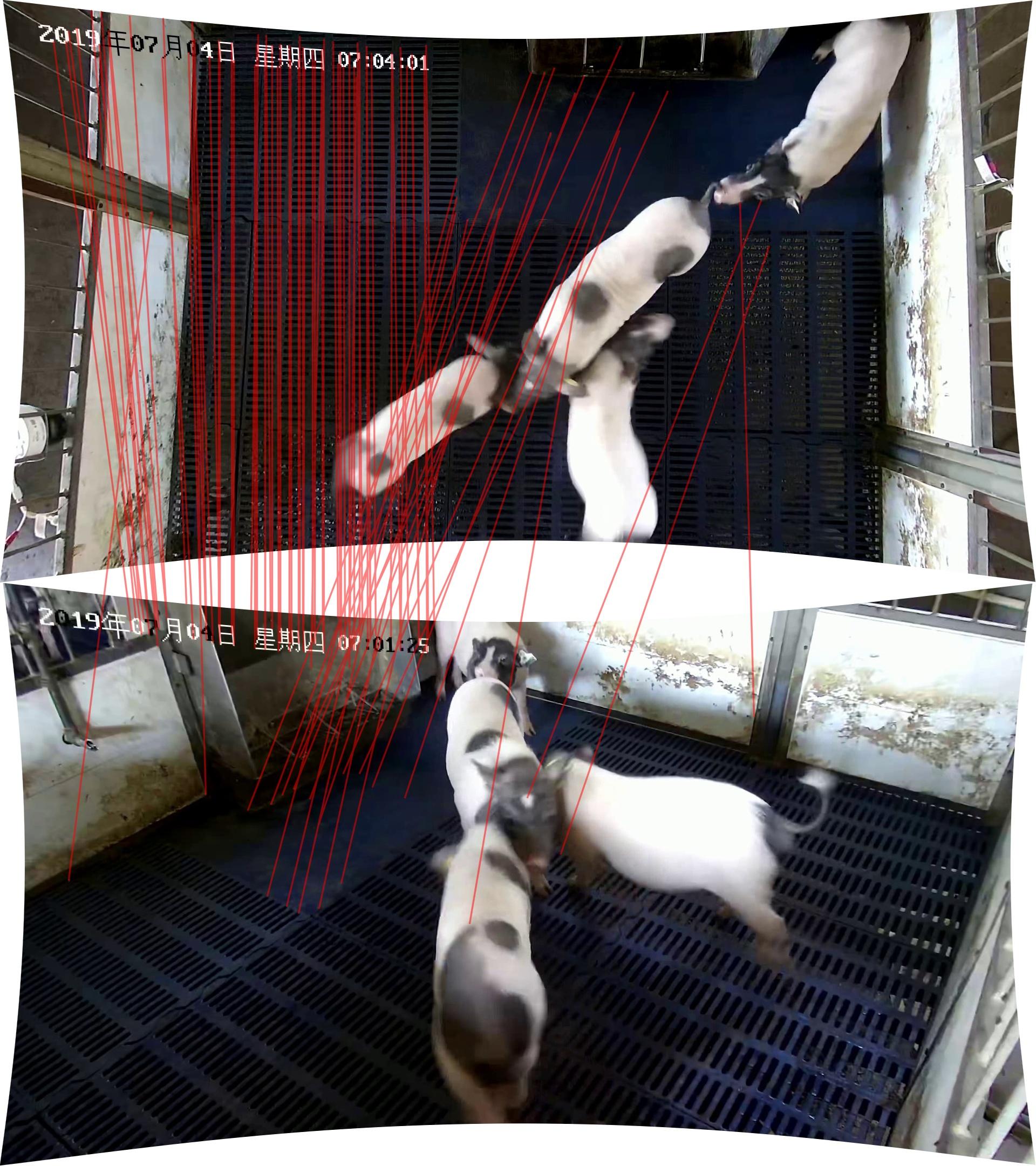}{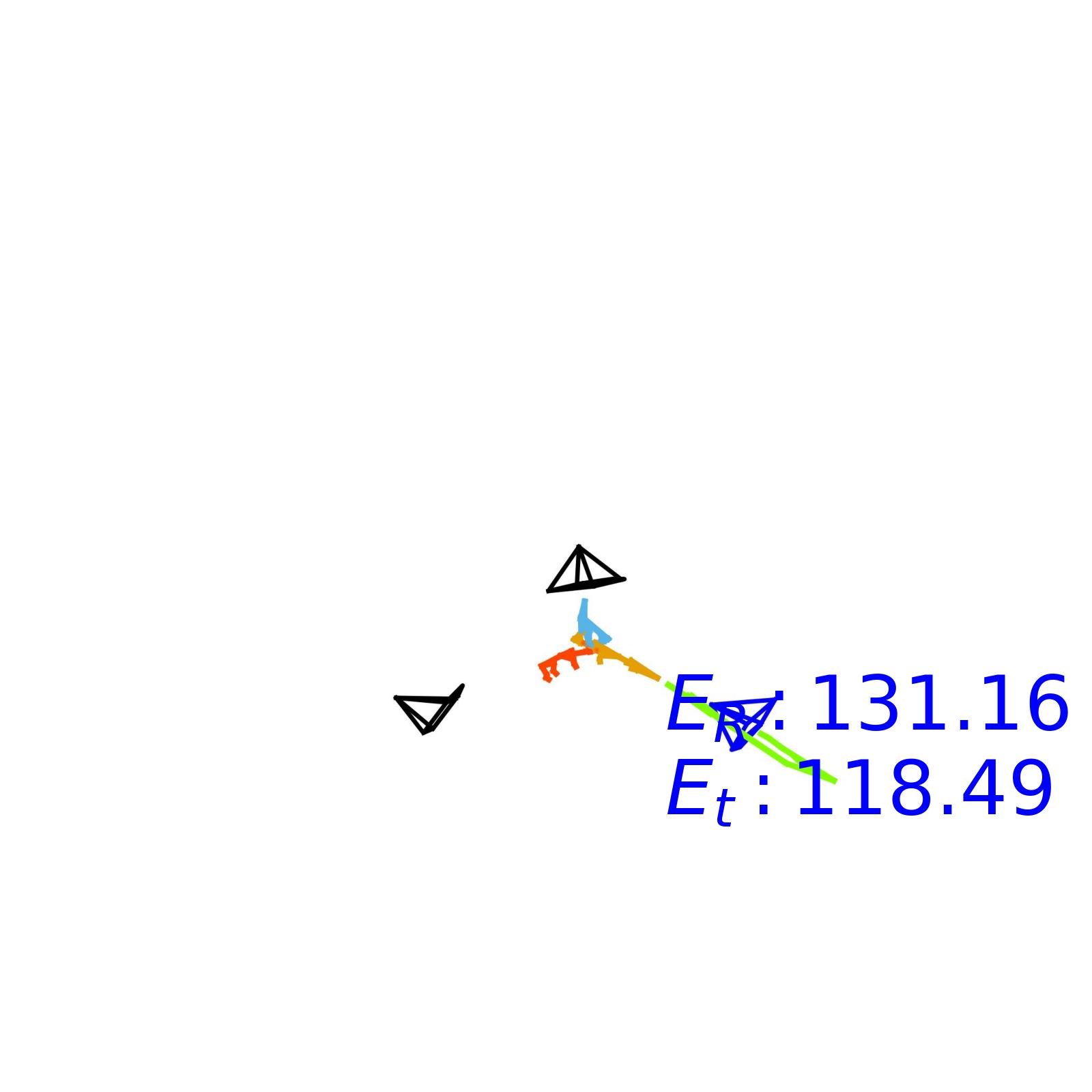} &        
    \incg{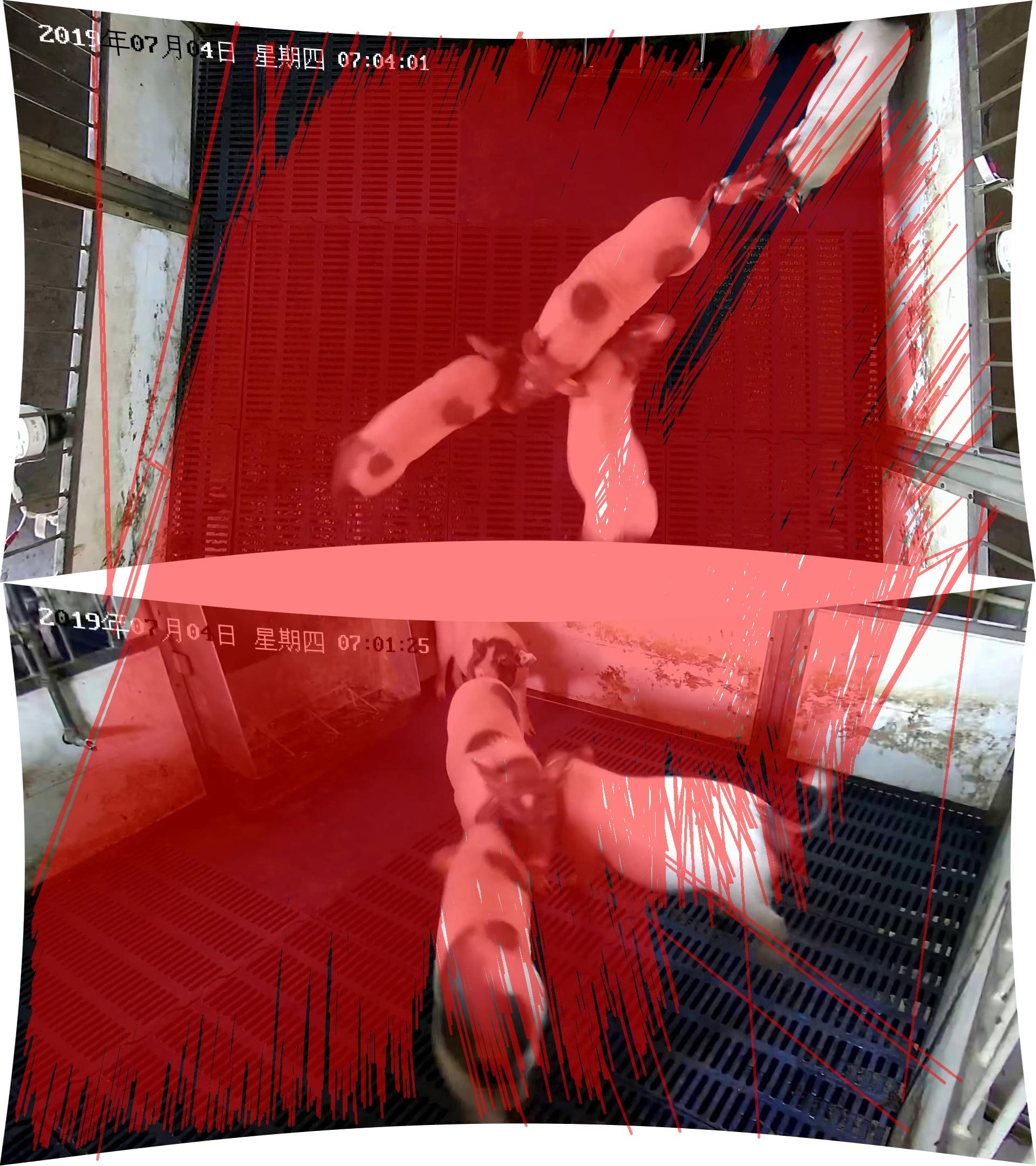}{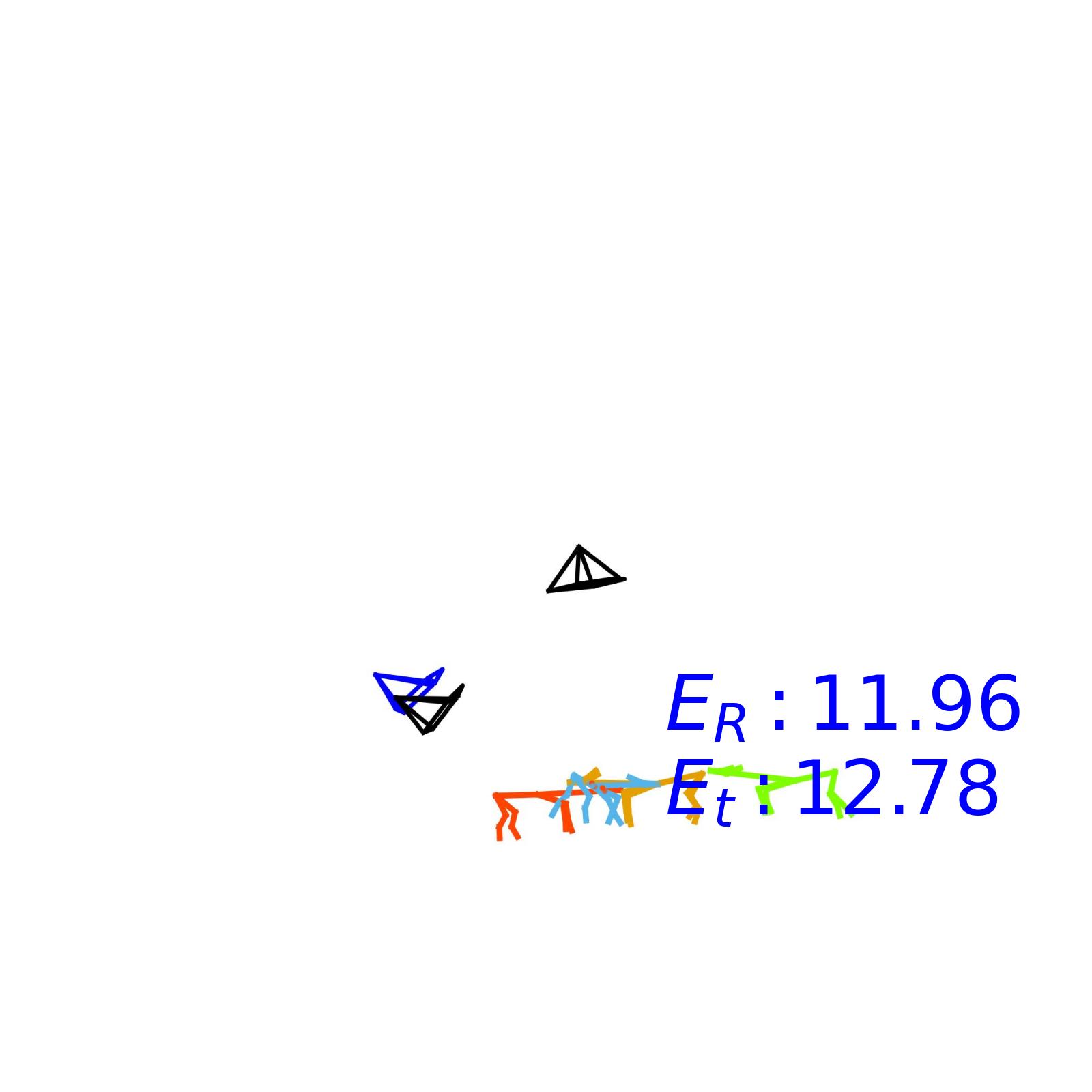}&
    \incg{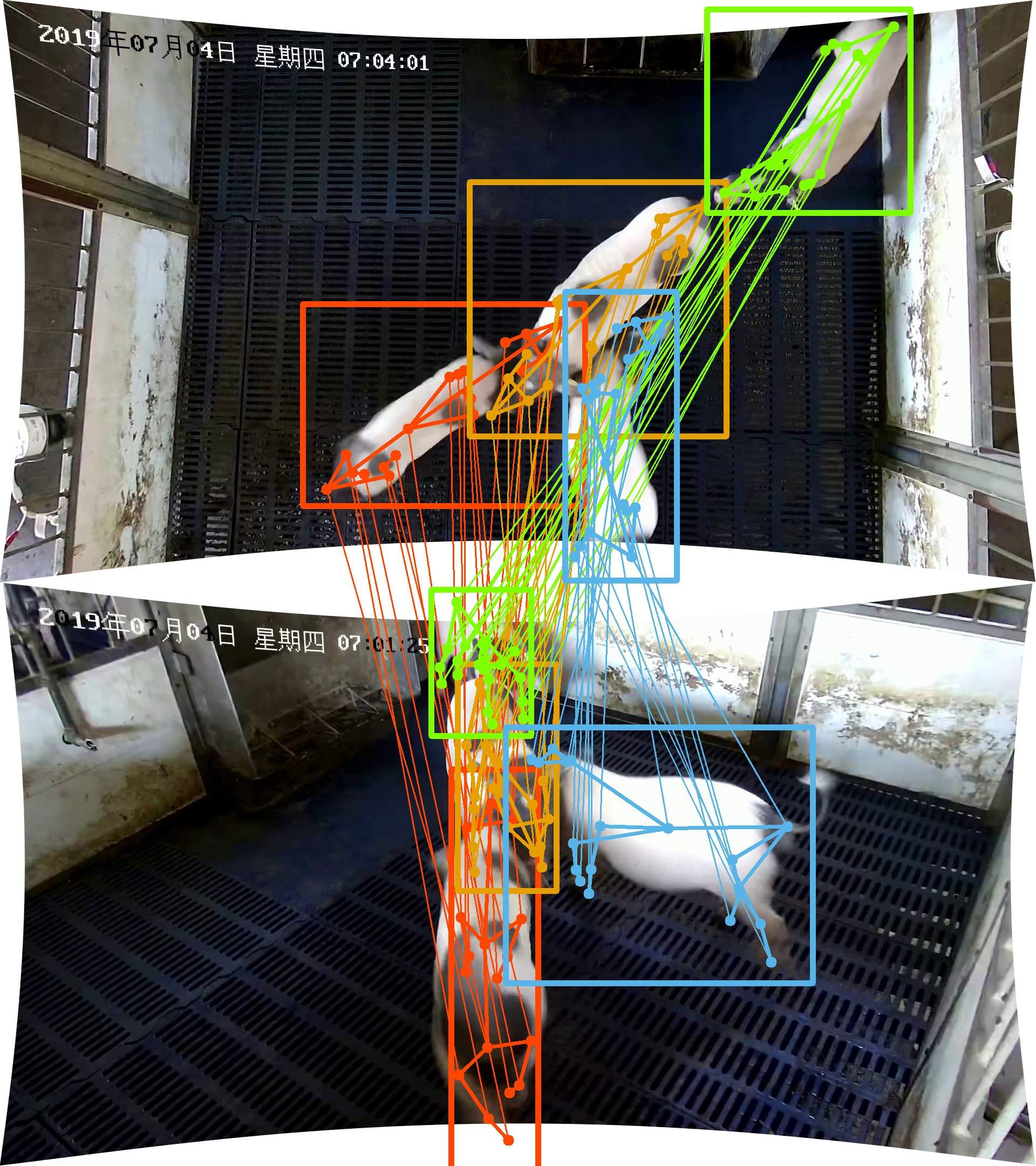}{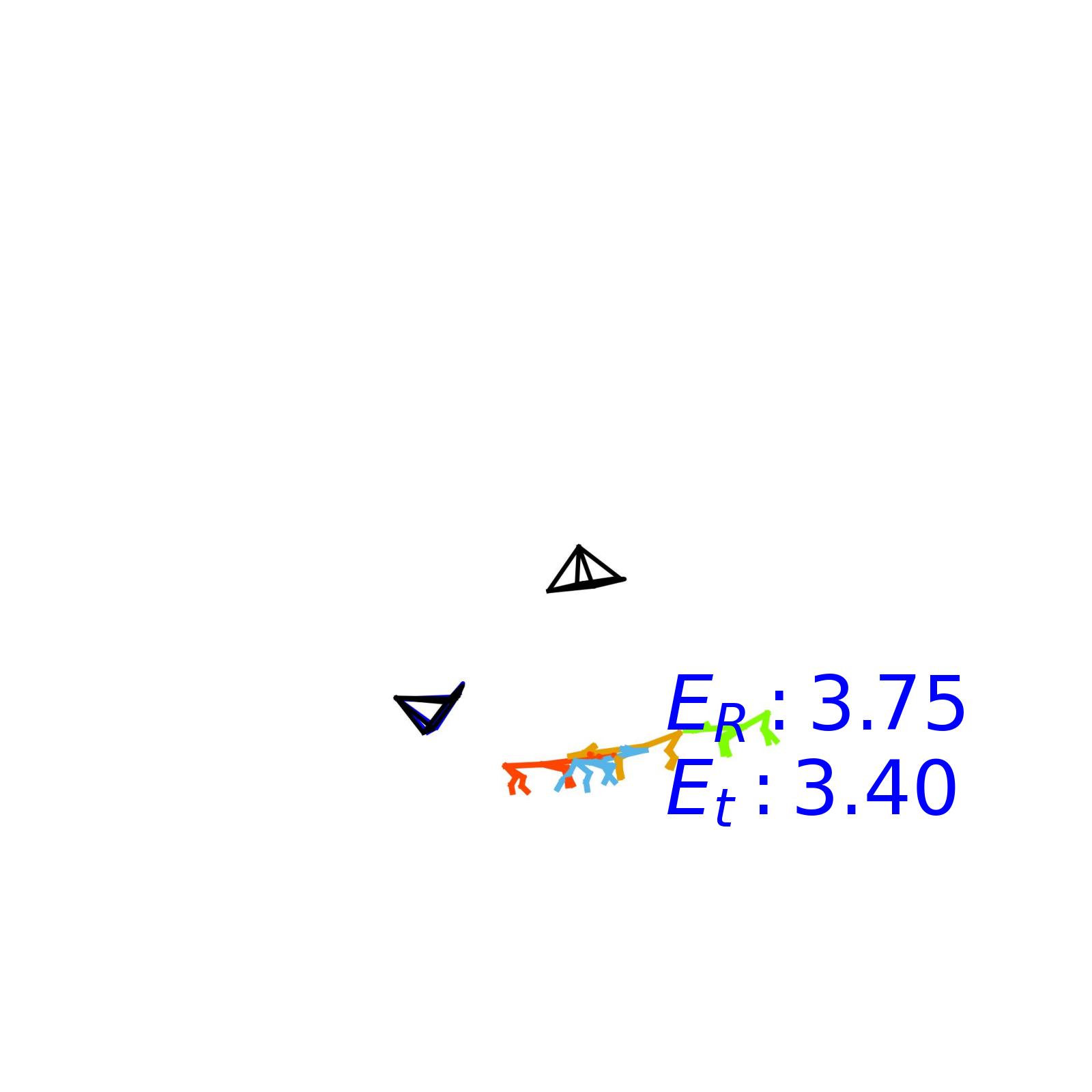} & 
    \raisebox{-0.5\totalheight}{%
    \includegraphics[width=0.13\linewidth]{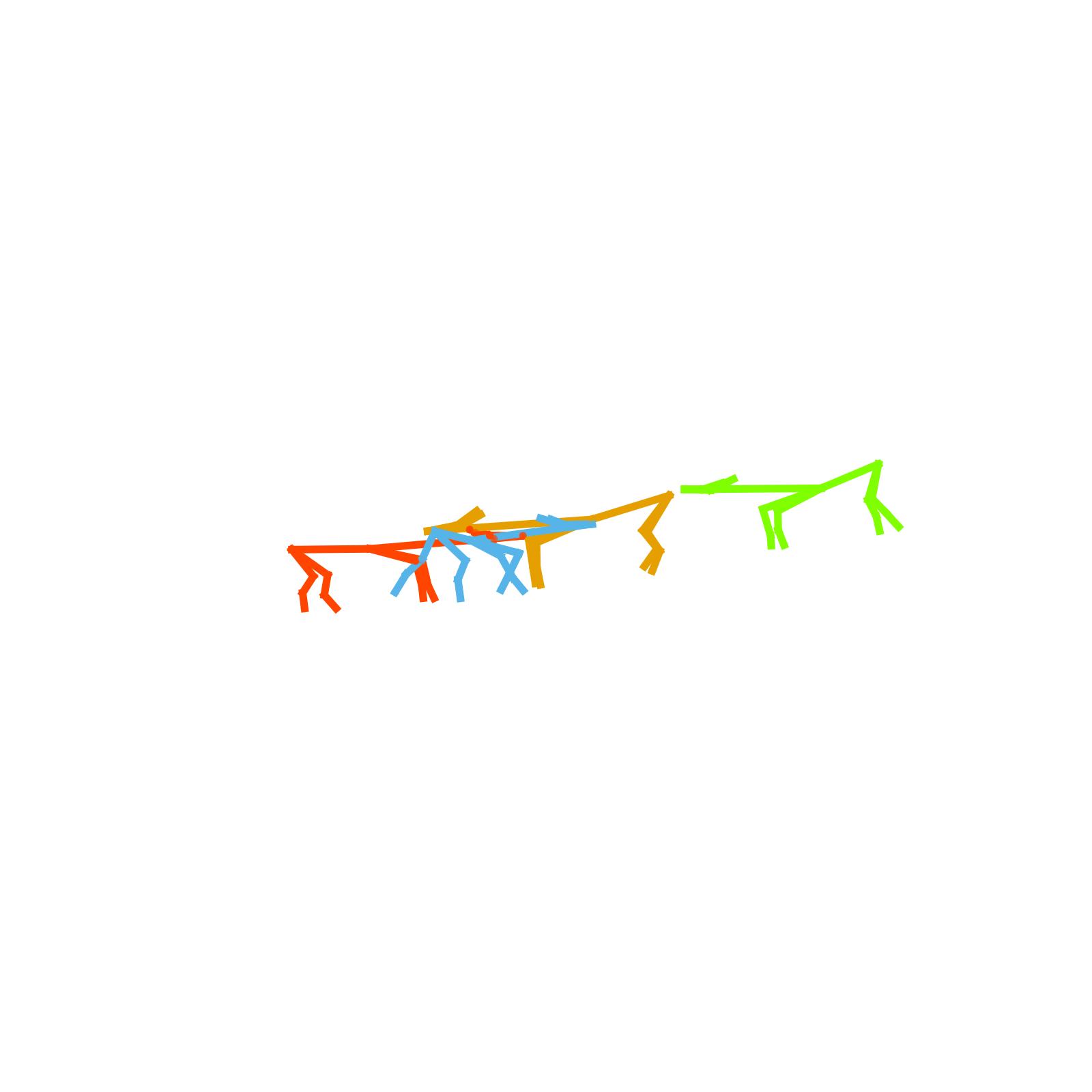}
  }%
    \\
    \hline
    \vlab{\Dpigeon} &
    \incg{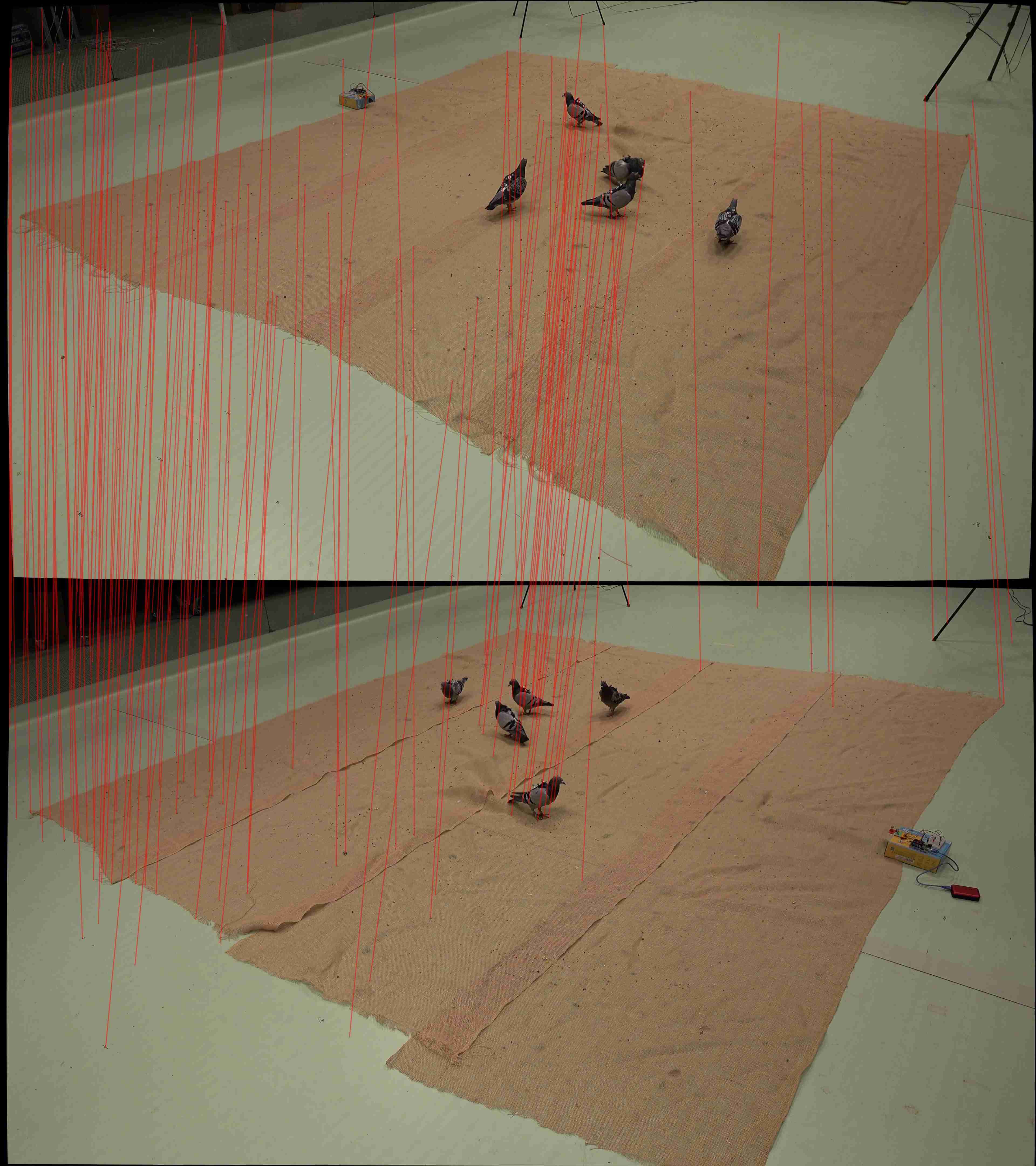}{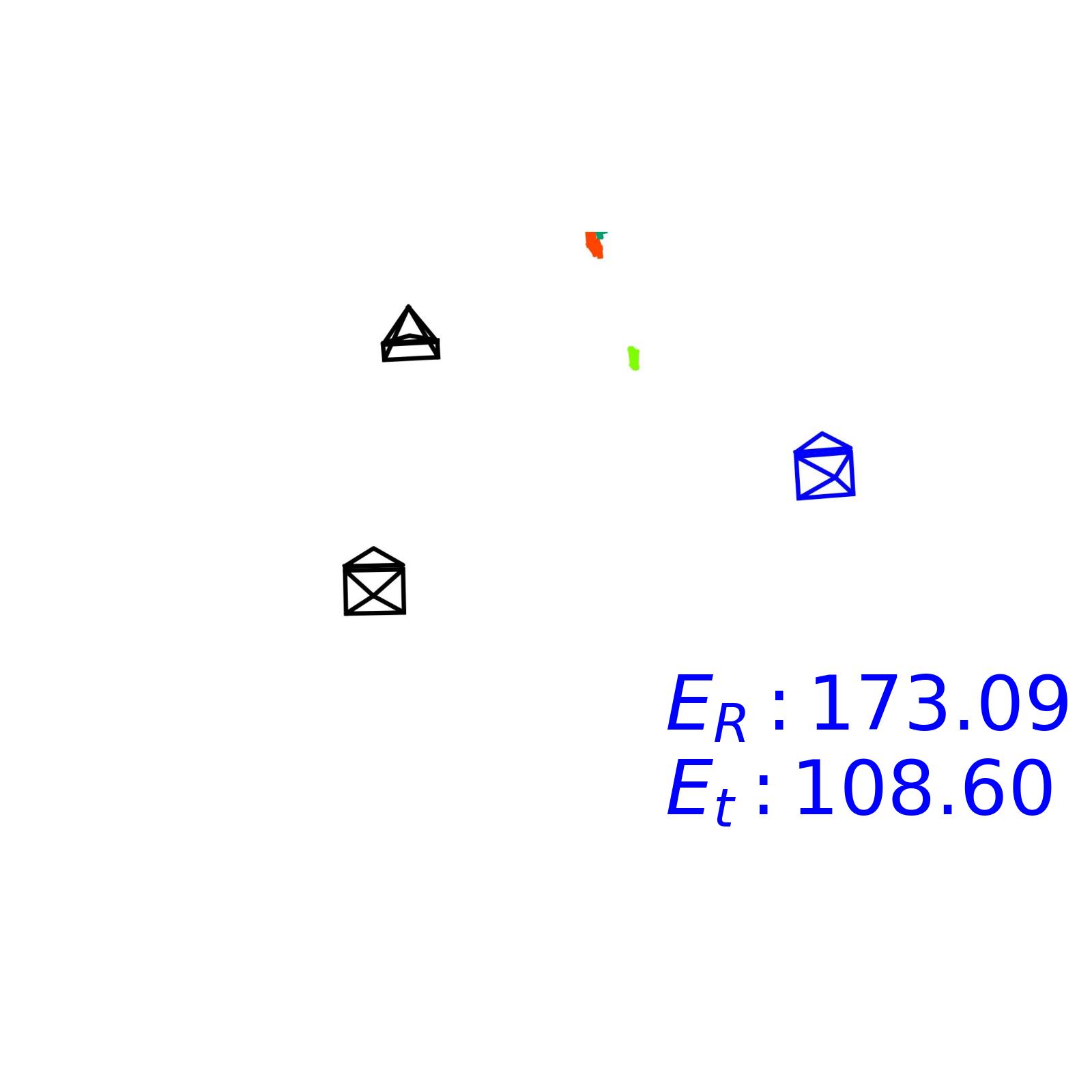} &        
    \incg{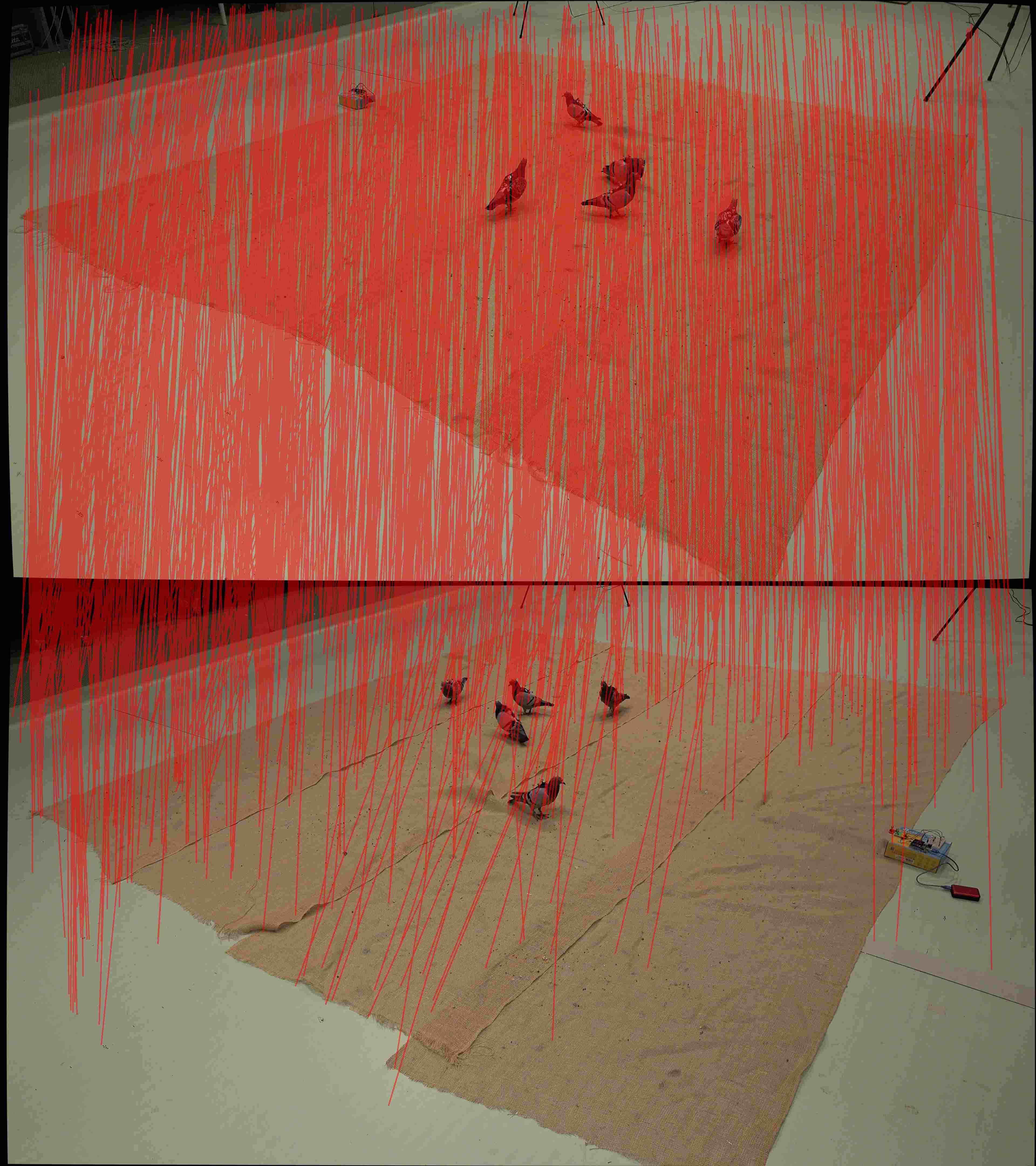}{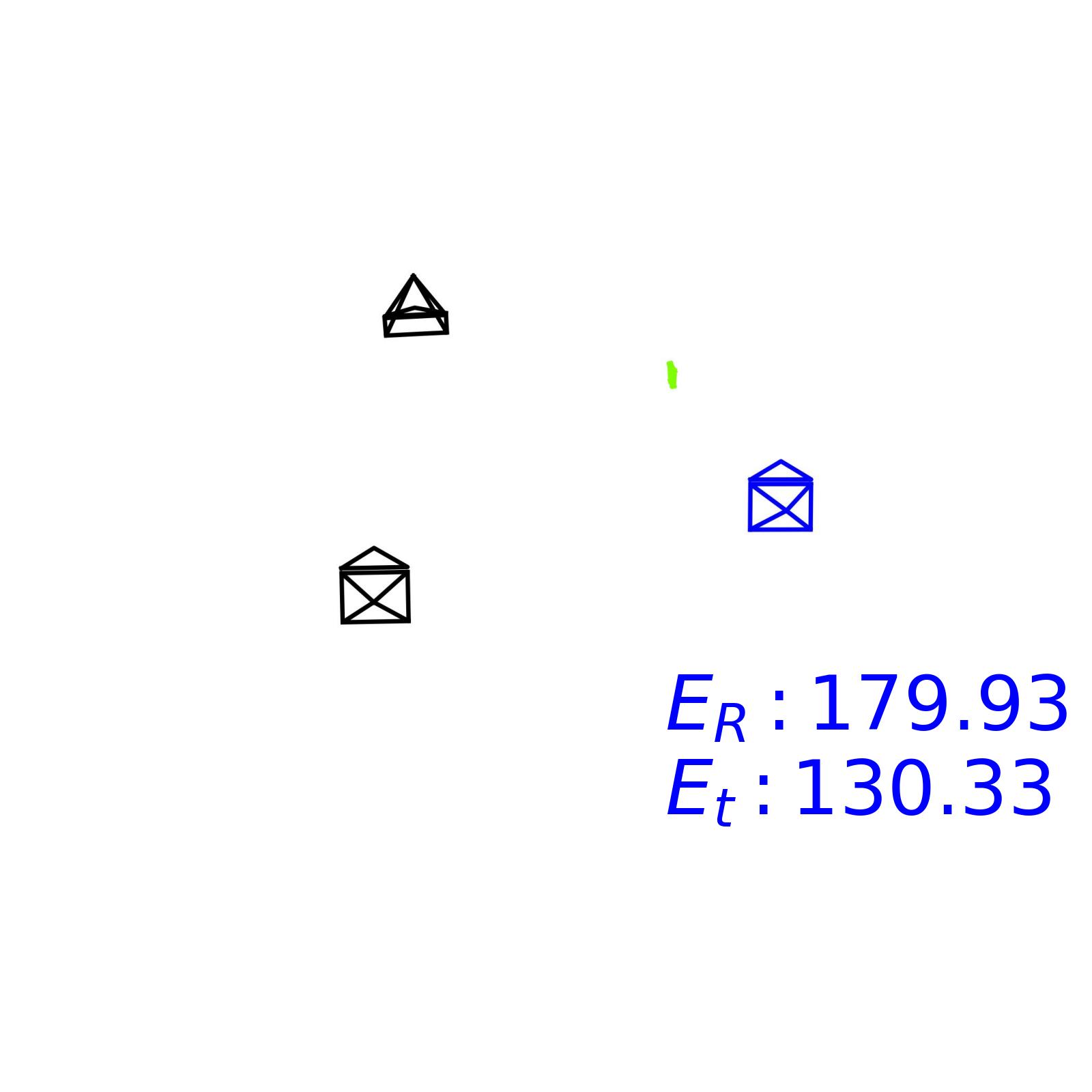}&
    \incg{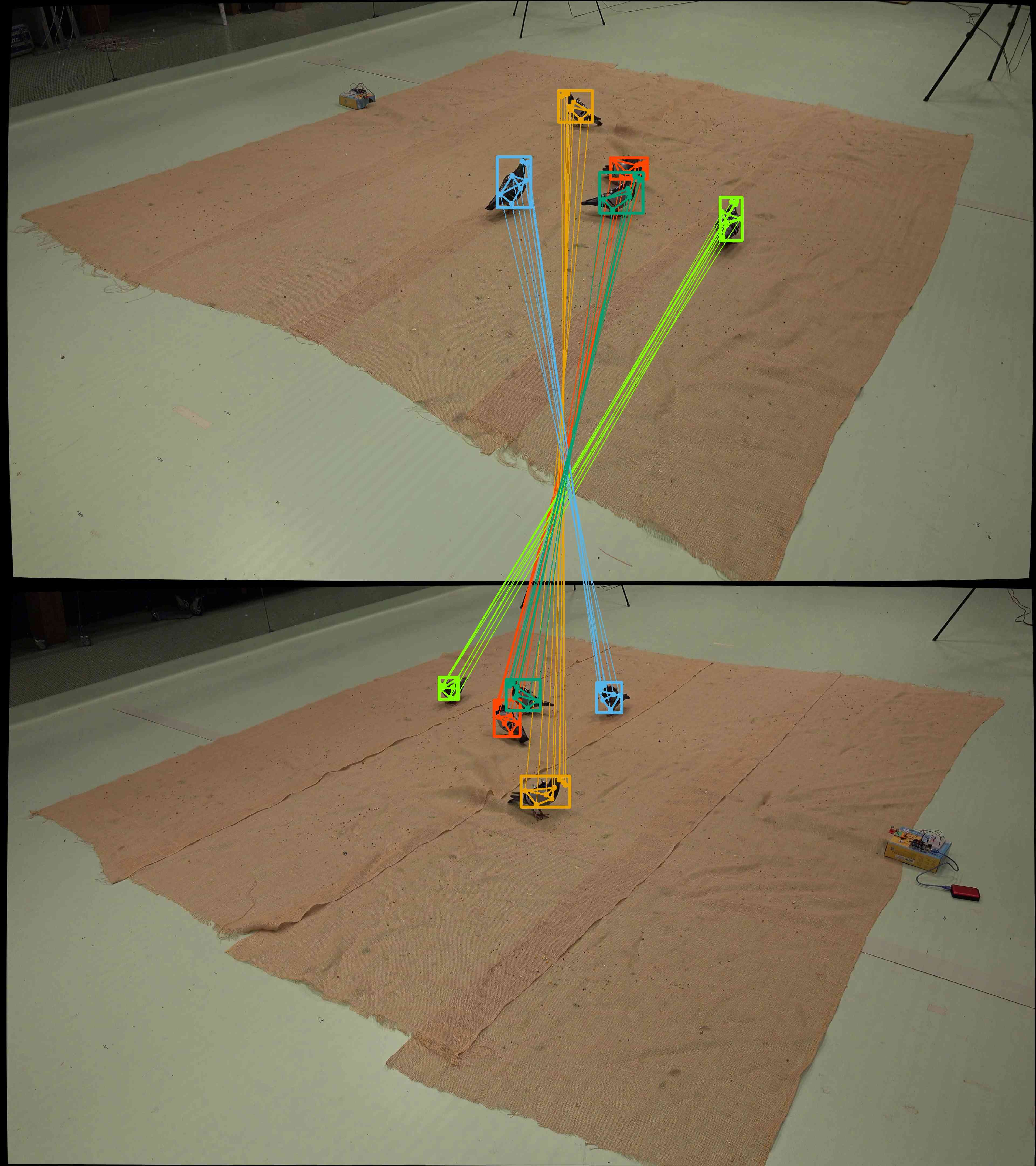}{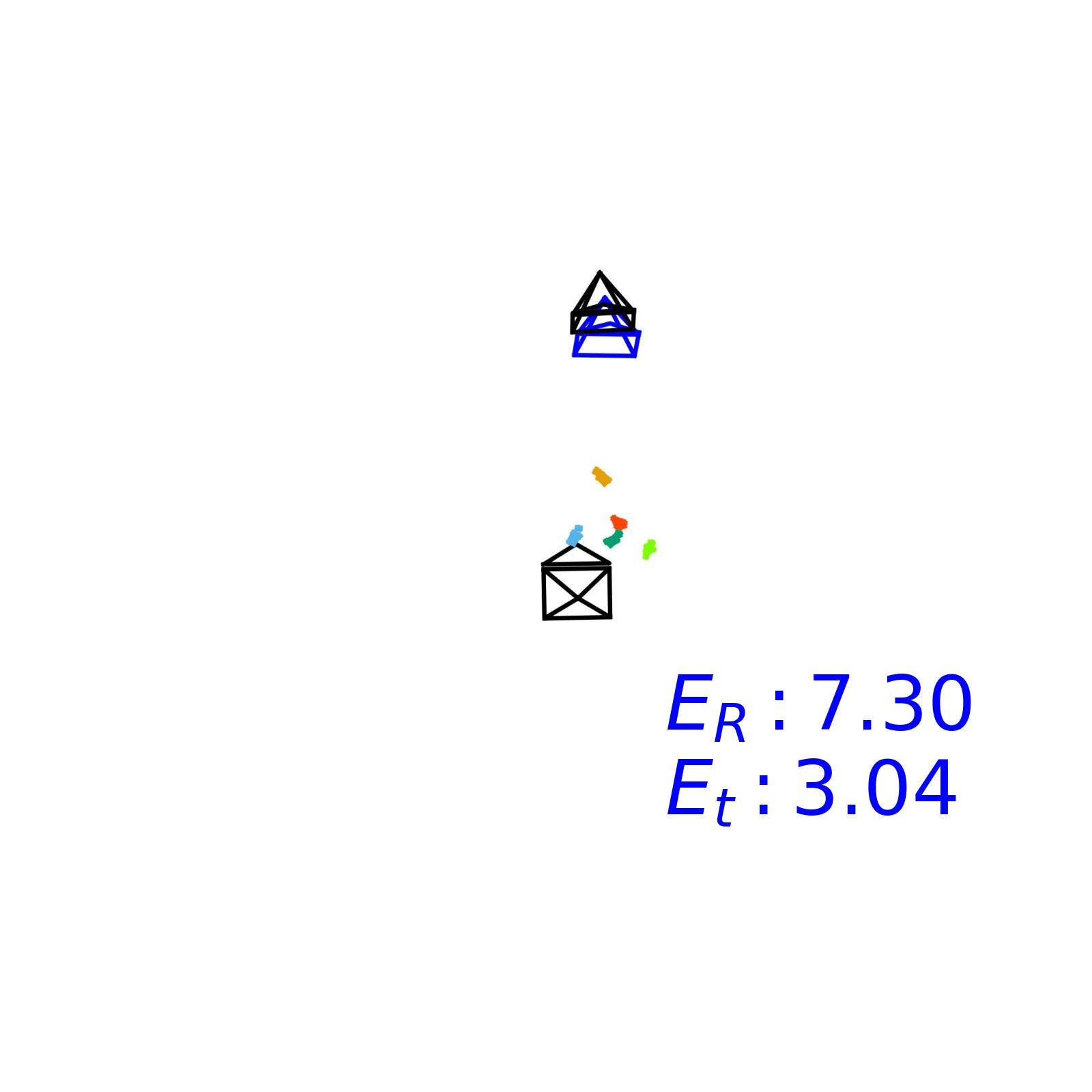} & 
    \raisebox{-0.5\totalheight}{%
    \includegraphics[width=0.13\linewidth]{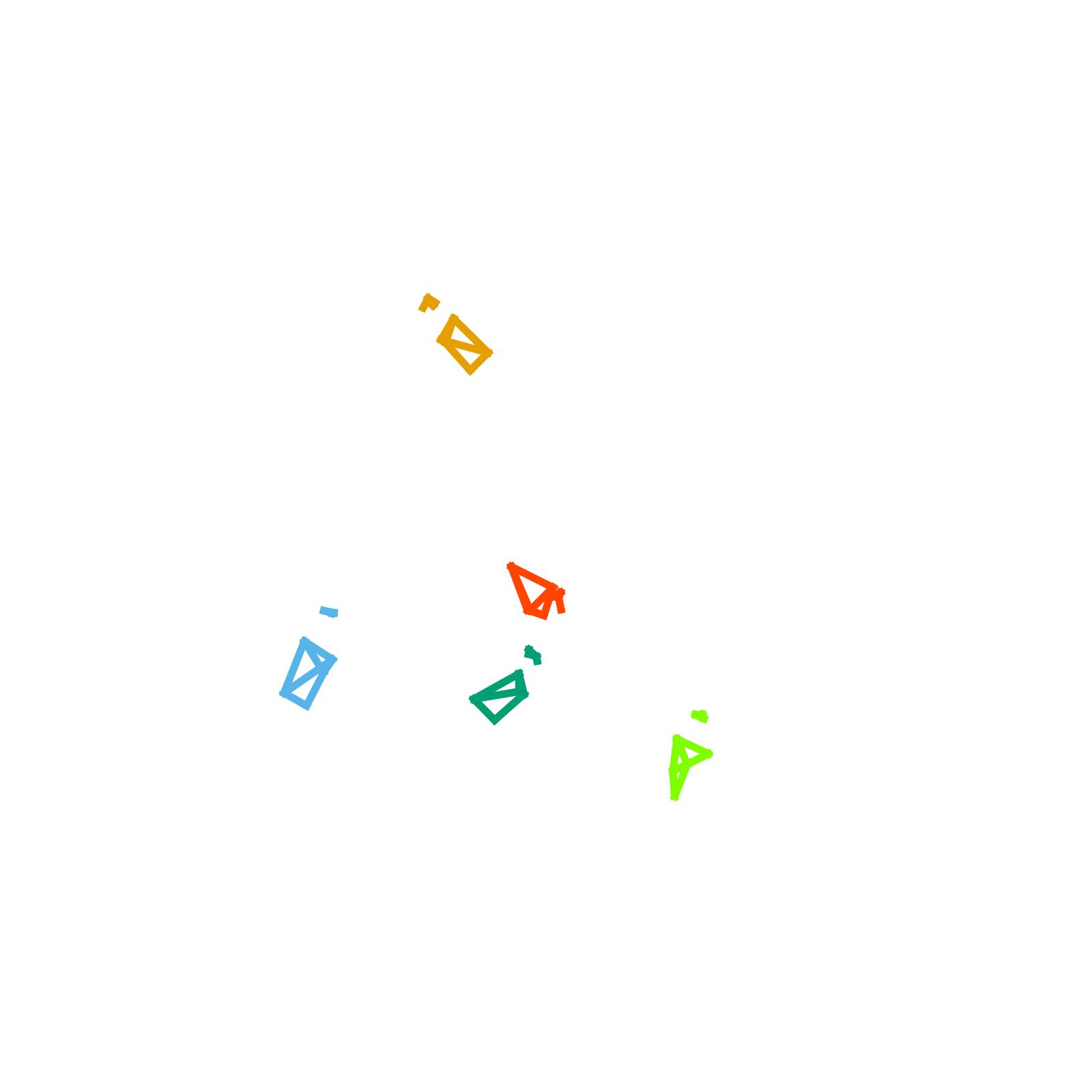}
  }%
    \\
    \hline

    \end{tabular}
    \caption{Two-view extrinsic calibration and matching results. 
    The ground-truth and estimated camera poses are shown in red and green, respectively. 
    Our method reliably establishes correspondences and accurately estimates camera poses for 3D reconstruction.}
    \label{fig:two_view_calib}
\end{figure}

We evaluate on five animal datasets:  \Dcheetah from Acinoset~\cite{joska2021acinoset}, \Dpig and \Ddog from MAMMAL~\cite{an2023three}, and \Dpigeon from 3D-POP~\cite{naik20233d}. For human subjects, we include \Dtoddler from CMU Panoptic dataset~\cite{Joo_2017_TPAMI} and \Dvolleyball from EgoHumans dataset~\cite{khirodkar2023egohumans}. Except for \Dcheetah, all datasets contain multiple instances in the scene, as shown in Fig.~\ref{fig:two_view_calib}. 
To evaluate our method, we employed both synthetic and real-world datasets. For the former, we project ground-truth 3D poses from these datasets into multi-view images and inject Gaussian noise with a standard deviation of 3 pixels to generate noisy 2D keypoints. For the latter, we utilize the 2D poses obtained from off-the-shelf pose estimators~\cite{ye2024superanimal,lu2023rtmo} that are tailored to each animal class. 

\paragraph{Metrics}

We assess the camera pose error using rotation error $E_R$ and translation error $E_{\mathbf{t}}$, both expressed as angular differences in degrees, and report the median projection error $E_{\mathrm{2D}}$ in pixels between the reprojected 3D keypoints and their corresponding 2D observations. For the matching evaluation, we measure the precision $P$ using the ground-truth instance correspondences across views. We further evaluate the precision of the estimated camera pose using three metrics: the relative rotation accuracy (RRA), relative translation accuracy (RTA), and area under curve (AUC)~\cite{dust3r_cvpr24}.  For a given threshold $\tau$, AUC@$\tau$ is computed as the minimum of RRA@$\tau$ and RTA@$\tau$.

\begin{table}[t]
    \centering
    \def\RRA{\scriptsize{RRA@30$\uparrow$}}
    \def\RTA{\scriptsize{RTA@30$\uparrow$}}
    \def\AUC{\scriptsize{AUC@30$\uparrow$}}
    \resizebox{\linewidth}{!}{%
    \begin{tabular}{
    lc|ccc|ccc|ccc}
    \toprule
    \multirow{2}{*}{Target}
    & \multirow{2}{*}{\shortstack[c]{\# of\\ subjects}}
    & \multicolumn{3}{c|}{5-point~\cite{nister2004efficient}}
    & \multicolumn{3}{c|}{Class-specific SteerPose}
    & \multicolumn{3}{c}{Class-agnostic SteerPose}
    \\
     &  
     & \RRA & \RTA & \AUC
     & \RRA & \RTA & \AUC
     & \RRA & \RTA & \AUC
     \\
    \hline
\Dcheetah~\cite{joska2021acinoset} & 1 & 0.5 & 0.61 & 0.43 & 0.82 & 0.86 & 0.58 & 0.81 & 0.85 & 0.57 \\
\Dpig~\cite{an2023three} & 4 & 0.64 & 0.77 & 0.56 & 0.88 & 0.89 & 0.76 & 0.92 & 0.97 & 0.81 \\
\Ddog~\cite{an2023three} & 2 & 0.47 & 0.61 & 0.42 & 0.96 & 0.98 & 0.79 & 0.96 & 0.96 & 0.79 \\
\Dpigeon~\cite{naik20233d} & 5 & 0.77 & 0.81 & 0.58 & 0.94 & 0.94 & 0.75 & N/A & N/A & N/A \\
\Dtoddler~\cite{Joo_2017_TPAMI} & 4 & 0.4 & 0.54 & 0.35 & 0.68 & 0.76 & 0.52 & N/A & N/A & N/A \\
    \bottomrule
    \end{tabular}
    }
    \caption{Comparison to a conventional geometry-based pose estimation method. Both the class-specific and class-agnostic models outperform the baseline method. }

    \label{tab:2view_geometric}
\end{table}

\begin{table}[t]
    \centering
    \def\RRA{\scriptsize{RRA@20$\uparrow$}}
    \def\RTA{\scriptsize{RTA@20$\uparrow$}}
    \def\AUC{\scriptsize{AUC@20$\uparrow$}}
    \resizebox{\textwidth}{!}{%

    \begin{tabular}{lc|ccc|ccc|ccc}
    \toprule
    \multirow{2}{*}{Target}&\multirow{2}{*}{\# of Subject}
    & \multicolumn{3}{c}{LightGlue~\cite{lindenberger2023lightglue}}
    & \multicolumn{3}{c}{\DMaster~\cite{mast3r_arxiv24}}
    & \multicolumn{3}{c}{Ours} \\
    
    &
    & \RRA & \RTA & \AUC
     & \RRA & \RTA & \AUC
     & \RRA & \RTA & \AUC
    \\ 
        \toprule
        \rowcolor{gray!30}
    \multirow{2}{*}{\cellcolor{white}{\Dcheetah~\cite{joska2021acinoset}}} & \multirow{2}{*}{\cellcolor{white}{1}}
    & 0.30 & 0.25 & 0.30 & 0.44 & 0.59 & 0.42 & 0.62 & 0.89 & 0.48 \\
    &
    & 0.30 & 0.24 & 0.30 & 0.44 & 0.58 & 0.42 & 0.41 & 0.55 & 0.26 \\
    \midrule
    \rowcolor{gray!30}
    \multirow{2}{*}{\cellcolor{white}{\Dpig~\cite{an2023three}}} & \multirow{2}{*}{\cellcolor{white}{4}}
    & 0.24 & 0.23 & 0.22 & 0.96 & 0.96 & 0.88 & 0.99 & 0.99 & 0.85 \\
    &
    & 0.38 & 0.38 & 0.35 & 0.89 & 0.90 & 0.81 & 0.77 & 0.73 & 0.45 \\
    \midrule
    \rowcolor{gray!30}
    \multirow{2}{*}{\cellcolor{white}{\Ddog~\cite{an2023three}}} & \multirow{2}{*}{\cellcolor{white}{2}}
    & 0.31 & 0.38 & 0.30 & 1.00 & 1.00 & 0.92 & 0.85 & 0.97 & 0.77 \\
    &
    & 0.30 & 0.41 & 0.29 & 1.00 & 1.00 & 0.94 & 0.76 & 0.76 & 0.47 \\
    \midrule
    \rowcolor{gray!30}
    \multirow{1}{*}{\cellcolor{white}{\Dpigeon~\cite{naik20233d}}} & \multirow{1}{*}{\cellcolor{white}{5}}
    & 0.02 & 0.13 & 0.07 & 0.78 & 0.78 & 0.64 & 0.98 & 1.00 & 0.75 \\
    \midrule
    \rowcolor{gray!30}
    \multirow{2}{*}{\cellcolor{white}{\Dtoddler~\cite{Joo_2017_TPAMI}}} & \multirow{2}{*}{\cellcolor{white}{4}}
    & 0.86 & 0.78 & 0.78 & 0.99 & 0.99 & 0.89 & 0.73 & 0.75 & 0.66 \\
    &
    & 0.83 & 0.77 & 0.76 & 1.00 & 1.00 & 0.90 & 0.75 & 0.75 & 0.66 \\
    \midrule
    \rowcolor{gray!30}
    \multirow{2}{*}{\cellcolor{white}{\Dvolleyball~\cite{khirodkar2023egohumans}}} & \multirow{2}{*}{\cellcolor{white}{4}}
    & 0.64 & 0.88 & 0.78 & 1.00 & 1.00 & 0.97 & 0.94 & 0.95 & 0.91 \\
    &
    & 0.64 & 0.88 & 0.78 & 1.00 & 1.00 & 0.97 & 0.77 & 0.86 & 0.72 \\
    \bottomrule
    \end{tabular}
    }
    \caption{Evaluation of two-view camera pose estimation. Gray-shaded rows use the ground-truth 2D poses with Gaussian noise with $\sigma=3$px, while white rows utilize 2D poses from off-the-shelf estimators. Our method achieves a performance comparable to that of SOTA methods. }

    \label{tab:two_view_calib}
\end{table}

\subsection{Comparison with Existing Methods}\label{sec:eval_ransac}

\paragraph{Geometric Method}
Table~\ref{tab:2view_geometric} presents a comparison of SteerPose with the standard five-point method~\cite{nister2004efficient}, which is a point-correspondence-based approach. We employ the RANSAC algorithm to define correspondences for the five-point method in multi-instance scenarios. This evaluation includes a diverse set of five species, including birds and humans. For quadruped species, we compared the results of class-specific and class-agnostic models.

The five-point algorithm struggles with pose estimation failures in challenging scenarios, particularly in degenerate configurations where the calibration targets are coplanar or when dealing with wide-baseline conditions.  In contrast, SteerPose demonstrates a robust performance across these scenarios and effectively handles species with varying skeletal structures beyond quadrupeds. Both class-specific and class-agnostic SteerPose outperform the standard correspondence-based methods, demonstrating their versatility and reliability in pose estimation and matching. For detailed information about the class-agnostic SteerPose, please see Appendix~\ref{sec:gen_steerpose}.

\paragraph{Learning-based Methods}

Table~\ref{tab:two_view_calib} and Fig.~\ref{fig:two_view_calib} compare our method with the state-of-the-art learning-based methods. SuperPoint~\cite{detone2018superpoint} and LightGlue~\cite{lindenberger2023lightglue} leverage learned priors for keypoint detection, description, and matching in 2D images. The relative poses are then estimated from these matches using RANSAC~\cite{fischler_bolles:RANSAC} and the essential matrix decomposition~\cite{Hartley00}. Then, a bundle adjustment~\cite{Hartley00} is employed to refine the poses. \DMaster~\cite{mast3r_arxiv24} simultaneously performs 3D scene reconstruction and matching based on monocular 3D point-map estimation~\cite{dust3r_cvpr24}.

The results show that SteerPose outperforms Lightglue, both quantitatively and qualitatively. Our method, which leverages structured keypoints obtained from articulated poses, achieves accuracy comparable to that of \DMaster. As shown in Fig~\ref{fig:two_view_calib}, our method is more robust in challenging scenarios, such as those with poor background textures or wide baselines between cameras (\eg, \Dcheetah and \Dpigeon), where \DMaster often struggles.

\begin{figure}[t]
    \centering
    \def\vlab#1{\rotatebox[origin=c]{90}{#1}}
    \newcommand{\incgthree}[3]{%
    \raisebox{-0.5\totalheight}{%
      \includegraphics[width=0.32\linewidth]{#1}%
    } & 
    \raisebox{-0.5\totalheight}{%
      \includegraphics[width=0.32\linewidth]{#2}%
    } & 
    \raisebox{-0.5\totalheight}{%
      \includegraphics[width=0.23\linewidth]{#3}%
    }%
  }
    \begin{tabular}{c|@{}c@{}c@{}c@{}}
    & Input &  Cross-view Correspondences & 3D Reconstruction \\
    
    \hline
    \vlab{\Dpig} &
    \incgthree{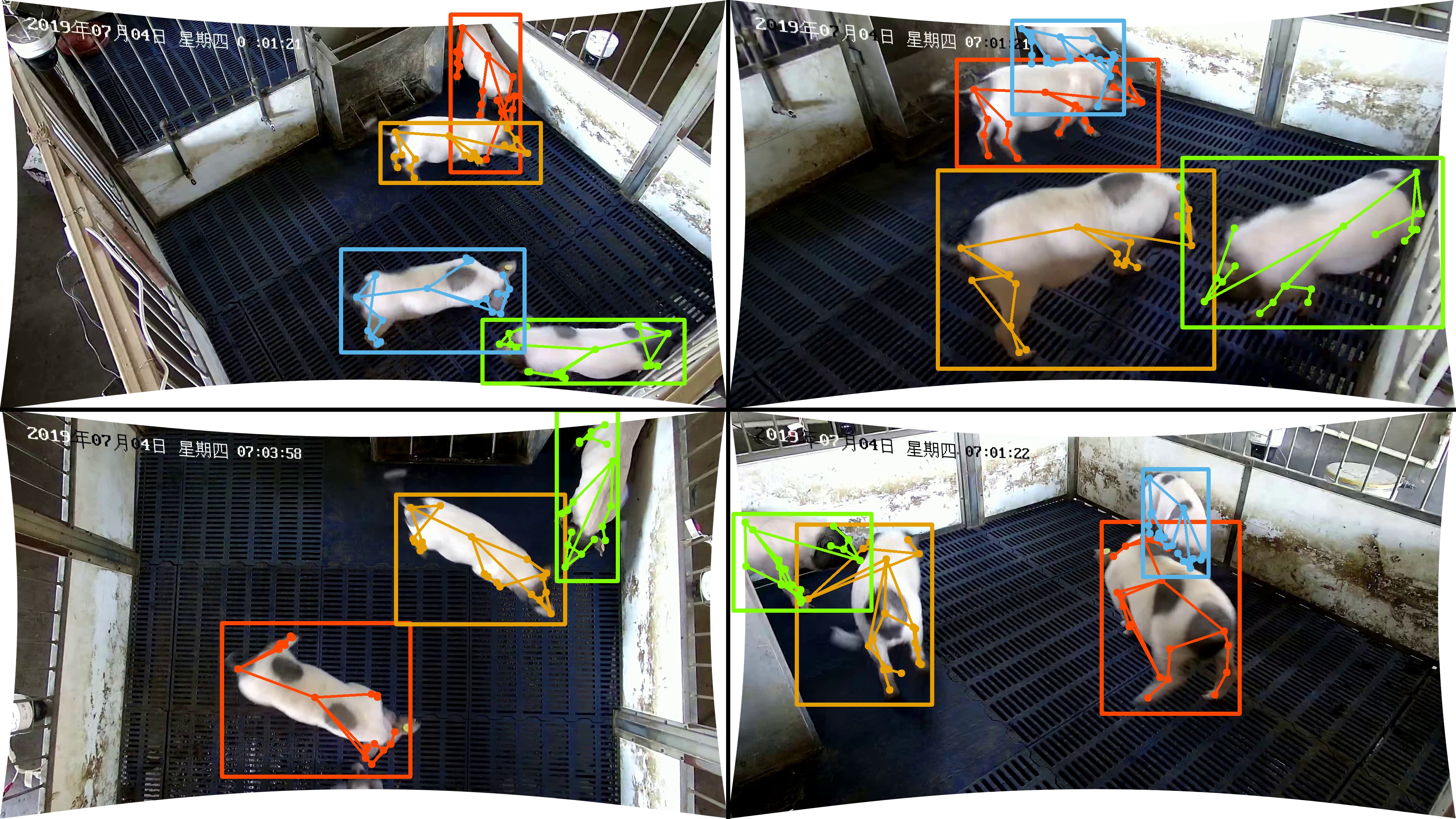}{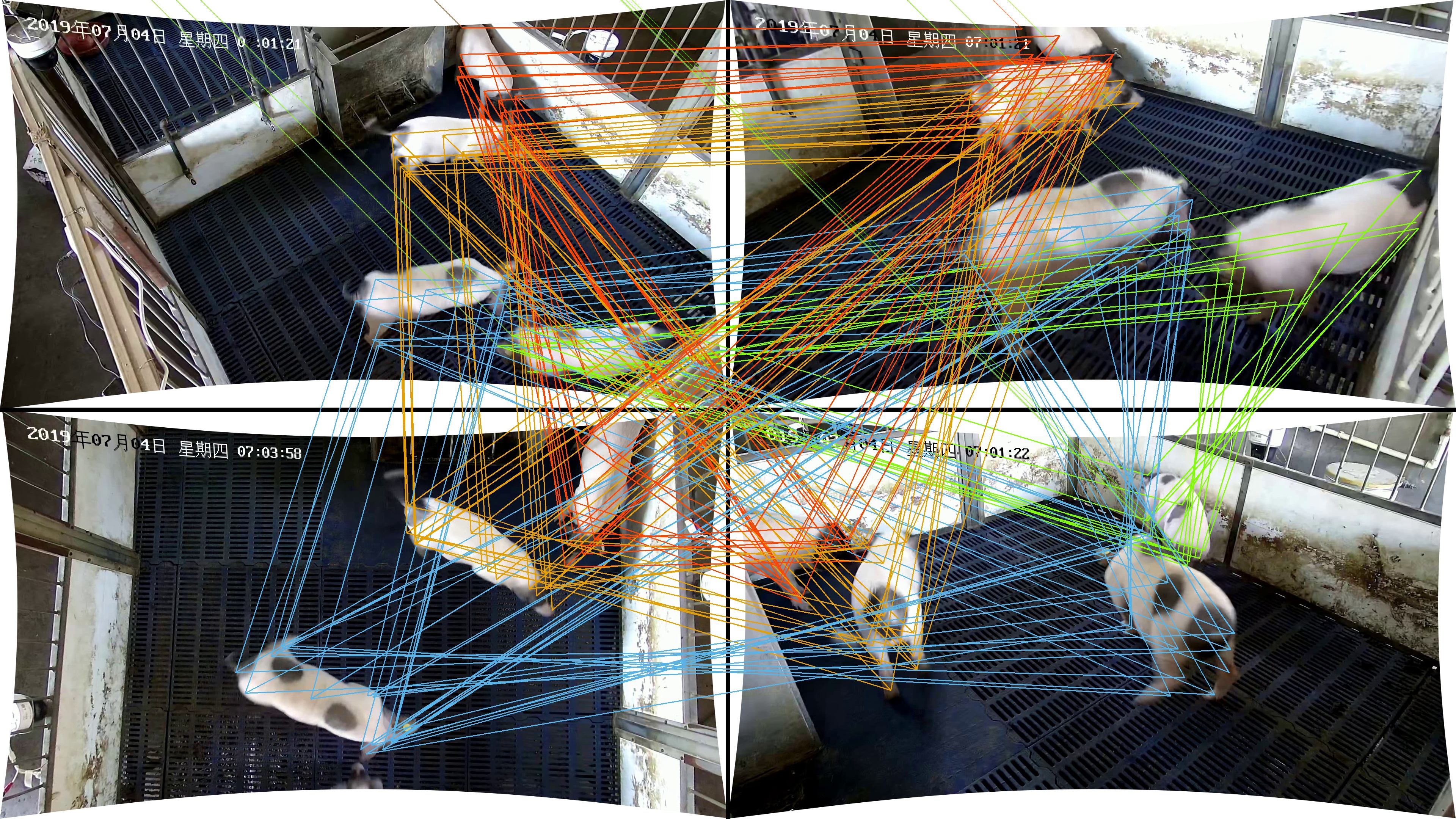}{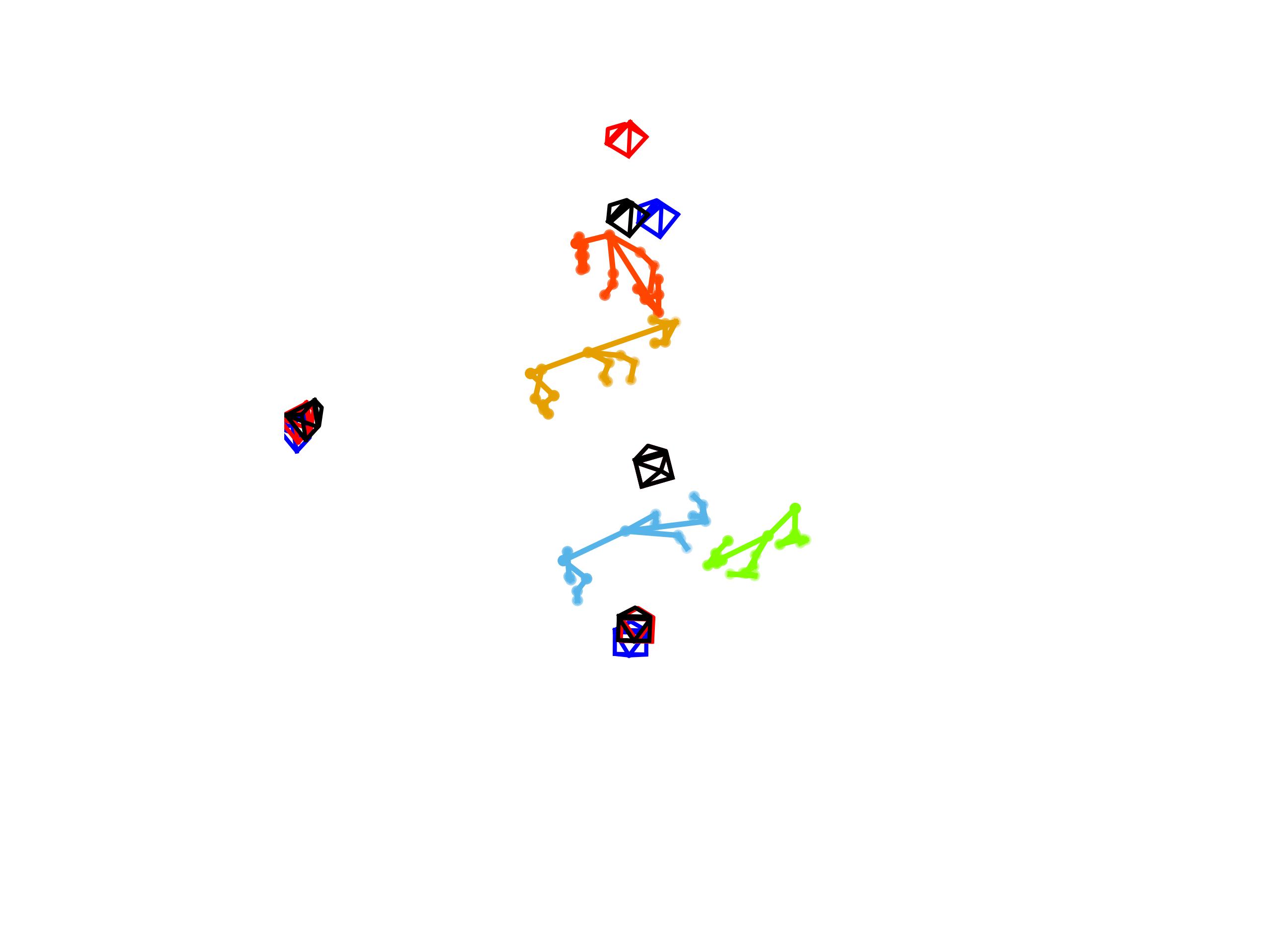}\\ \hline

    \vlab{\Dtoddler} &
    \incgthree{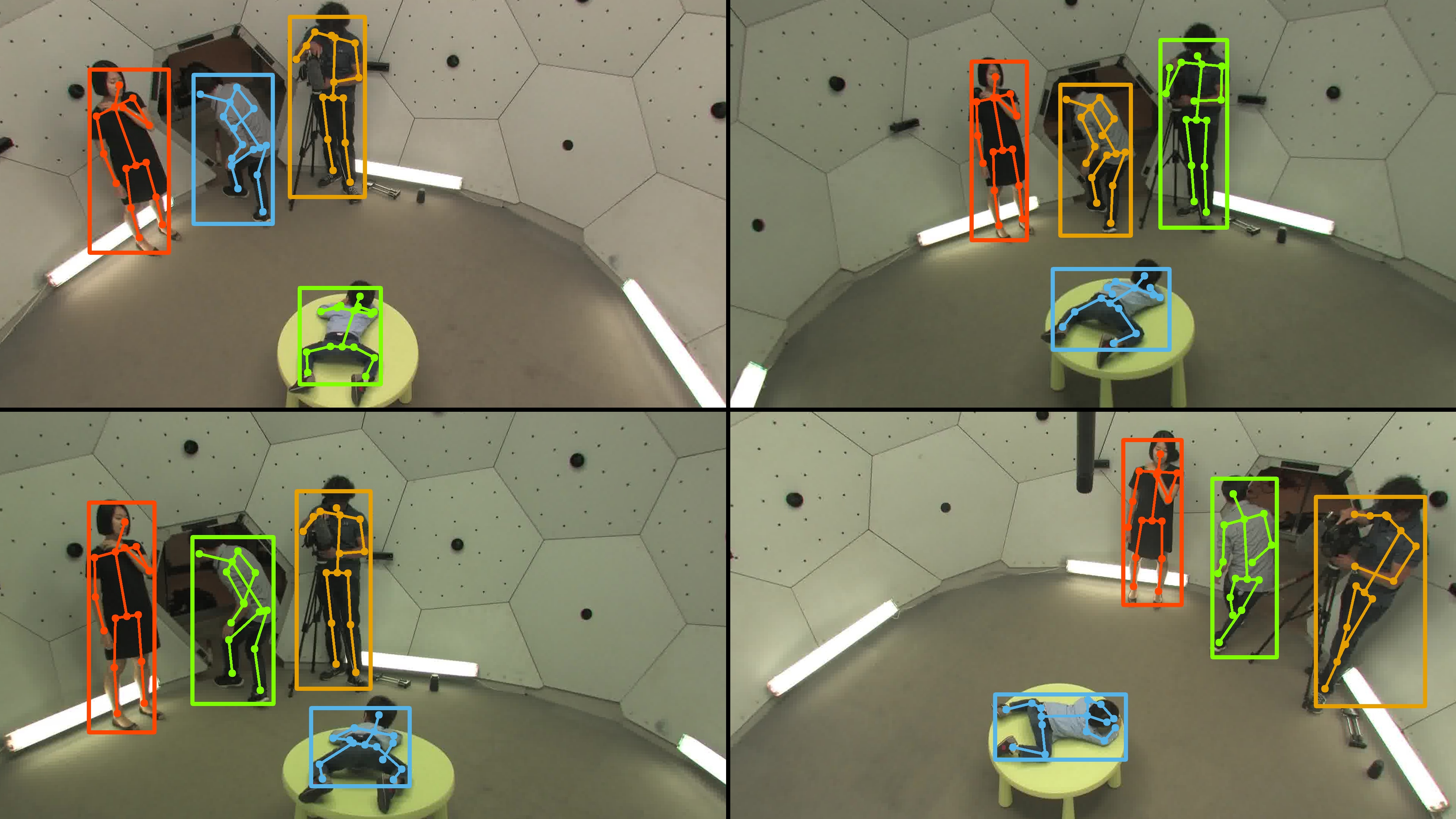}{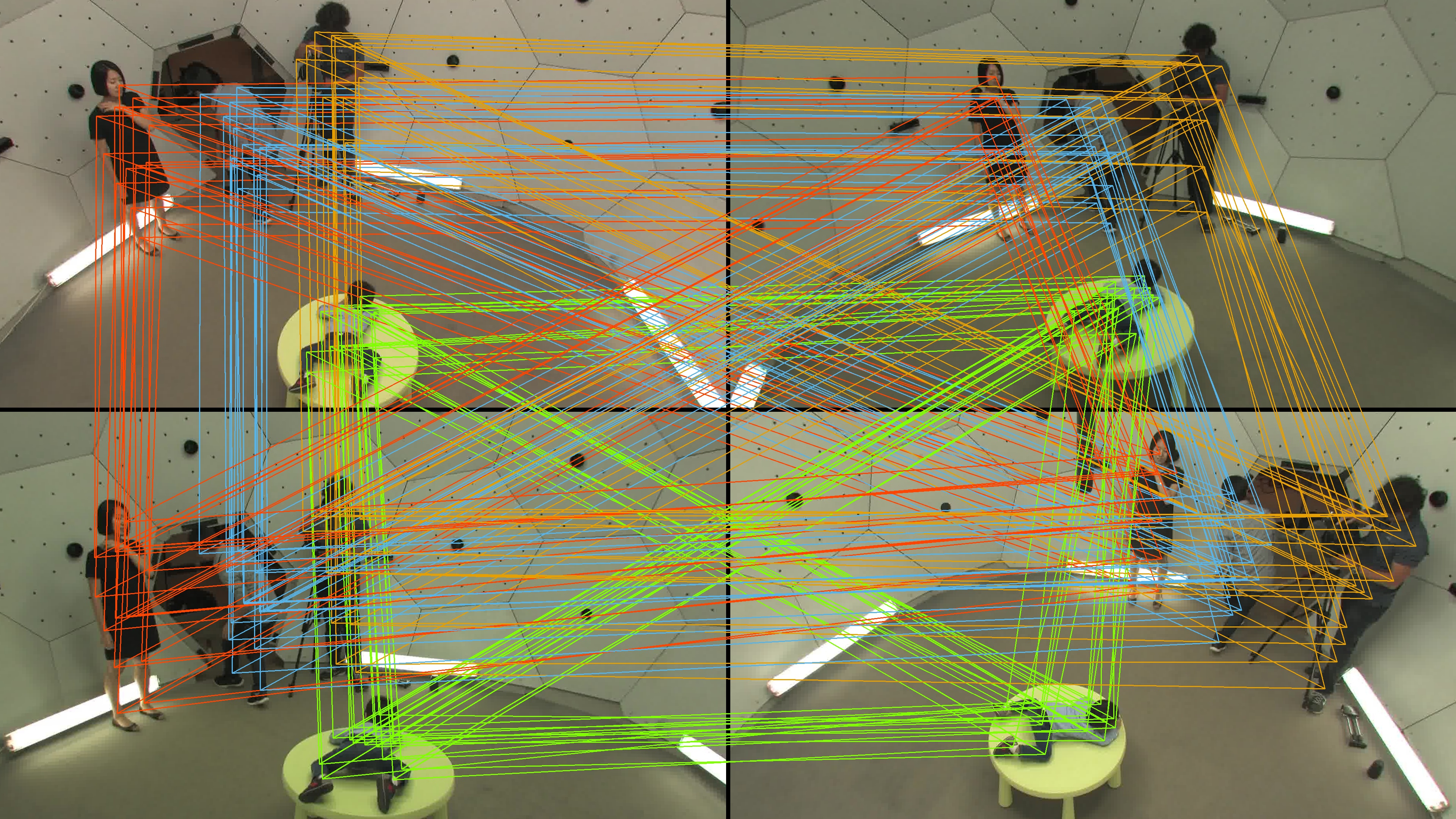}{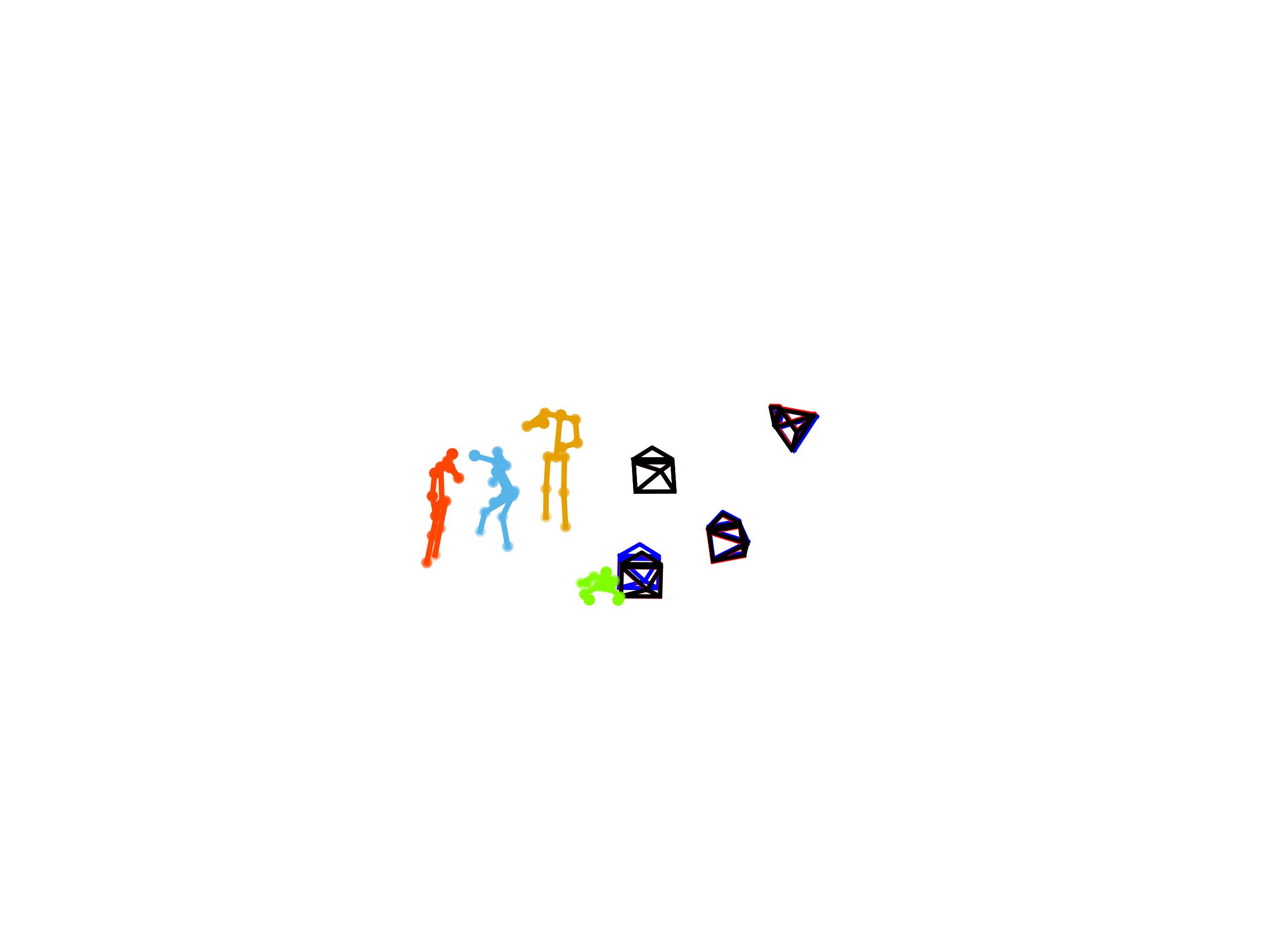}\\ \hline

    \end{tabular}

    \caption{Visualization for multi-view calibration results. 
    The ground-truth cameras are in black, the initial camera poses after registration are in blue, and the final bundle-adjustment results are shown in red. Our method establishes reliable correspondences and accurately estimates camera poses for 3D reconstruction. For quantitative evaluation, please refer to Table~\ref{table:multi_quan}.}
    \label{fig:multi_qual}
\end{figure}

\subsection{Multi-view Integration}\label{sec:eval_nview}

Fig.~\ref{fig:multi_qual} illustrates the multi-view camera calibration results. We utilize RANSAC~\cite{fischler_bolles:RANSAC} to identify the best camera poses for each pair of views by evaluating the geometric consistency based on the reprojection error. For class-specific models, we use the results in Table~\ref{tab:two_view_calib}. By integrating the pairwise camera poses and cross-view correspondences into a unified coordinate system, we further refine them through a nonlinear optimization of the reprojection errors. This highlights the capability of our method to automatically achieve extrinsic camera calibration, correspondence finding, and 3D reconstruction as a markerless motion capture within a single framework by simply capturing novel articulated animals in uncontrolled environments.

\subsection{Ablation Study}\label{sec:eval_ablation}

Table~\ref{tab:ablation} reports the contribution of our geometric consistency loss $\mathcal{L}_\mathrm{geom}$.  We use two class-specific models trained on quadrupeds (\Dpig) and bipeds (\Dpigeon) to demonstrate the effectiveness of this loss term. We observed that incorporating $\mathcal{L}_\mathrm{geom}$ significantly improves extrinsic calibration and matching accuracy by enforcing keypoint geometric consistency in both scenarios.

\begin{table}[]
    \centering
    \begin{tabular}{l|ccc|ccc}
    \toprule
         \multicolumn{1}{c}{}
         & \multicolumn{3}{c}{\Dpig~\cite{an2023three}}
         & \multicolumn{3}{c}{\Dpigeon~\cite{naik20233d}} \\ 
         &$E_R\downarrow$ &$E_\mathbf{t}\downarrow$&$P\uparrow$&$E_R\downarrow$&$E_\mathbf{t}\downarrow$&$P\uparrow$ \\ \midrule

w/o $\mathcal{L}_\mathrm{geom}$   & 11.35 & 7.78 & 0.83 & 20.70 & 12.02 & 0.99  \\
w $\mathcal{L}_\mathrm{geom}$   & 8.30 & 4.71 & 0.90 & 9.20 & 5.70 & 0.99  \\
    \bottomrule
    
    \end{tabular}
    \caption{Contribution of the geometric consistency loss $\mathcal{L}_\mathrm{geom}$. }
    \label{tab:ablation}
\end{table}

\section{Conclusion}

We proposed SteerPose which learns to \textit{rotate} an input 2D pose to match one in another view at a relative rotation $R$. SteerPose leverages the articulation of 3D poses as prior knowledge, and functions as $g'(\cdot)$ which performs \textit{mental rotation} to form a rotation-covariant transform of 3D poses under projection $f(\cdot)$ satisfying $f(g(\cdot))=g'(f(\cdot))$ where $g(\cdot)$ represents the 3D rotation by $R$.

Our method combines pretrained SteerPose with the Sinkhorn algorithm to perform simultaneous extrinsic calibration and correspondence search as an optimization problem during inference. The novel geometric consistency loss explicitly guarantees that the optimized rotation and match have a valid solution for relative translation. Experimental results on diverse in-the-wild datasets of humans and animals demonstrated the effectiveness and robustness of the proposed method. In particular, by leveraging the articulation of human or animal poses, our calibration approach effectively addresses challenging wide-baseline configurations and textureless backgrounds, which are common in uncontrolled environments such as those involving animals in the wild. Furthermore, we showed that our class-agnostic model enables the 3D reconstruction of novel animals in multi-view setups.

Our proposed calibration requires only 2D pose estimation or annotation, which is becoming widely available in many contexts.  We believe that the proposed method can be used as a practical calibration tool for studies involving on-site multi-view 3D captures, and can provide triangulated 3D poses for training monocular 3D pose estimators of novel targets.

\bibliographystyle{unsrtnat}
\bibliography{egbib}  

\clearpage

\begin{appendices}
\appendixpage
\suppressfloats

\section{Generalizability of SteerPose} \label{sec:gen_steerpose}

Table~\ref{tab:agnostic} reports the test loss of SteerPose trained under different conditions.  The ``class-specific'' column represents SteerPose trained on the training set of the target dataset itself.  The ``class-agnostic'' columns represent SteerPose trained on Animal3D dataset~\cite{xu2023animal3d}, which contains 40 quadruped species.  The results indicate that SteerPose, when trained for quadrupeds as a class-agnostic model, performs comparably well on the same animals from a different dataset, \Ddog, and generalizes effectively to unseen species, \ie, species not included in Animal3D, such as \Dcheetah and \Dpig. 

Note that there were small discrepancies between the skeleton structures of \Ddog and Animal3D.  For example, Animal3D has a neck, whereas \Ddog does not.  To evaluate the class-agnostic SteerPose trained with Animal3D using the test set from \Ddog, we used SteerPose itself to align the skeleton structure by applying identity rotation to the 2D poses.  That is, by masking the missing 2D joints in the input, rotating 2D poses to the same viewpoint by SteerPose can estimate the 2D keypoint positions of the masked joints.  This can also fill in missing keypoints due to occlusions.  This ``keypoint completion'' step can enhance the generalizability of the class-agnostic model as shown in the two ``Class-agnostic'' columns in Table~\ref{tab:agnostic}.

Table~\ref{table:multi_quan} also demonstrates that the class-agnostic model achieves comparable results in multi-view calibration and cross-view correspondence matching tasks against the class-specific models, as detailed in Sec.~\ref{sec:additional_results}.

\section{Comparison with Other Methods} \label{sec:comp_others}

Table~\ref{tab:comp_approach_transposed} summarizes the advantages of leveraging articulation in our approach for extrinsic camera calibration.  As shown in Fig.~3 of the main text, the 2D poses as structured correspondences can realize robust calibrations, in particular for wide-baseline scenarios.

\begin{table}[t]
\centering
    \begin{tabular}{l|c|c|c}
        \toprule
        Target & Class-specific & Class-agnostic$^\ast$ & Class-agnostic\\
        \hline
        \Dcheetah~\cite{joska2021acinoset} & 0.098 & 0.217 & 0.202 \\
\Dpig~\cite{an2023three} & 0.121 & 0.21 & 0.173 \\
\Ddog~\cite{an2023three} & 0.166 & 0.208 & 0.16 \\
         \bottomrule
    \end{tabular}
    \caption{Generalizability of SteerPose.  Both ``class-specific'' and ``class-agnostic'' columns list the average keypoint errors for the target test set.  The class-agnostic SteerPose generalizes well to novel animals. $^\ast$ indicates results obtained without applying keypoint completion.}
    \label{tab:agnostic}
\end{table}

\begin{table}[t]
    \centering
    \resizebox{\columnwidth}{!}{%
    \begin{tabular}{lccc}
        \toprule
        \textbf{Method} &
        Correspondence Search &
        Joint Optimization  &  
        Structured Correspondence\\
        \midrule
        RelPose++~\cite{lin2024relpose++}          & $\times$     & $\times$     & $\times$ \\
        SuperGlue~\cite{sarlin2020superglue}       & $\checkmark$ & $\times$     & $\times$ \\
        Diff. 8-Pts~\cite{roessle2023end2end}      & $\checkmark$ & $\checkmark$ & $\times$ \\
        \DMaster~\cite{mast3r_arxiv24}               & $\checkmark$ & $\checkmark$ & $\times$ \\
        \textbf{Ours}                              & $\checkmark$ & $\checkmark$ & $\checkmark$ \\
        \bottomrule
    \end{tabular}
    }
    \caption{Comparison of key properties of different camera pose estimation methods. Use of the target 2D poses as structured correspondences can contribute to both calibration and matching.}
    \label{tab:comp_approach_transposed}
\end{table}

\begin{figure}[t]
    \centering
    \def\vlab#1{\rotatebox[origin=c]{90}{#1}}
    \newcommand{\incg}[2]{%
  \raisebox{-0.5\totalheight}{%
    \includegraphics[width=0.13\linewidth]{#1}%
    \includegraphics[width=0.13\linewidth]{#2}%
  }%
  }

    \begin{tabular}{c|@{}c@{}c@{}c@{}c@{}}
    & LightGlue~\cite{lindenberger2023lightglue} & \DMaster~\cite{mast3r_arxiv24} & Ours & 3D Pose (GT) \\
    \hline
    \vlab{\Ddog} &
    \incg{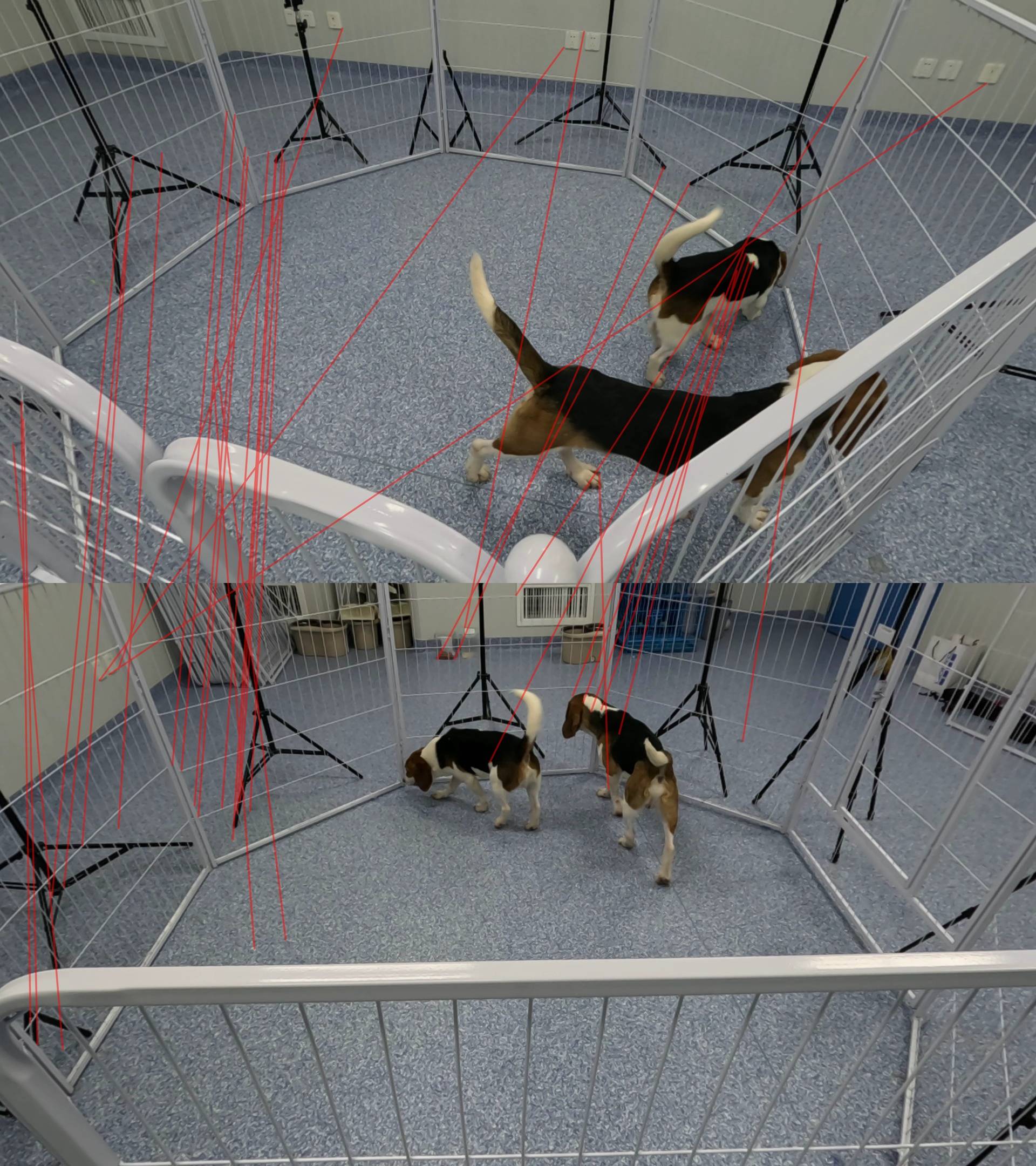}{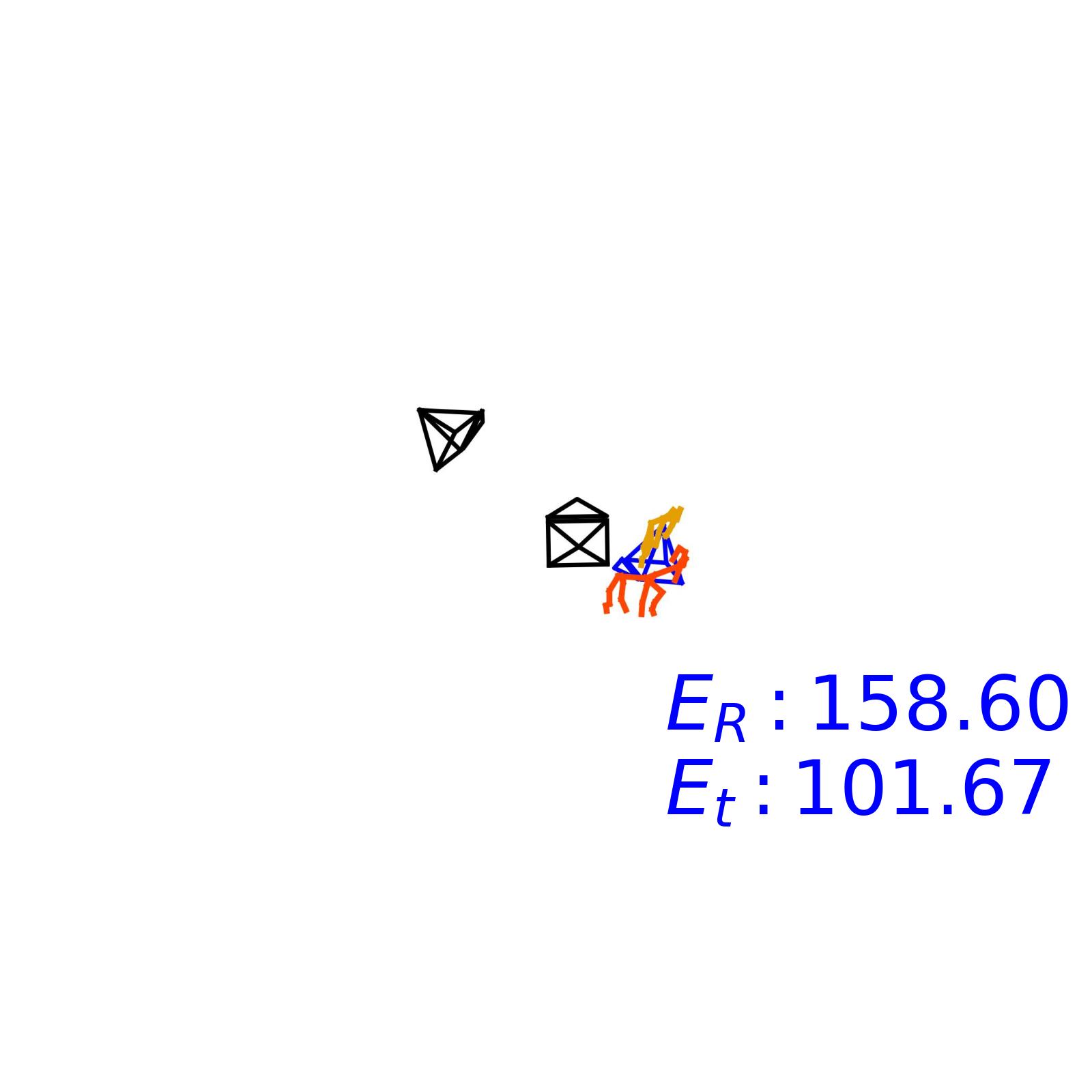} &        
    \incg{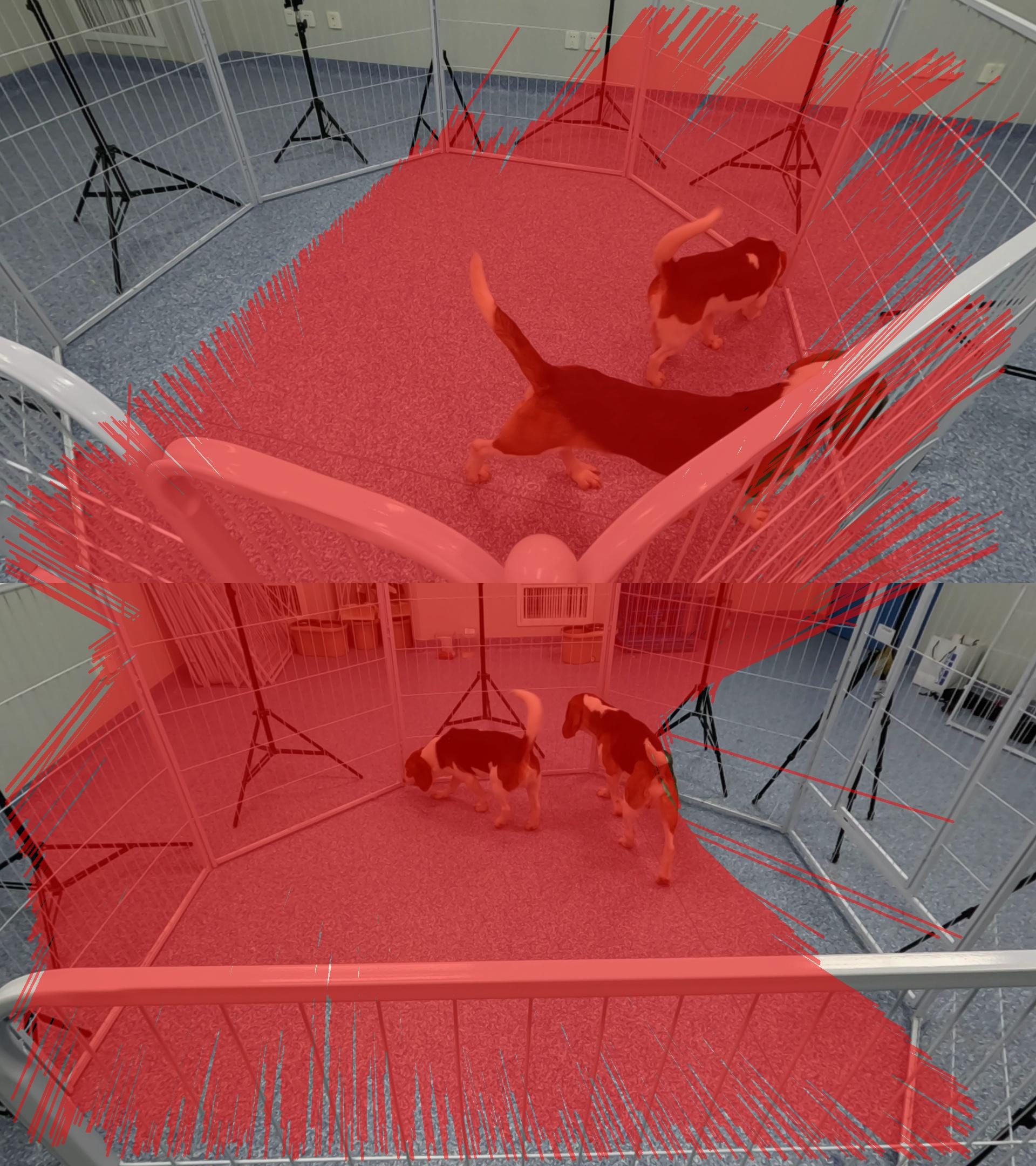}{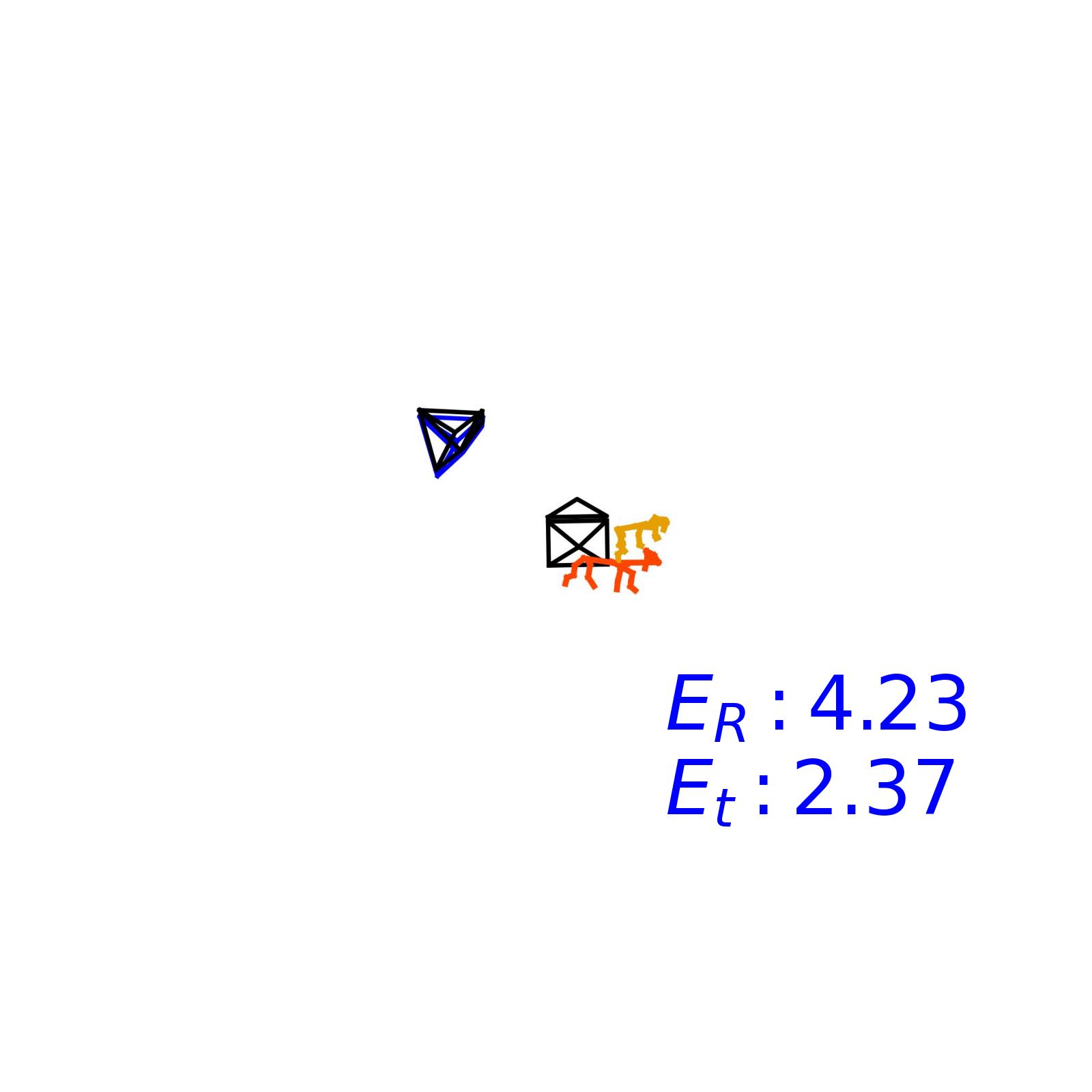}&
    \incg{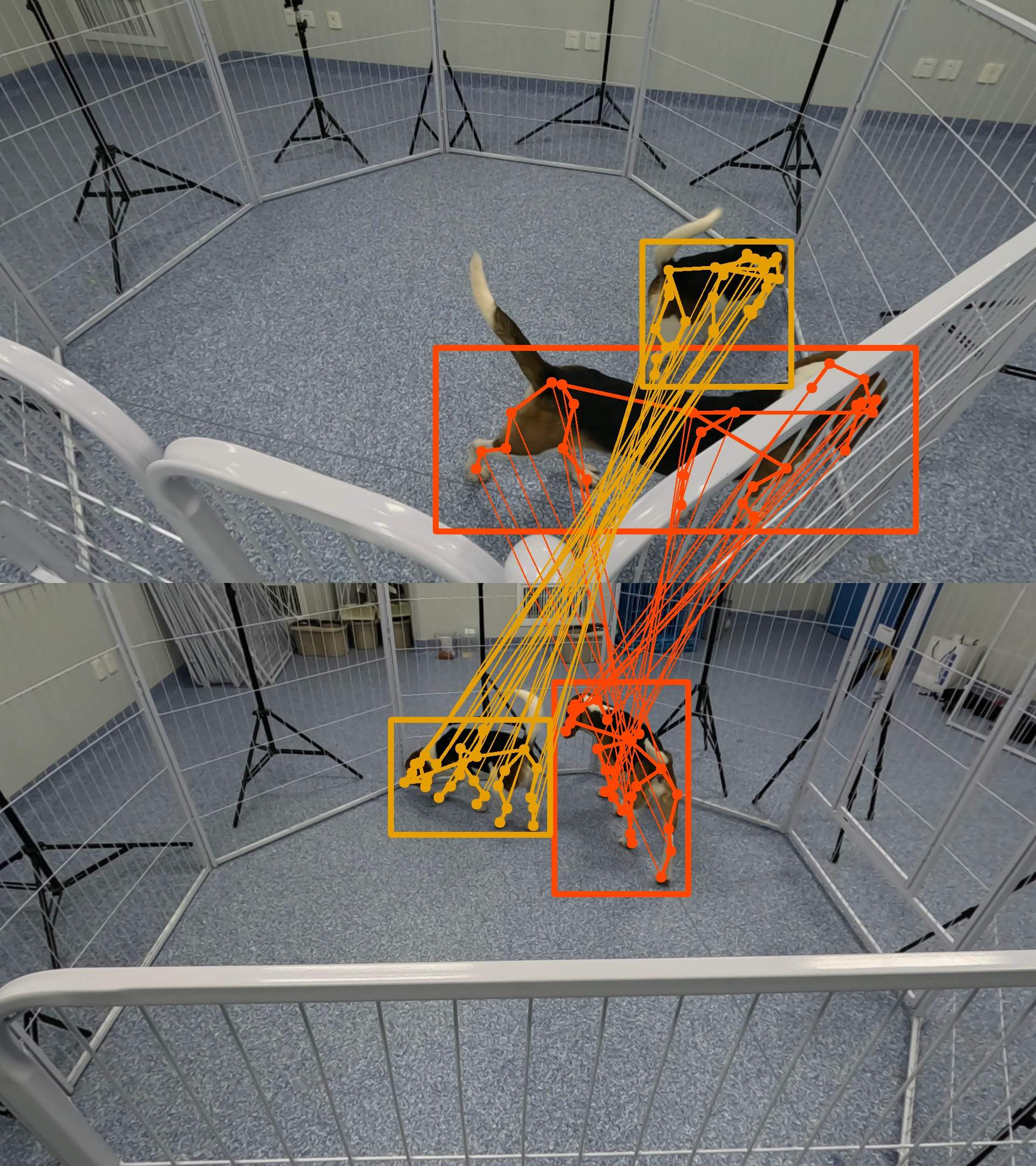}{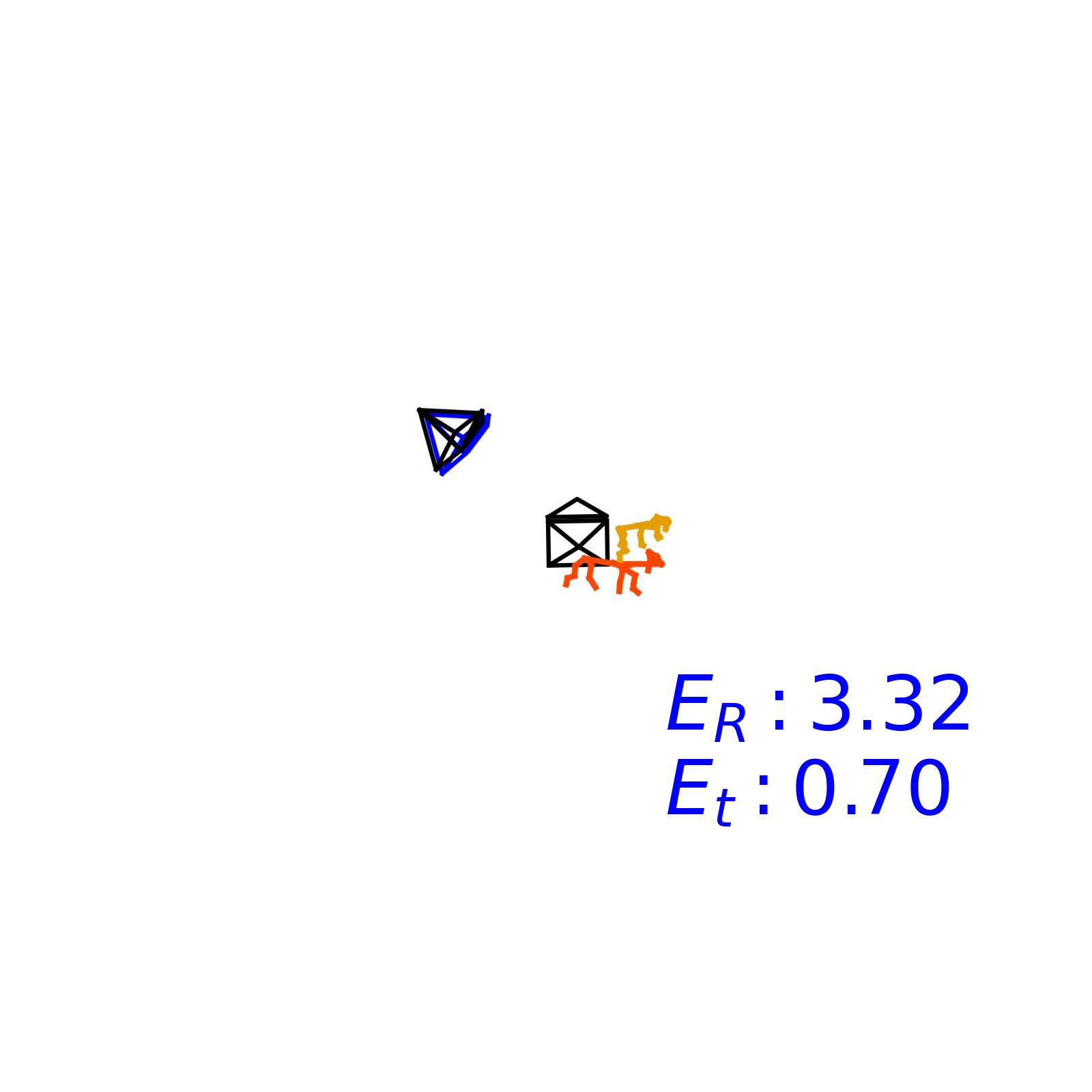} &
    \raisebox{-0.5\totalheight}{%
    \includegraphics[width=0.13\linewidth]{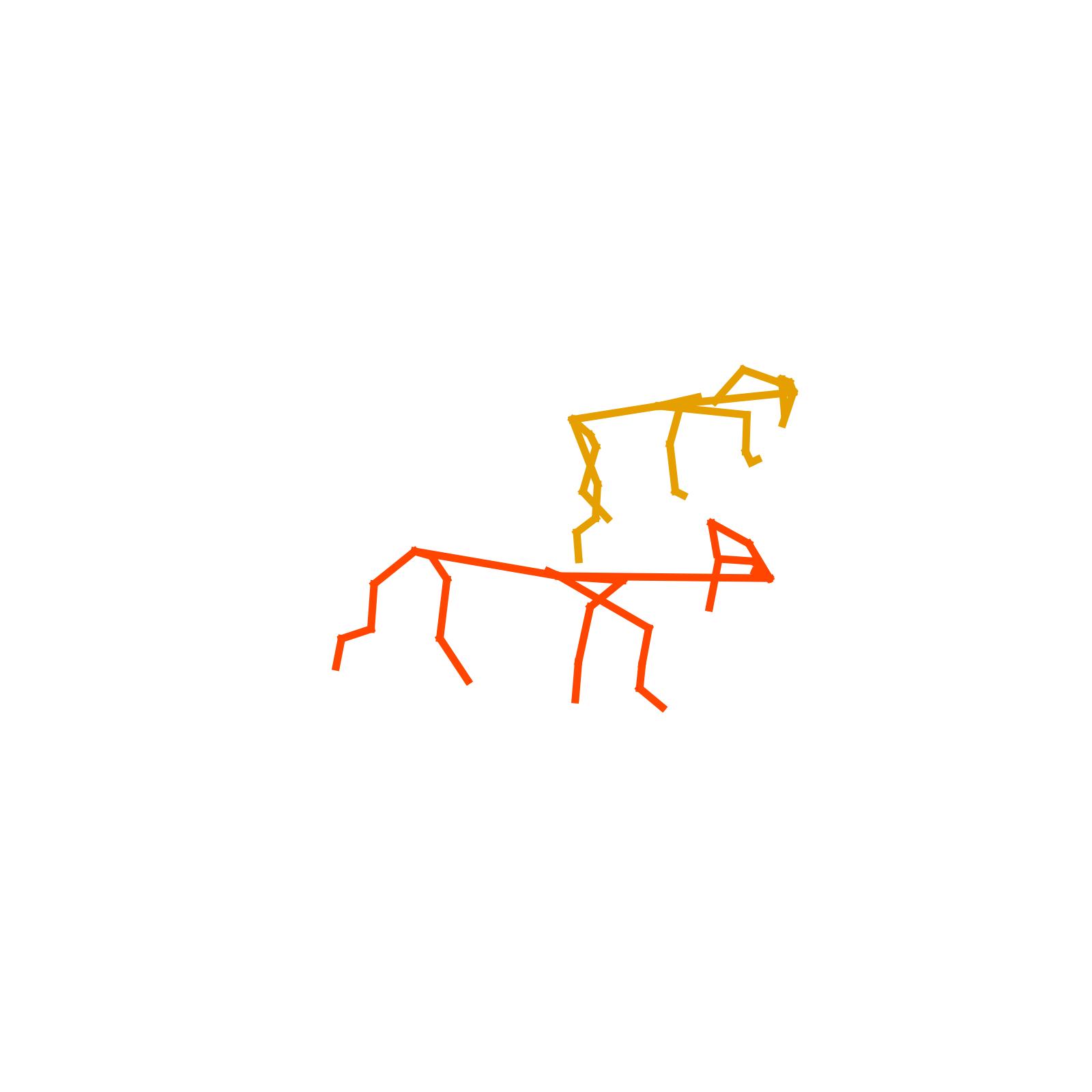}
  }%
    
    \\
    \hline
    \vlab{\Dtoddler} &
    \incg{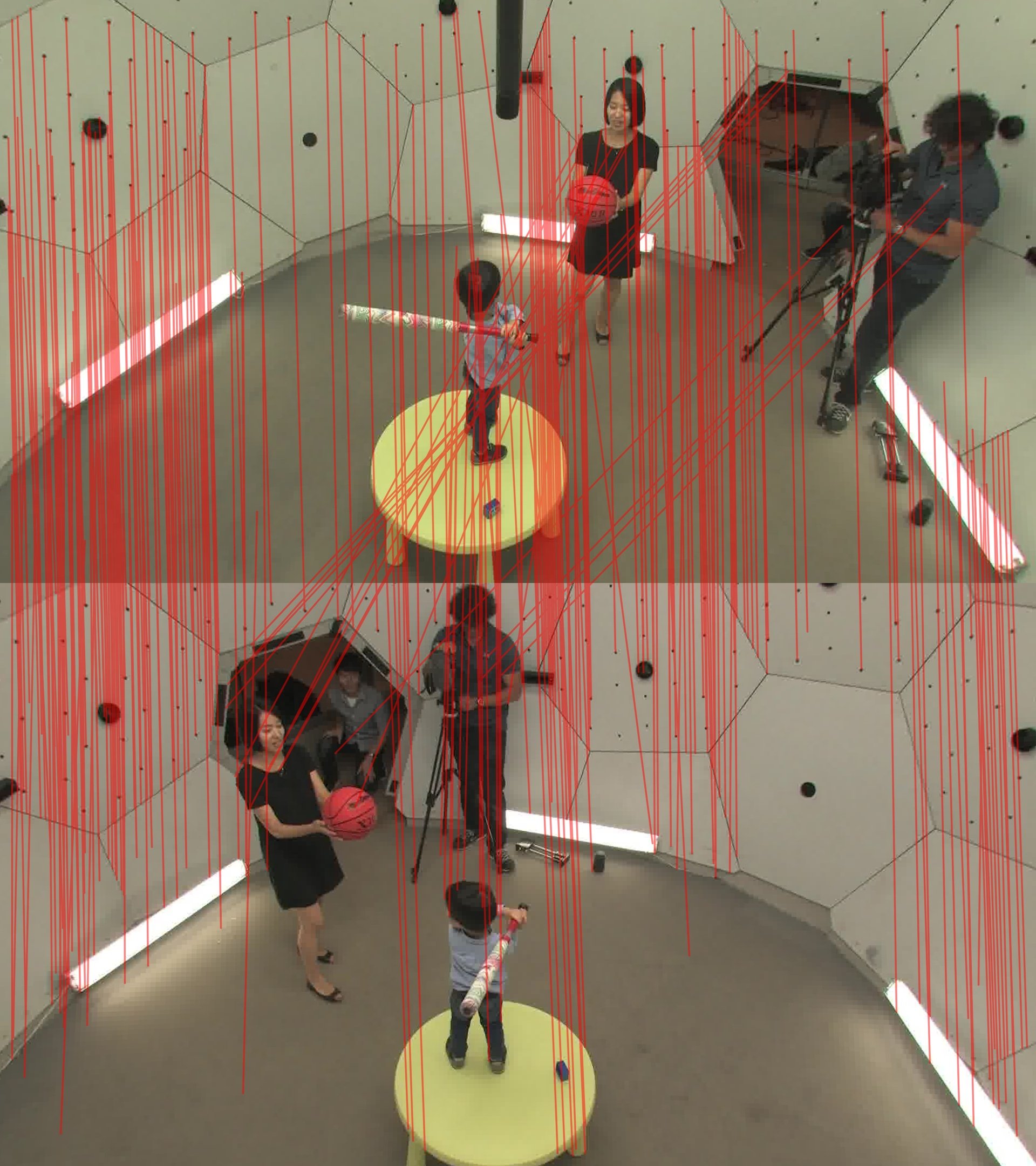}{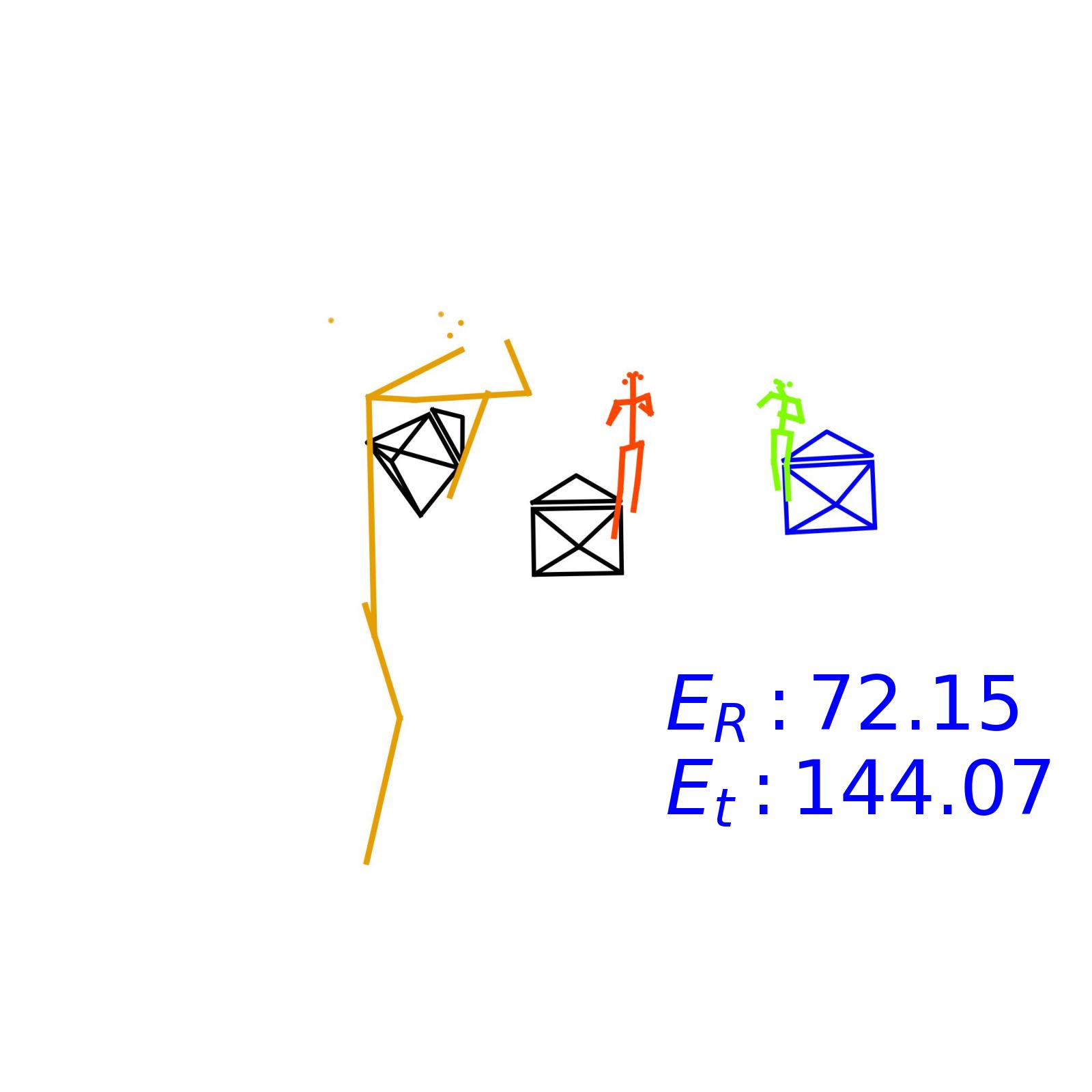} &        
    \incg{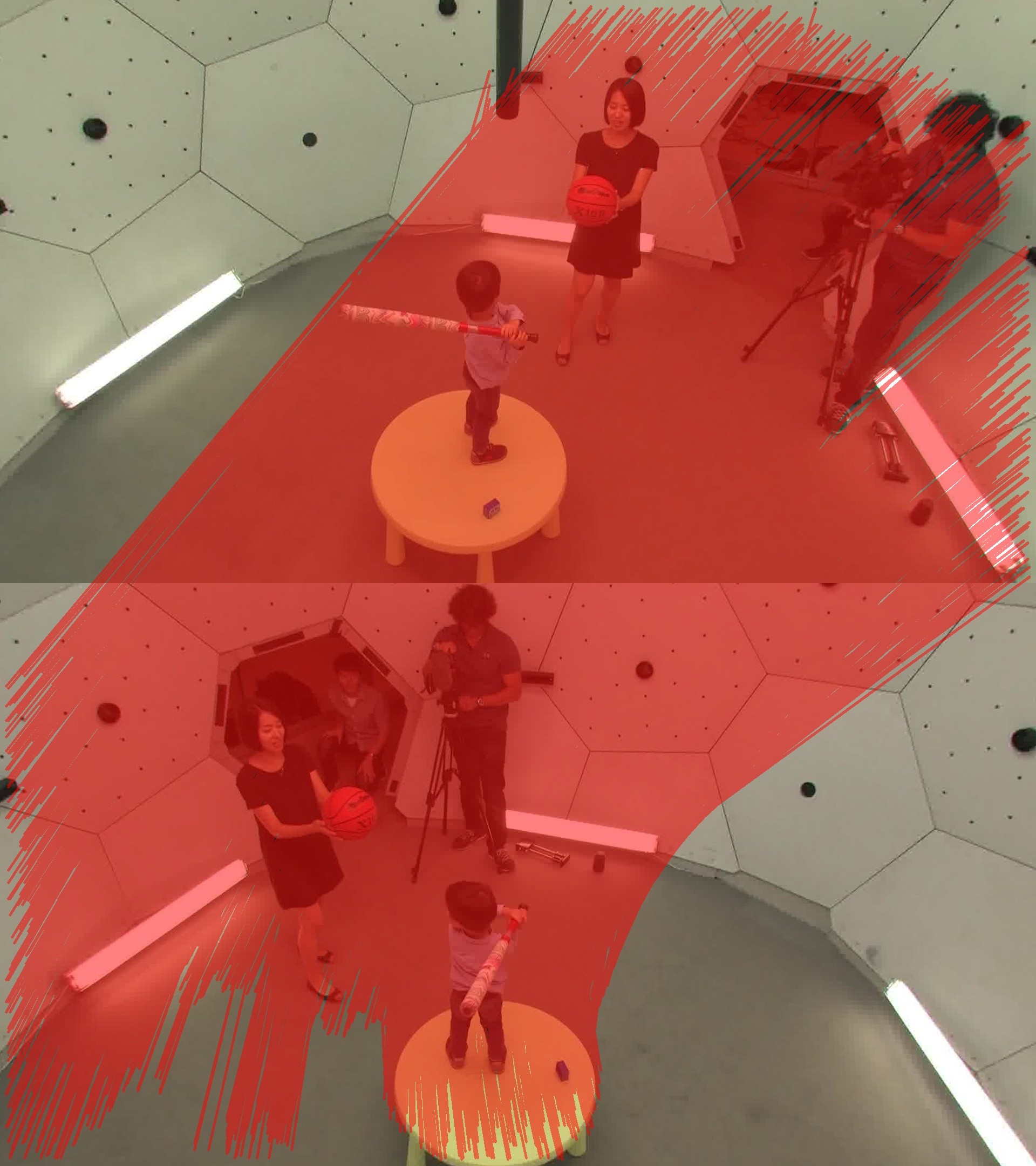}{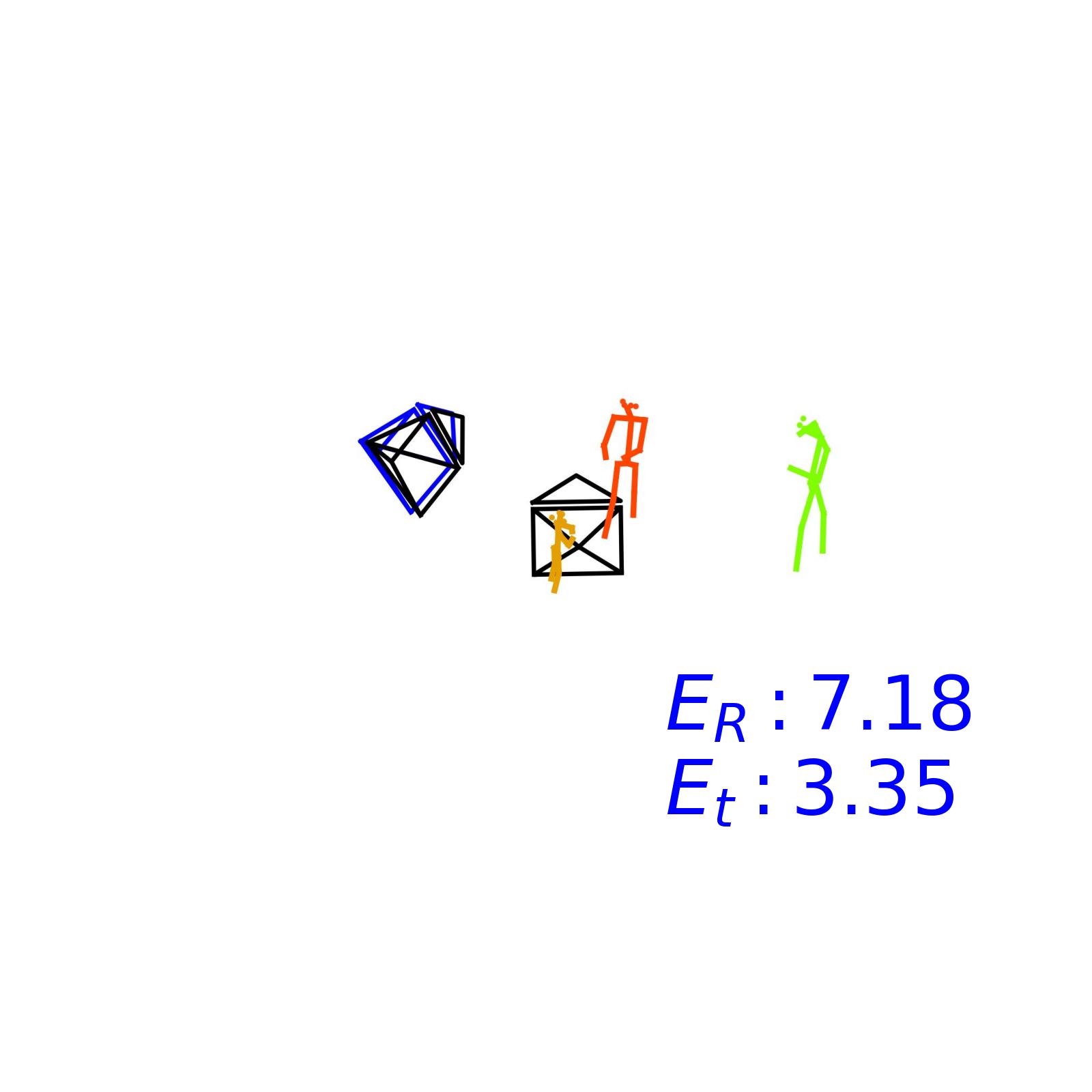}&
    \incg{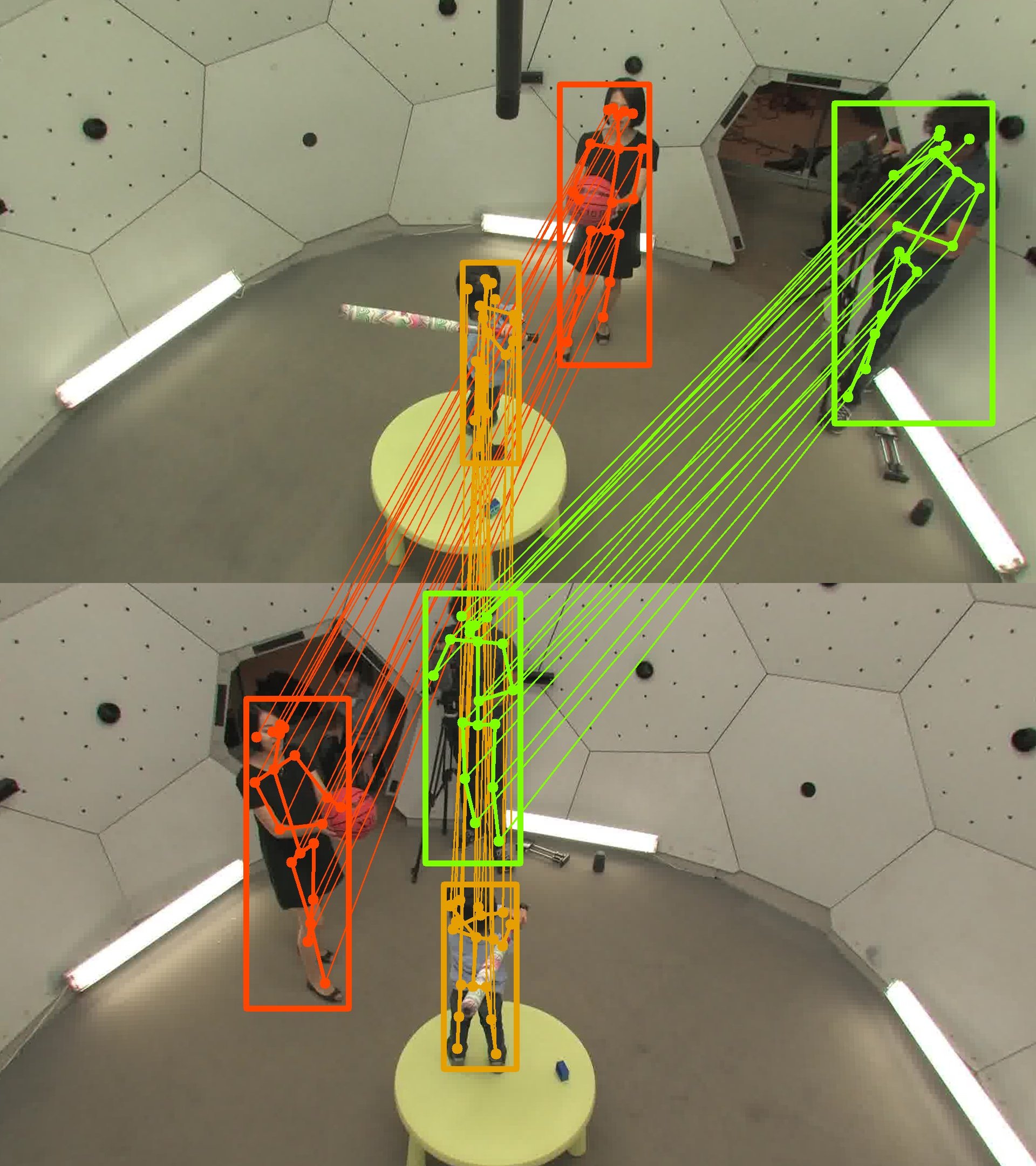}{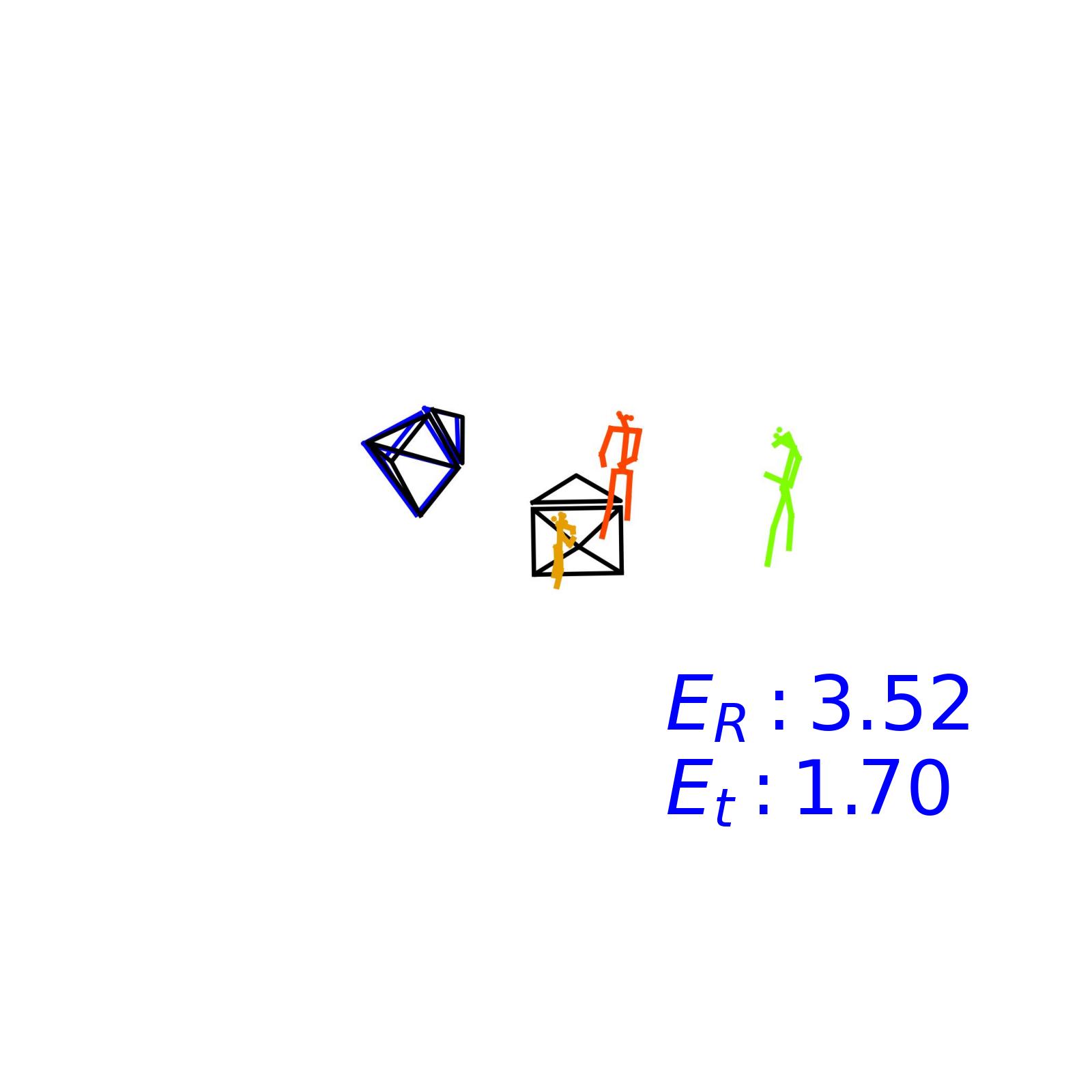} &
    \raisebox{-0.5\totalheight}{%
    \includegraphics[width=0.13\linewidth]{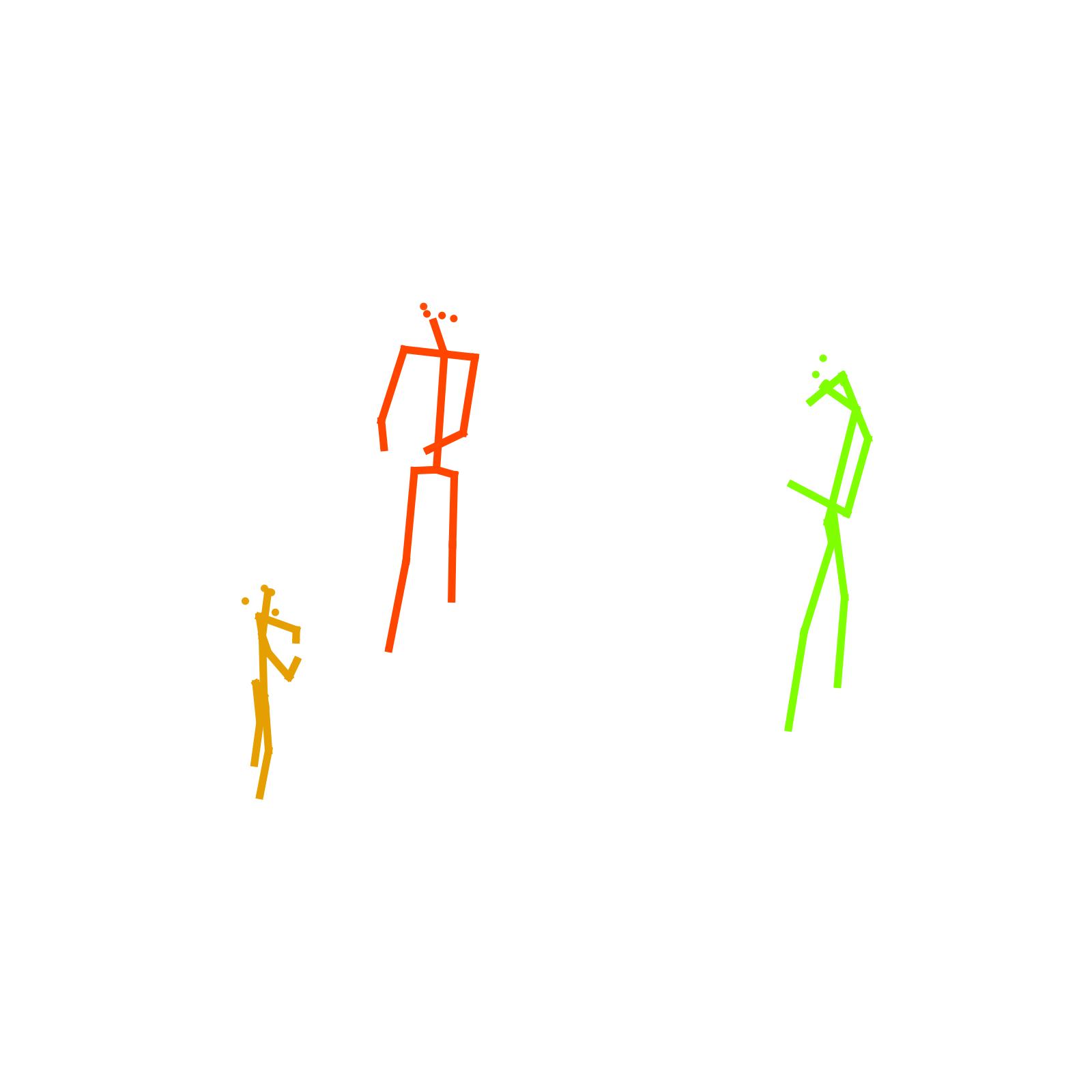}
  }%
    \\
    \hline
    \vlab{\Dvolleyball} &
    \incg{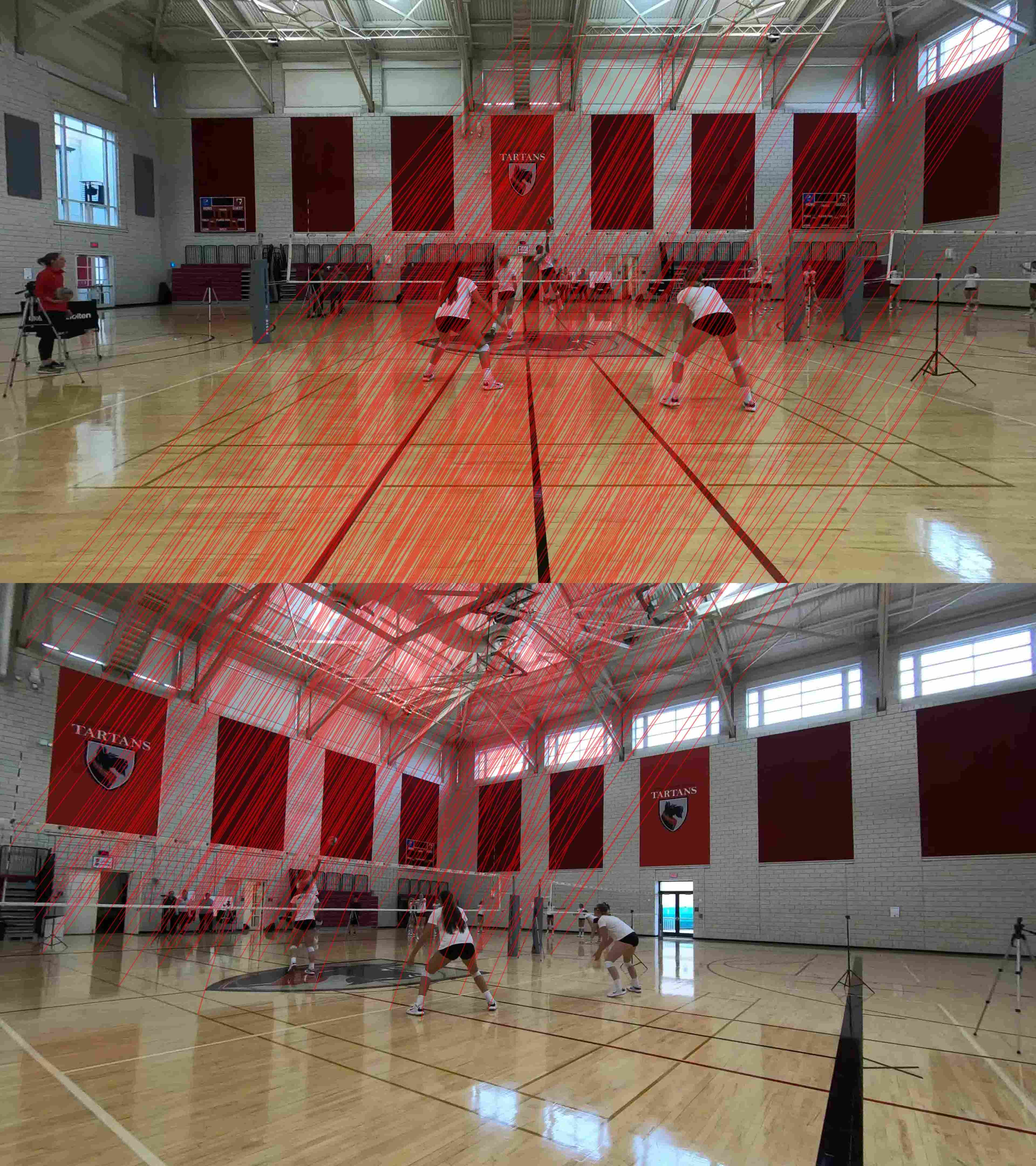}{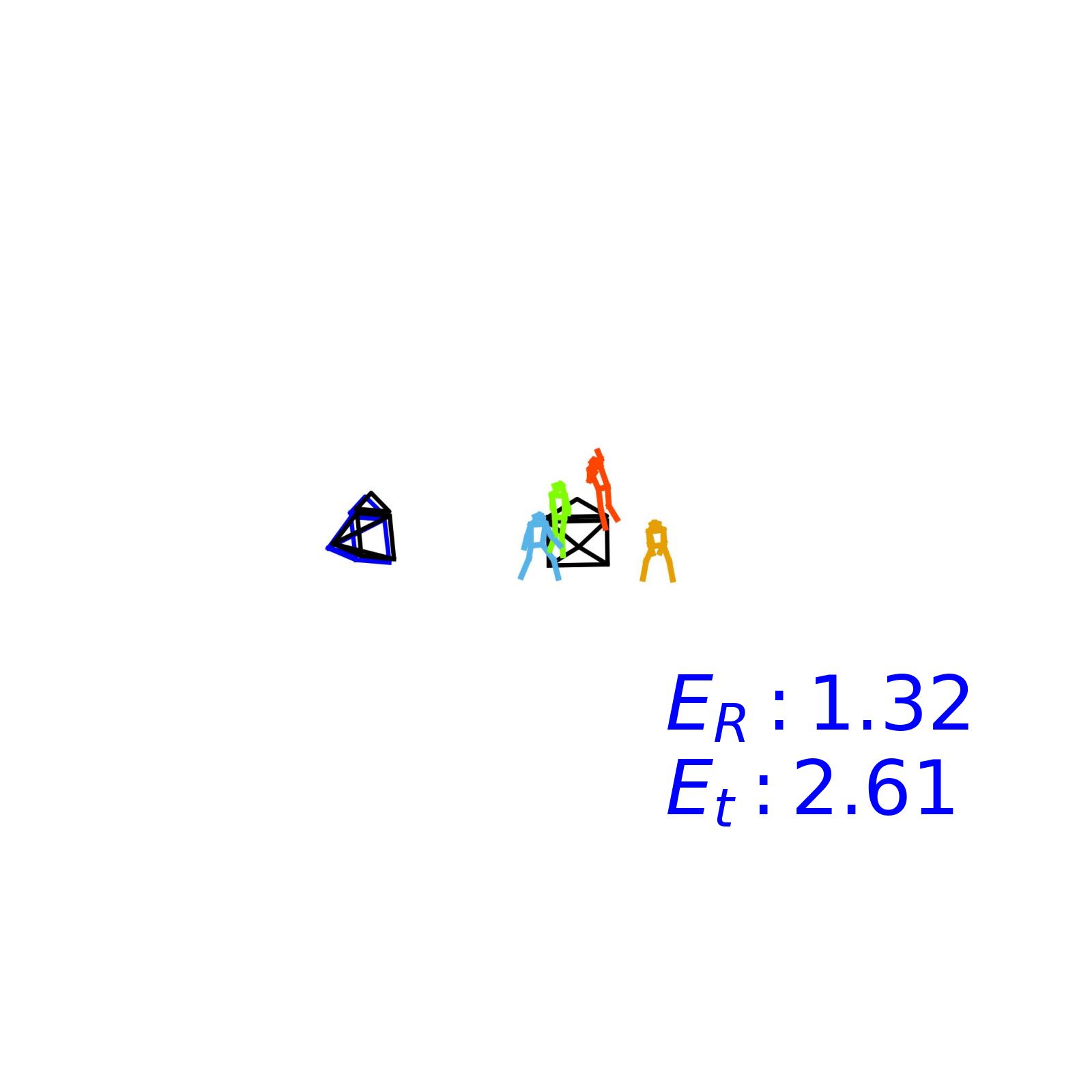} &        
    \incg{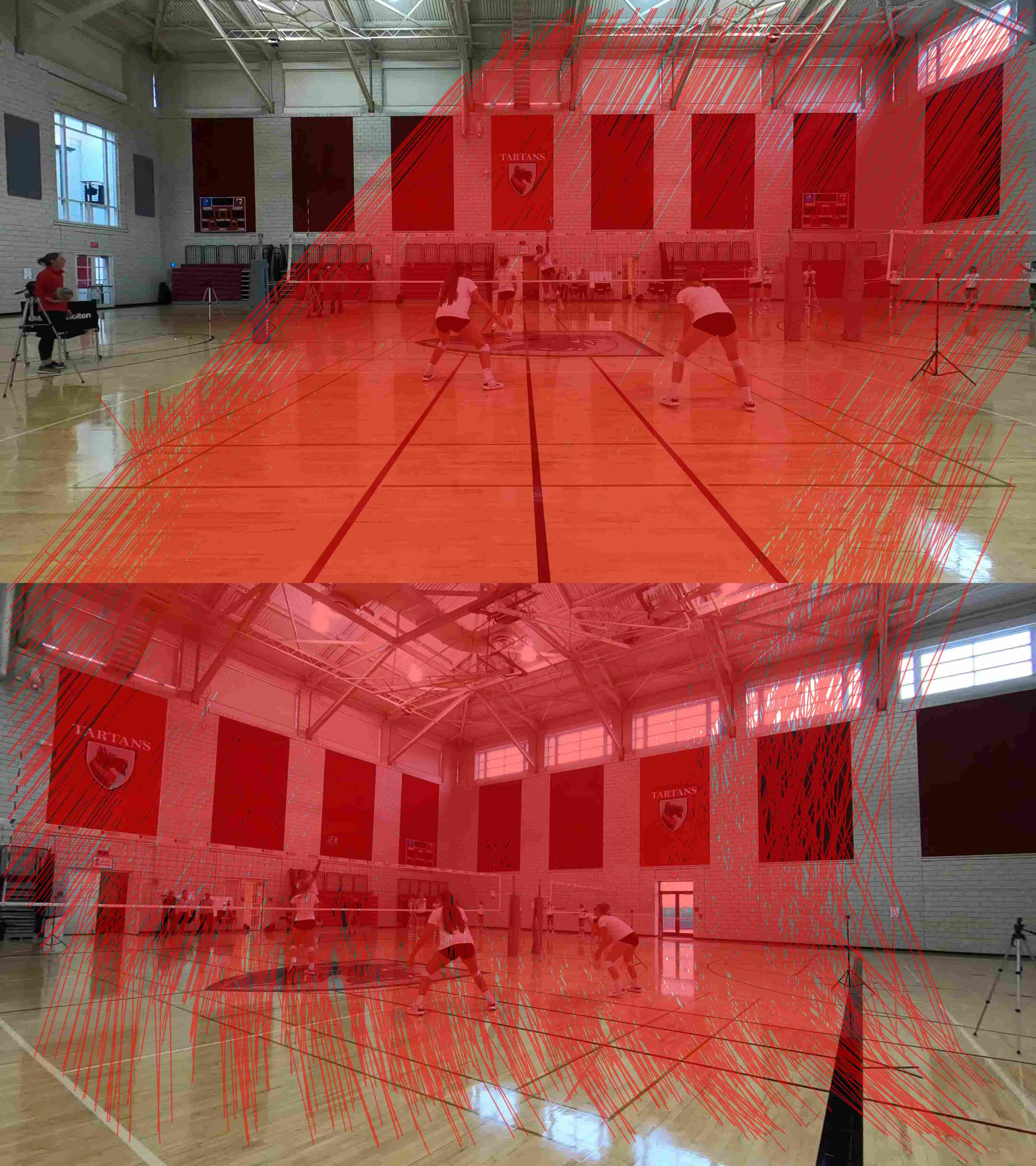}{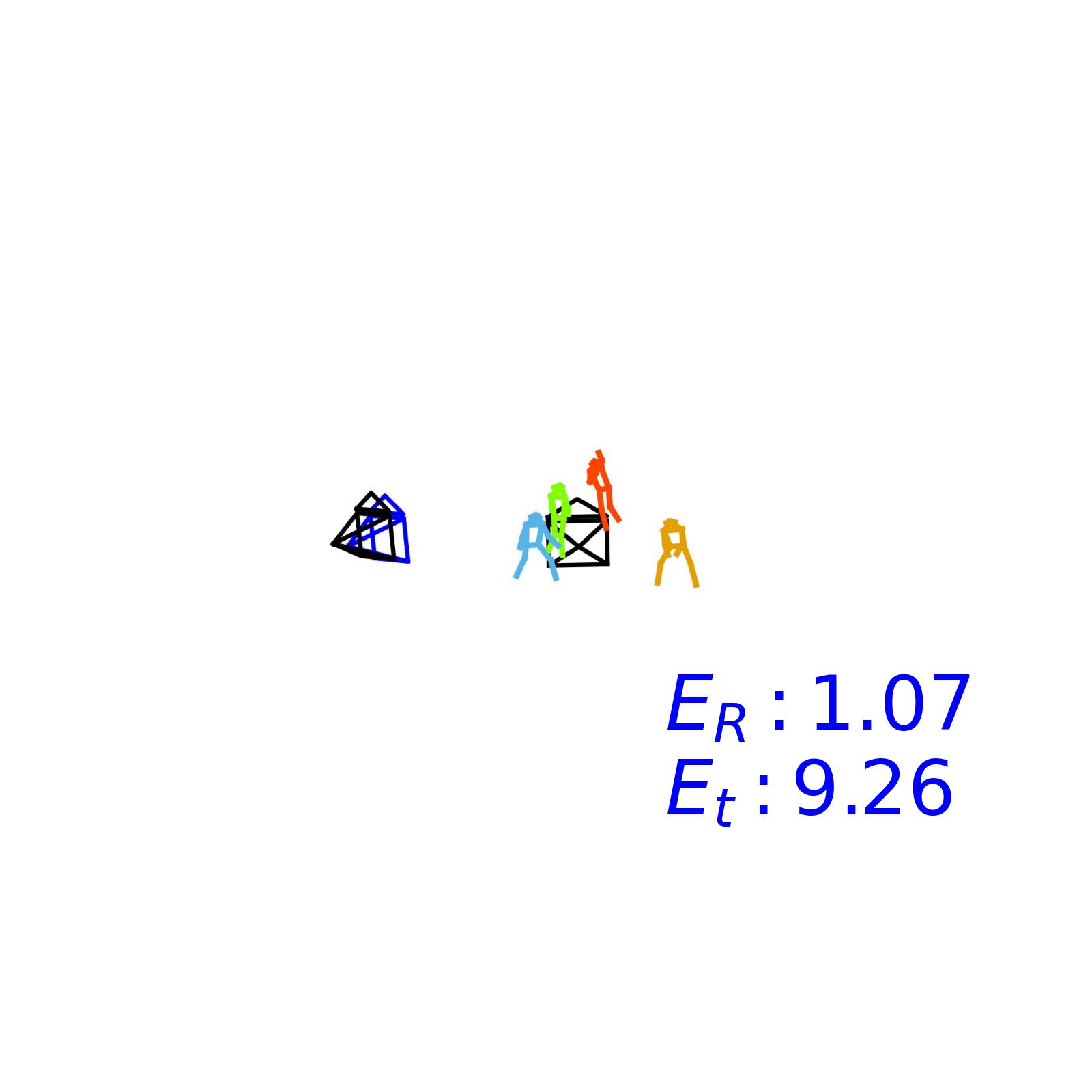}&
    \incg{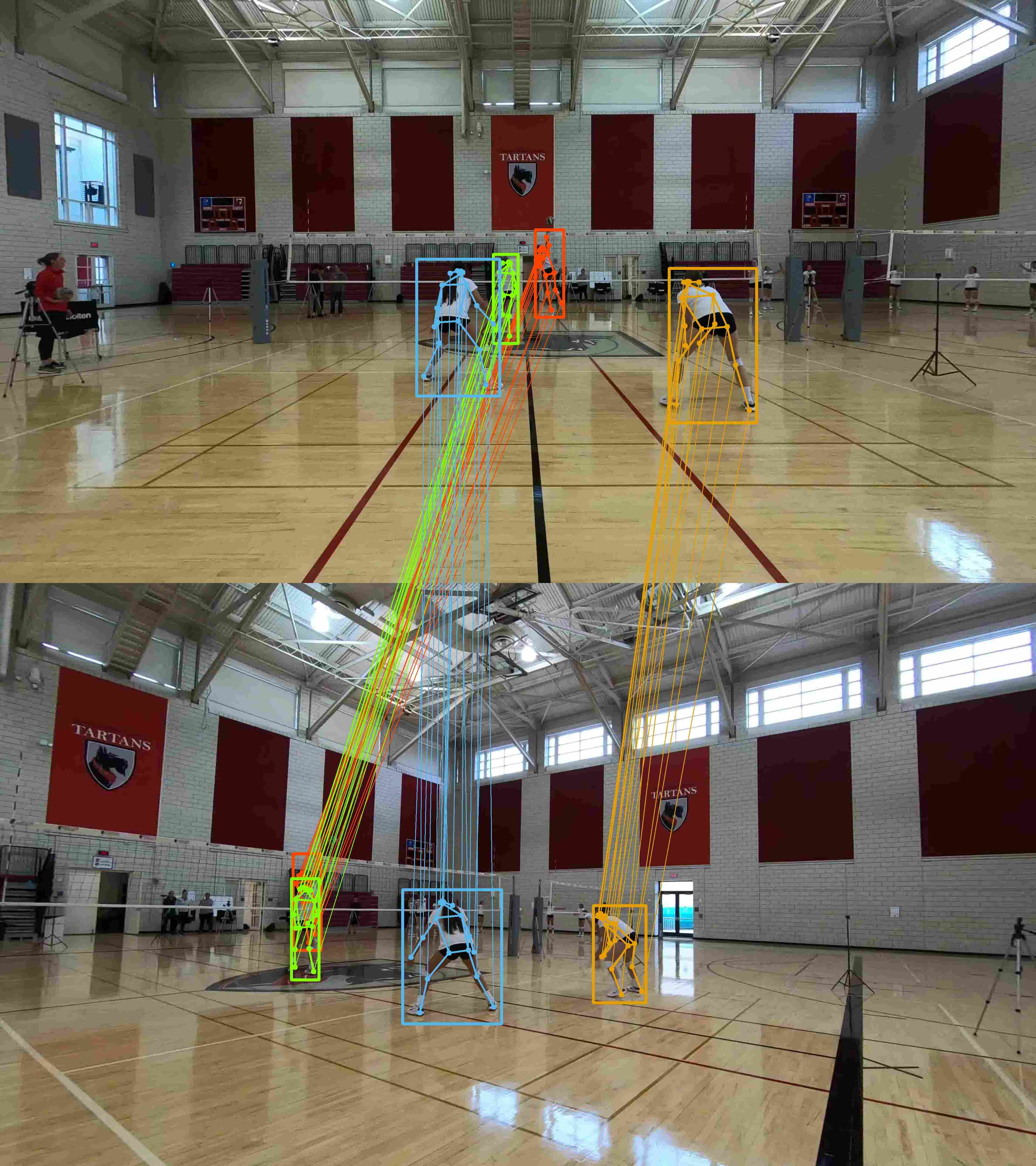}{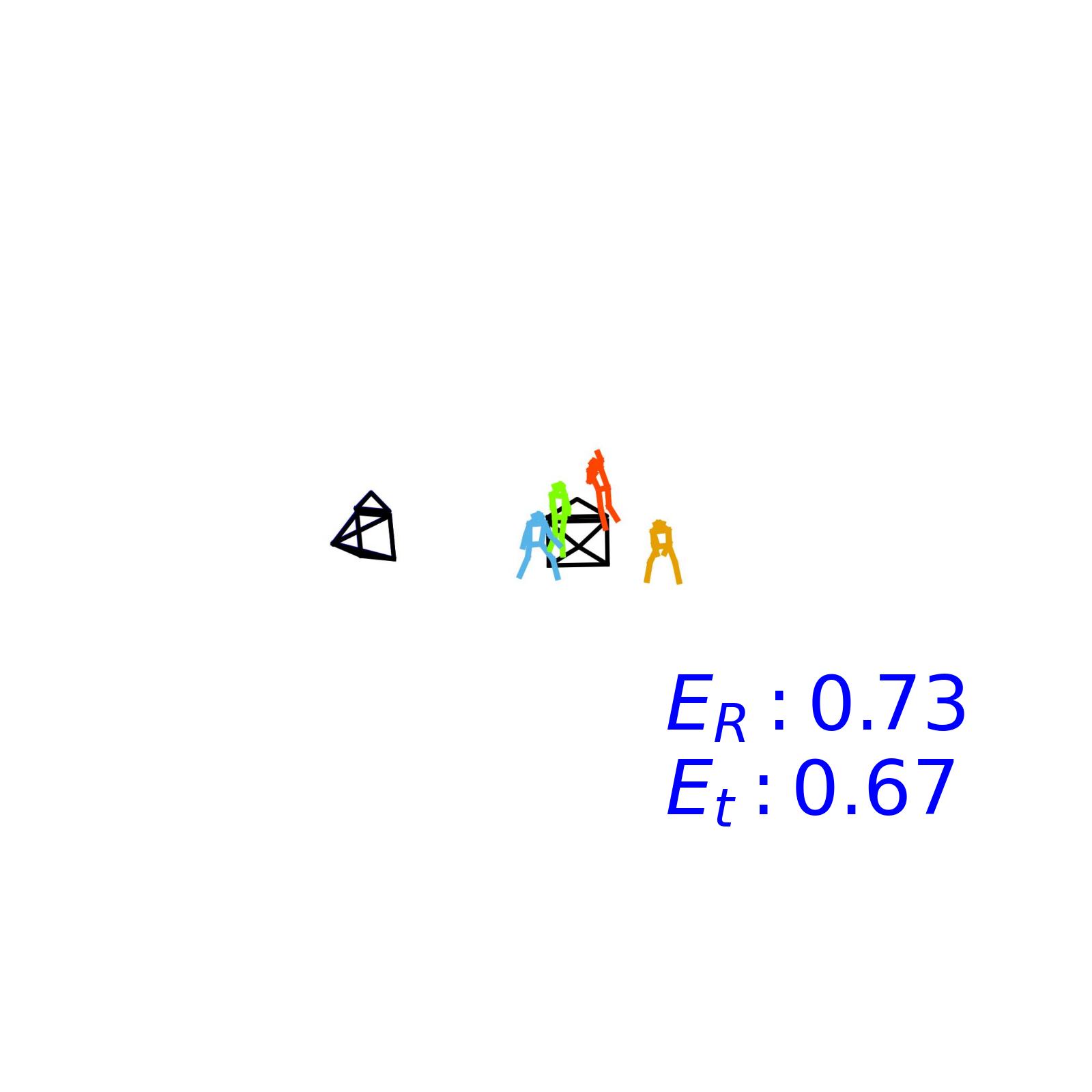} & 
    \raisebox{-0.5\totalheight}{%
    \includegraphics[width=0.13\linewidth]{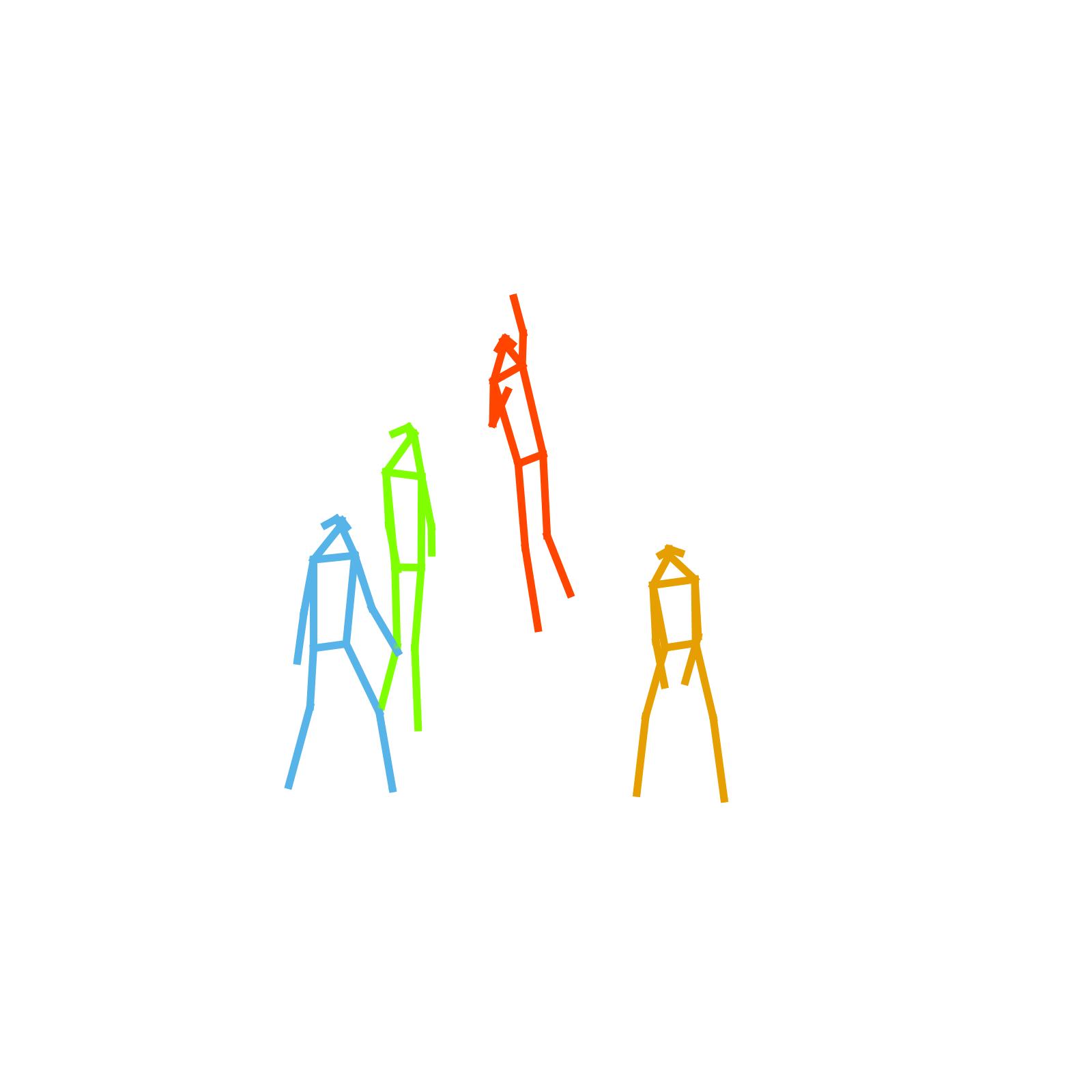}
  }%
    \\
    \hline
    \end{tabular}
    \caption{Two-view extrinsic calibration and matching results. Colored lines across the views represent the estimated correspondences. The ground truth and estimated camera poses are shown in red and green, respectively. The 3D poses were triangulated using the estimated camera poses. }
    \label{fig:two_view_calib_appendix}
\end{figure}

\begin{table}[t]
    \centering
    \begin{tabular}{l |  c | c  c  c | c  c  c  }
    \toprule
    \multirow{2}{*}{Scenario}  & \multirow{2}{*}{Camera IDs} & \multicolumn{3}{c|}{Motion Averaging~\cite{govindu2001combining}}&\multicolumn{3}{c}{Bundle Adjustment~\cite{Hartley00}}\\
    &&$E_R$ & $E_{\mathbf{t}}$ &$E_{\mathrm{2D}}$ & $E_R$ & $E_{\mathbf{t}}$ & $E_{\mathrm{2D}}$  \\
    \midrule
    
    \multirow{2}{*}{\Dpig~\cite{an2023three}} &  \multirow{2}{*}{[0,1,5,9]} & 2.20& 4.16& 5.86& 2.42& 3.73& 3.23\\
& &\cellcolor{gray!30}4.39& \cellcolor{gray!30}4.91& \cellcolor{gray!30}4.86& \cellcolor{gray!30}2.43& \cellcolor{gray!30}3.75& \cellcolor{gray!30}3.23\\ \midrule
\multirow{2}{*}{\Ddog~\cite{an2023three}} &  \multirow{2}{*}{[4,6,7,9]} & 7.50& 5.24& 10.10& 5.93& 4.10& 3.09\\
& &\cellcolor{gray!30}7.39& \cellcolor{gray!30}4.90& \cellcolor{gray!30}9.47& \cellcolor{gray!30}5.97& \cellcolor{gray!30}4.11& \cellcolor{gray!30}3.09\\ \midrule
\Dtoddler~\cite{Joo_2017_TPAMI} &  [7,12,22,28] &0.85& 1.07& 2.62& 0.29& 0.38& 1.76\\ 

    \bottomrule
    \end{tabular}
    \caption{Evaluation of extrinsic calibration and matching in multi-view setups. Gray-shaded cells denote the class-agnostic results. }
    \label{table:multi_quan}
\end{table}
\begin{figure}[t]
    \centering
    \def\vlab#1{\rotatebox[origin=c]{90}{#1}}
    \newcommand{\incgthree}[3]{%
    \raisebox{-0.5\totalheight}{%
      \includegraphics[width=0.32\linewidth]{#1}%
    } & 
    \raisebox{-0.5\totalheight}{%
      \includegraphics[width=0.32\linewidth]{#2}%
    } & 
    \raisebox{-0.5\totalheight}{%
      \includegraphics[width=0.23\linewidth,trim=90 90 90 90,clip]{#3}%
    }%
  }
    \begin{tabular}{c|@{}c@{}c@{}c@{}}
    & Input &  Cross-view Correspondences & 3D Reconstruction \\
    
    \hline

    \vlab{\Ddog} &
    \incgthree{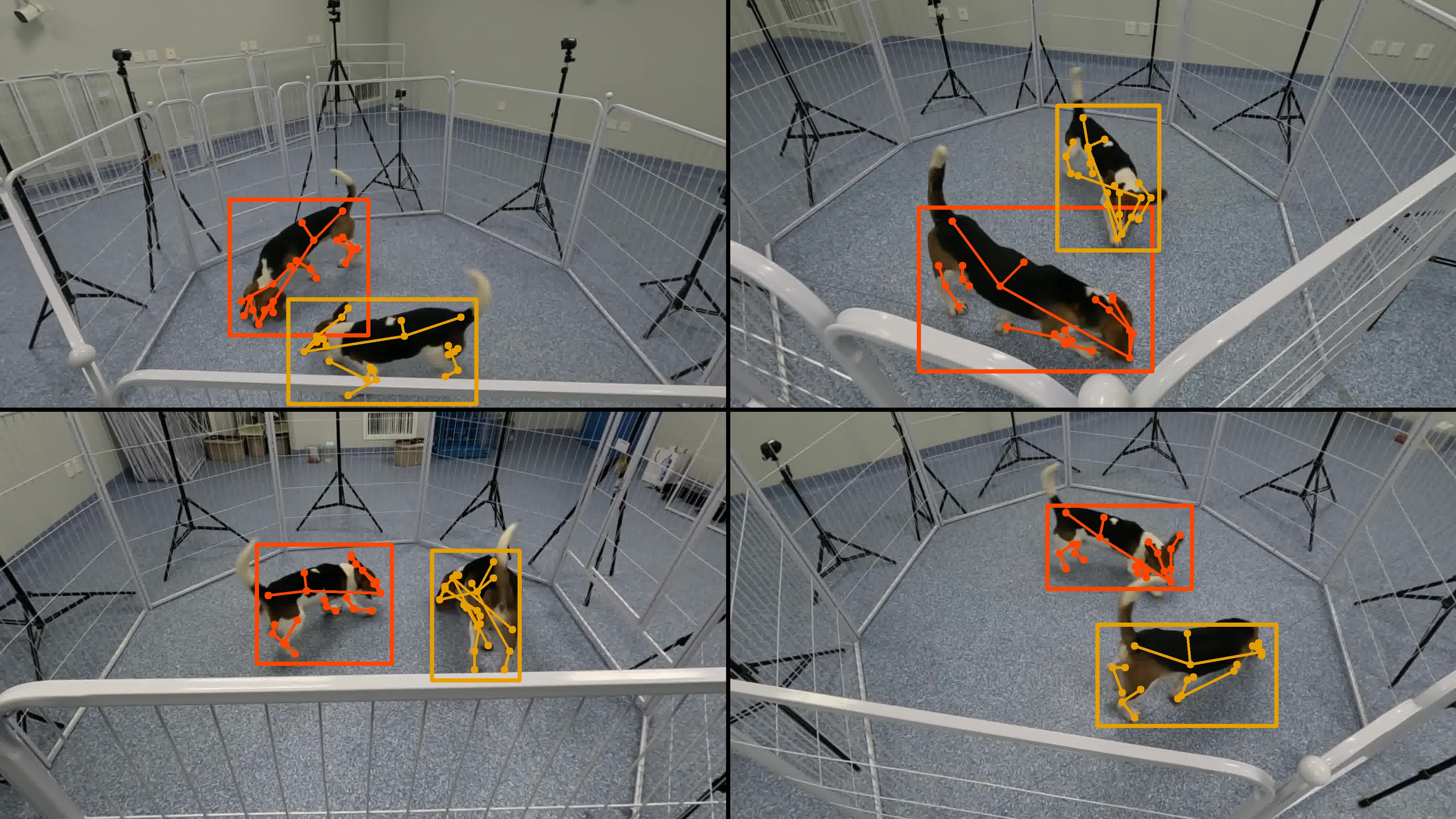}{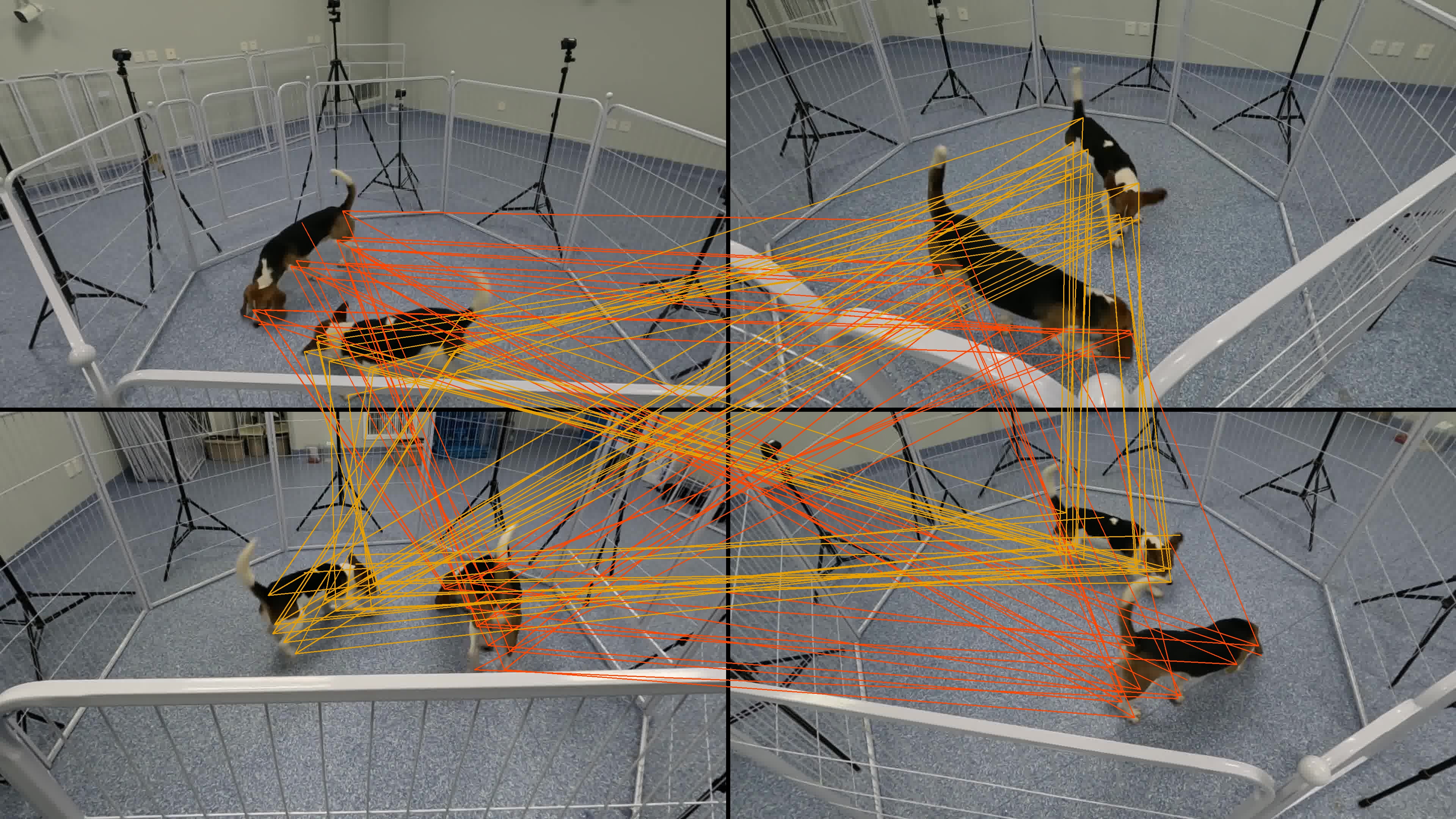}{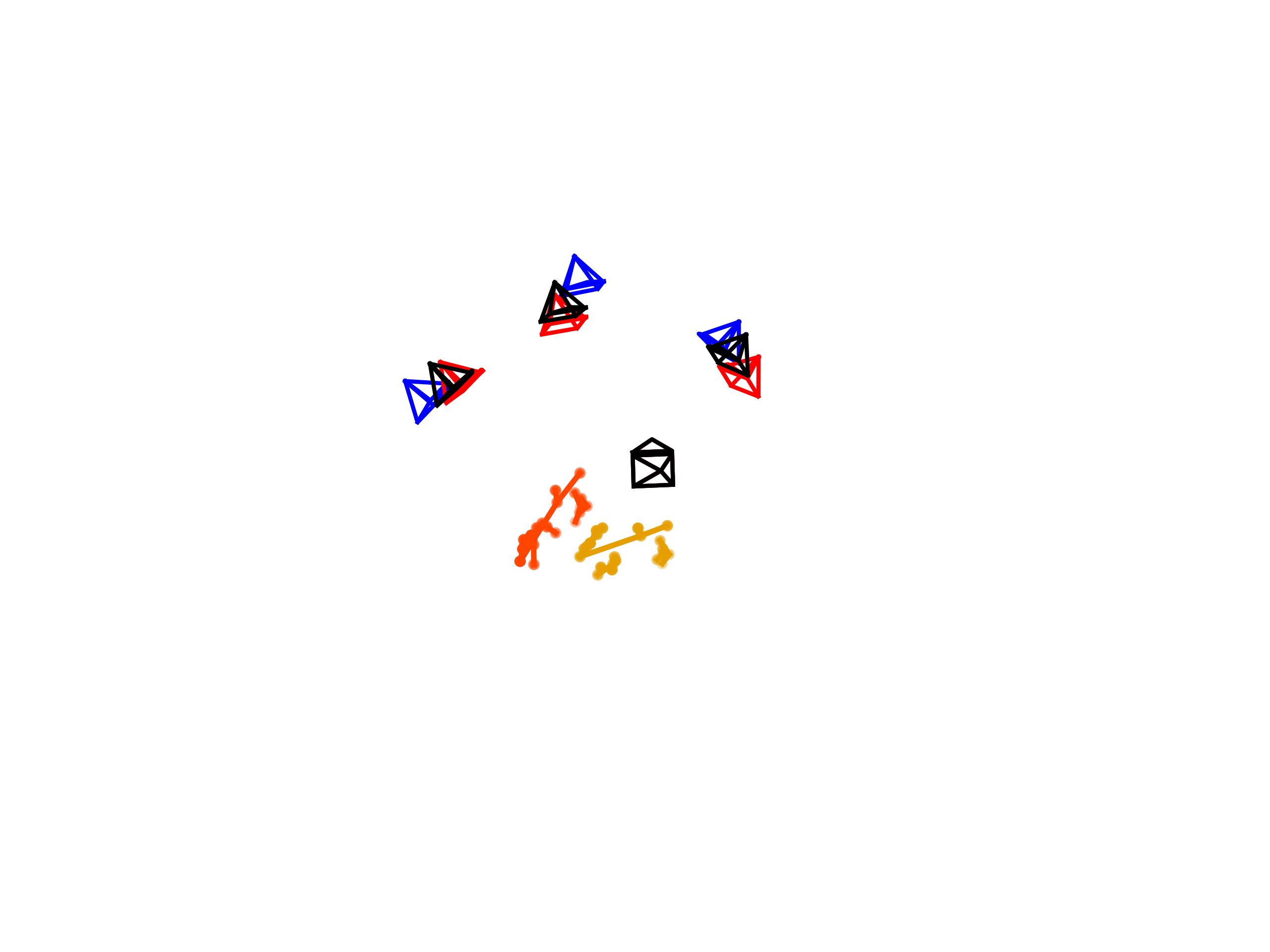}\\ \hline

    \end{tabular}
 
    \caption{Visualization for multi-view calibration results. 
    The inputs (left) show detected 2D poses of multiple instances from off-the-shelf pose estimators, with the colored bounding boxes indicating the instance IDs. Cross-view correspondences are shown as colored lines (Middle). Multi-view integration (Right) refines the camera poses and reconstructs the 3D poses.
    The ground-truth cameras are in black, the initial camera poses after motion averaging are in blue, and the final bundle-adjustment results are shown in red. }
    \label{fig:multi_qual_apendix}
\end{figure}

\section{Additional Calibration Results}\label{sec:additional_results}
Fig.~\ref{fig:two_view_calib_appendix} shows a qualitative comparison of our two-view calibration and matching results, demonstrating a consistent performance across different scenarios. Table~\ref{table:multi_quan} presents quantitative multi-camera calibration results for both the class-specific and agnostic models. This nearly identical performance highlights the effectiveness and generalizability of our approach to unseen animal species during training. The qualitative multi-view calibration results are shown in Fig.~\ref{fig:multi_qual_apendix}. Our method refines camera poses through the nonlinear optimization of reprojection errors within a unified coordinate system, leading to accurate camera poses and 3D reconstructions.
\section{Training Details and Implementation}
\subsection{Training Data Generation}
Training data were generated by sampling diverse camera viewpoints relative to the 3D poses of the target. Specifically, we placed 100 cameras uniformly over the unit hemisphere using the Fibonacci sphere method, with their orientations directed towards the origin. To model the camera roll variations, we uniformly sampled 20 rotations around the optical axis of the camera from each viewpoint. We randomly sampled 3,000 camera pairs and applied orthographic projections to obtain 2D poses for both views. Each training pair consists of 2D-2D pose correspondences and the associated relative camera rotation. The synthesized pairs are divided into training, validation, and test sets in a ratio of 7:2:1. To enhance the robustness against partial occlusions, we augmented the training data by randomly masking 10\% to 30\% of the joints in the 2D poses. Table \ref{tab:dataset_split} summarizes the details of the datasets used in the experiments.
\begin{figure}[t]
    \centering
    \includegraphics[width=\linewidth]{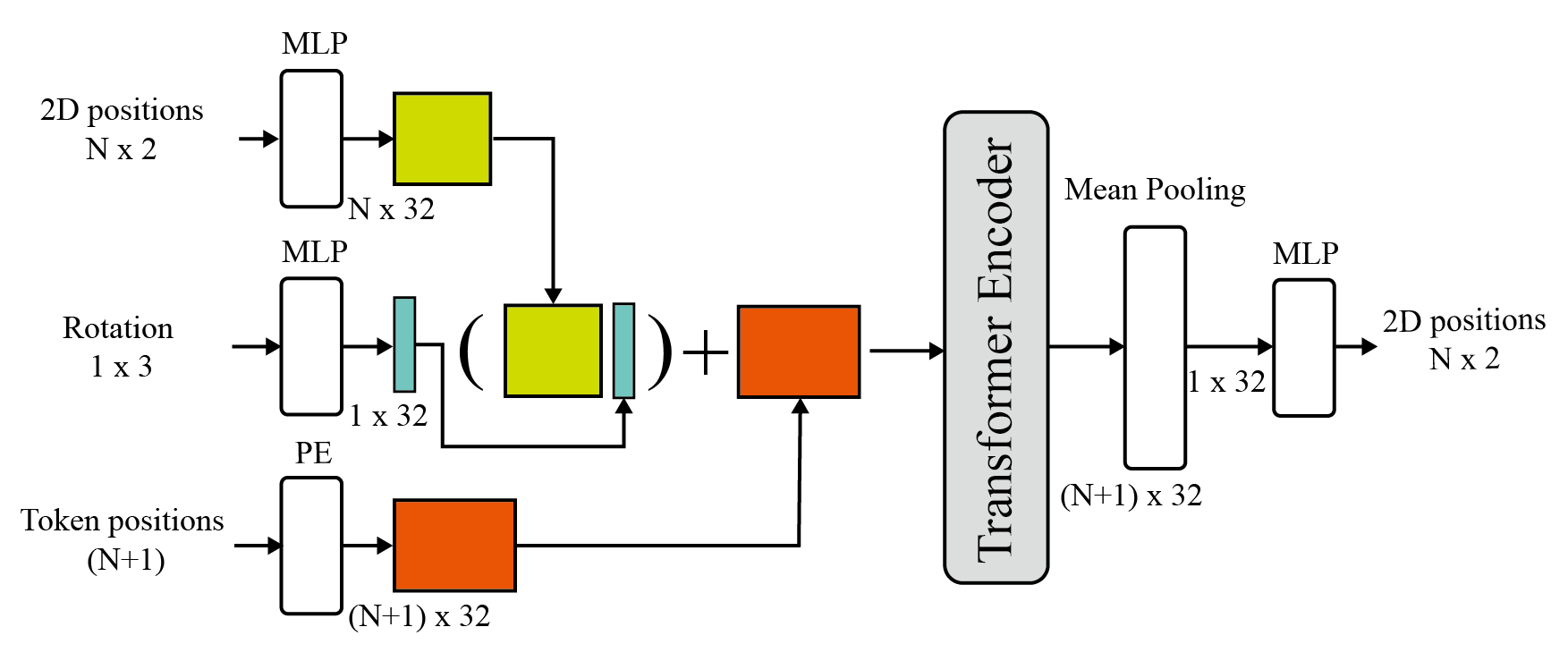}
    \caption{Network architecture of SteerPose}
    \label{fig:network_architecture}
\end{figure}

\subsection{Network architecture of SteerPose}
\def\embeddim{32}

Fig.~\ref{fig:network_architecture} illustrates the network architecture of the proposed SteerPose.  The input consists of 2D positions of $N$ joints in the image and three parameters representing the rotation by the Rodrigues vector.
Each of the $N$ 2D joint positions and the rotation vector are projected onto the \embeddim-dimensional vectors by MLP layers.

Each of the \embeddim-dimensional vectors is combined with a positional encoding by adding them element-wise.  Here we used a look-up embedding with $N+1$ tokens for positional encoding.  The embeddings with the positional encoding are used as input tokens of the transformer encoder with five layers.  The output tokens of the transformer encoder are averaged together to form a single \embeddim-dimensional vector by mean pooling and then transformed to $N$ 2D joint positions.

\subsection{Implementation Details}
We used the Adam optimizer with a learning rate of 0.01 for 1000 iterations. We schedule the learning rate using a linear scheduler with a start factor of 1.0 and an end factor of 0.01. The geometric consistency loss is activated after 40\% of the total optimization iterations and once a reliable set of multiple bi-directional matches has been established. The scale parameter $\alpha = 3$ was used to calculate the similarity score. 

For each scenario, we sampled 100-120 image pairs from multi-view sequences. The calibration success was determined by a reprojection error of less than 10 pixels for the reconstructed 3D poses. Otherwise, we retried the calibration with a new random initial rotation, up to a maximum of 5 attempts.

\begin{table}[t]
    \centering
    \resizebox{\columnwidth}{!}{%
    
    \begin{tabular}{l|c|c|c}
        \toprule
        & Train & Calibration (Synthetic) & Calibration (Real) \\
        \midrule
        \Dcheetah~\cite{joska2021acinoset}  & Jules flick1 (20190309) & Romeo flick (20190227) & Romeo flick (20190227) \\
        \Dpig~\cite{an2023three} & 0--1400 & 1400--1750 & 0--1750 (HRNet) \\
        \Ddog~\cite{an2023three} & 0--80 & 80--112 & 0--112 (SuperAnimal~\cite{ye2024superanimal}) \\
        \Dpigeon~\cite{naik20233d} & Sequence8\_n01\_01072022 & Sequence5\_n05\_01072022 & Sequence5\_n05\_01072022 \\
        \Dtoddler~\cite{Joo_2017_TPAMI} &171204\_pose1--6& 170915\_toddler5 & 170915\_toddler5 (RTMO~\cite{lu2023rtmo}) \\
        \Dvolleyball~\cite{khirodkar2023egohumans} &171204\_pose1--6& 001\_volleyball & 001\_volleyball (RTMO~\cite{lu2023rtmo})\\
         \bottomrule
    \end{tabular}
    }
    
\caption{ Dataset for training and calibration in synthetic and real scenarios, where the numbers indicate the frame indices. }

\label{tab:dataset_split}
\end{table}

\end{appendices}
\end{document}